\documentclass[11pt]{article}

\usepackage[preprint]{acl}

\usepackage{times}
\usepackage{latexsym}

\usepackage[T1]{fontenc}

\usepackage[utf8]{inputenc}

\usepackage{microtype}
\usepackage{booktabs}
\usepackage{multirow}

\usepackage[varqu]{inconsolata}
\usepackage{upquote}

\usepackage{graphicx}
\usepackage[most]{tcolorbox}
\tcbuselibrary{skins,breakable}

\usepackage{pifont}

\usepackage{array}
\usepackage{amsmath}
\usepackage{enumitem}
\usepackage{subcaption}
\usepackage{multirow}
\usepackage{booktabs}
\usepackage{marvosym}
\usepackage{stfloats}

\title{Benchmarking Large Language Models for Detecting Satirical Code-Mixed Sentences}

\author{
  \textbf{Răzvan-Alexandru Smădu\textsuperscript{1,2}},
  \textbf{Ioana Buhnila\textsuperscript{3,}\thanks{Equal contribution.}},
  \textbf{Daniela Tanase\textsuperscript{1,}\footnotemark[1]},
\\
  \textbf{Dumitru-Clementin Cercel\textsuperscript{1,2,\Letter}},
  \textbf{Florin Pop\textsuperscript{1,2,4}},
  \textbf{Mihaela-Claudia Cercel\textsuperscript{5,6}}
\\
\\
  \textsuperscript{1}National University of Science and Technology POLITEHNICA Bucharest, \\
   Bucharest, Romania \\
\textsuperscript{2}{Academy of Romanian Scientists, Bucharest, Romania} \\
  \textsuperscript{3}Center for Data Science in Humanities, Chosun University, South Korea \\
  \textsuperscript{4}National Institute for Research \& Development in Informatics - ICI Bucharest, \\
Bucharest, Romania \\
  \textsuperscript{5}University of Bucharest, Bucharest, Romania \\
  \textsuperscript{6}Paris 1 Panthéon-Sorbonne University, Paris, France \\
  \texttt{dumitru.cercel@upb.ro}
}

\newtcolorbox{promptbox}[2][]{
  enhanced,                     
  colback=gray!5,               
  colframe=gray!50,             
  title={#2},                   
  colbacktitle=gray!20,         
  coltitle=black,               
  fonttitle=\bfseries\sffamily\small, 
  fontupper=\ttfamily\small,    
  arc=3pt,                      
  boxrule=0.6pt,                
  toptitle=3pt,                 
  bottomtitle=3pt,
  left=8pt,                     
  right=8pt,
  top=8pt,
  bottom=8pt,
  boxsep=0pt,
  breakable,                    
  before skip=10pt,             
  after skip=10pt,              
  #1
}

\newcommand{\prompt}[3]
{
\begin{promptbox}{#2}
#3
\end{promptbox}
}

\begin{document}
\maketitle
\begin{abstract}
Code-switching is a natural linguistic phenomenon, yet its impact on high-level reasoning tasks remains largely unexplored in under-resourced languages like Romanian. Natural language processing tasks, such as satire detection, rely heavily on nuanced cultural and syntactic cues that large language models struggle to capture. In this paper, we investigate four methods (i.e., random replacement, phrase translation, dependency parsing, and large language models) for generating synthetic code-mixed Romanian sentences with English, German, French, and Italian. We start with an existing dataset for satire detection, compiled from Romanian news articles and annotated at the sentence level by humans, and augment it with over 600k samples, namely CS-SeLeRoSa. We provide an in-depth analysis of the dataset generation process. Then, we evaluate several multilingual large language models across zero-shot, chain-of-thought, and fine-tuning regimes on both the synthetic corpus and a human-adjudicated gold benchmark, namely CS-SeLeRoSa-Gold. While older open models suffer substantial degradation (dropping to 37\%--55\% macro-F1), newer models like Gemma 4 31B achieve 68.1\% zero-shot, rivaling frontier models like GPT-5 mini (69.5\%). Furthermore, fine-tuning significantly reduces formatting errors and improves the results, increasing Gemma 4 31B to 77.5\% on CS-SeLeRoSa and 72.6\% on the gold standard, outperforming frontier zero-shot baselines.
\end{abstract}

\section{Introduction}
\label{sec:introduction}

Code-switching or code-mixing is a common phenomenon in multilingual communities in which speakers alternate between two or more languages within the same sentence \cite{guzman17_interspeech}. 
While early natural language processing (NLP) and linguistic studies primarily mapped constraints on widely spoken pairs, the structural behavior of code-switching shifts significantly when applied to morphologically rich languages \cite{10.1093/oso/9780198240594.001.0001}.

Code-switching was previously explored in the context of satire, irony, and humor detection, but primarily in other combinations, such as Hindi-English \cite{bansal-etal-2020-code,aggarwal-etal-2020-really,mazumder-etal-2025-revealing} or Spanish-English. For example, \citet{bansal-etal-2020-code} investigated Hindi-English code-mixed tweets for detecting humor, sarcasm, and hate speech, arguing that code-mixing can be used to emphasize these phenomena. They also show that it can improve the model's performance on downstream tasks. \citet{belani-flanigan-2023-automatic} proposed a dataset generated from transcribing day-by-day conversations for Spanish-English with various labels, including usage of code-mixing for jokes. \citet{pant-dadu-2020-towards} proposed mapping the code-switched text to a higher-resource language through transliteration and showing improved performance for sarcasm and hate speech detection. However, this remains an underexplored area due to data scarcity. Consequently, many studies, including ours, rely on synthetic generation. This practice has positively spurred the community to develop diverse generation techniques, even though most evaluations continue to focus on downstream tasks using existing, naturally code-mixed text \cite{aguilar-etal-2020-lince,sheth-etal-2025-comi}.

Mixing Romanian with other European languages is a naturally occurring phenomenon \cite{wilhelm2017,buzila2025}. In the diaspora, many Romanians live, especially in countries such as Germany, France, Italy, Spain, and the UK, making it one of the largest diasporas \cite{oecd_talent_2019}. Therefore, learning these languages is very common. Given these socio-linguistic factors and the scarcity of code-mixed satire data, we study four European languages, namely, English, German, French, and Italian, as target languages for code-switching with Romanian.

Automatic satire detection in news articles was introduced by \citet{burfoot-baldwin-2009-automatic}, which defines a satirical article as one that deliberately exposes real-world individuals, organisations or events to ridicule, incorporating devices such as irony and exaggeration. Unlike fake news, which shares similar linguistic features but has divergent intent, satire is often misinterpreted as misinformation \cite{levi-etal-2019-identifying}. For the Romanian language, SaRoCo \cite{rogoz-etal-2021-saroco} represents the first dataset for satire news detection, addressing the corpus-level classification. Recently, \citet{smuadu2025selerosa} proposed the SeLeRoSa dataset with higher sentence-level granularity. However, the intersection of satire detection with code-mixed text remains largely underexplored.

In this work, we create the first code-switched sentence-level Romanian satirical news dataset, namely CS-SeLeRoSa. It employs four different code-switching methods and four language pairs (English, German, French, and Italian). We provide an extensive analysis of the generated dataset using various metrics and benchmark several LLMs on the code-mixed dataset.
We summarize the main contributions of this work as follows:
\begin{itemize}
\item We introduce CS-SeLeRoSa, the first synthetic sentence-level code-mixed Romanian satirical news dataset, exploring four generation methods and four language pairs.
\item We provide an in-depth linguistic and semantic analysis of the generated dataset across diverse code-mixing metrics.
\item We curate a human-adjudicated gold subset, CS-SeLeRoSa-Gold, verified by expert annotators, thoroughly analyzed.
\item We benchmark several LLMs on both datasets and evaluate their performance across zero-shot, chain-of-thought, and fine-tuning settings.
\end{itemize}

\section{Related Work}

\textbf{Code-Switching Tasks.}
Code-switching was evaluated in various natural language processing tasks. For instance,
\citet{krishnan-etal-2021-multilingual} address slot filling and intent prediction. They take the set of text in the original language and, for each item, generate random translations for chunks, then perform label-level alignments.
A similar approach was applied by \citet{tarunesh-etal-2021-machine} for the Hindi language. In addition, they proposed a machine translation system which uses an encoder-decoder architecture trained using multiple losses: denoising autoencoding, backtranslation, and cross-entropy.
\citet{sterner-teufel-2025-minimal} investigated the alignment in choosing natural code-switched sentences over manipulated variants and showed that larger models tend to align better with the preferences of fluent bilingual speakers, who understand the social and syntactic nuances of code-switching. In a follow-up work, \citet{sterner-teufel-2025-code} analyzed the syntactic difference between naturally occurring code-switched sentences and manipulated ones. They replaced the words with their parts-of-speech tags and, from the resulting structure, distinguished between sentences in minimal code-switching pairs. They proposed a graph neural network-based approach that achieves scores comparable to those of large language models across various language pairs.

Previously, code-switching was evaluated alongside the detection of satire, irony, and sarcasm \cite{bansal-etal-2020-code}, specifically for Hindi-English (also known as Hinglish). \citet{shah-maurya-2021-effective} detected sarcasm from code-mixed social media posts, incongruity through sub-word embeddings in a bidirectional LSTM-based architecture~\cite{DBLP:journals/neco/HochreiterS97}. \citet{10.1007/978-3-030-45442-5_18} proposed the first multilingual and multicultural detection system for irony, in French, English, and Arabic. They evaluated feature-based models, such as random forests, and neural network models, such as a convolutional neural network, noting challenges in the dataset, such as informal language use and Arabic-specific challenges. Recently, more work has begun to address code-mixing in Arabic \cite{hamed-etal-2025-survey}.

\textbf{Code-Switched Datasets.}
\citet{pratapa-etal-2018-language} proposed to investigate the role of the equivalence constraint theory from linguistics to generate code-mixed text. Their approach suggested that using curriculum learning on generated samples can reduce the perplexity of language models.
\citet{salaam-etal-2022-offensive} created a real-world dataset and generated 30k synthetic sentences for code-mixing, for English to French, Spanish, and German. Multiple languages have been evaluated for code-mixing, and one of the most common combinations is Hindi-English \cite{dogruoz-etal-2021-survey}.
Large language models (LLMs) have been utilized in various code-mixed tasks \cite{zhang2025chaillmsimprovingcodemixed,lee-etal-2024-commit}.
Recently, \citet{xie2026switchlingua} proposed SwitchLingua, the first large-scale multilingual and multi-ethnic code-switching dataset, consisting of 12 languages and over 80 hours of audio. However, it does not include Romanian. They also proposed LinguaMaster, a data synthesis method based on an agentic framework. 

\textbf{Satire Detection.}
Automatic satire detection has been studied extensively in the literature, with approaches progressing from surface-level lexical features to structured reasoning, adversarial debiasing, and multi-modal architectures.
\citet{goldwasser-zhang-2016-understanding} argued that satire detection should not be treated as a traditional text classification task but rather as a common-sense reasoning task. Focusing on political satire, they showed that satirical cues rely on finding patterns of absurdity rather than lexical word-level features.
Similarly, \citet{yang-etal-2017-satirical} put satirical news classification in the context of fake news detection, noting both share similarities and can potentially be deceptive or even harmful. They showed that satirical cues are not uniformly distributed across articles and argued that paragraph-level features are more relevant than document-level features for psycholinguistic, stylistic, and structural signals.
\citet{mchardy-etal-2019-adversarial} proposed adversarial training as a method to identify and remove confounding variables from model representations. They showed that attention-based models were relying on superficial source-specific cues rather than genuine satirical features, and demonstrated that adversarial training can help the model focus on content-based signals.
\citet{li-etal-2020-multi-modal} investigated multi-modal inputs for satire detection, focusing on image and text modalities that could be combined in an unrelated fashion to make the news content absurd or ridiculous. Their results showed that unimodal models are outperformed by multimodal ones.
Later, \citet{nandy-etal-2024-yesbut} framed satire as involving conflicting scenarios, in which a scenario is juxtaposed with an ironical one that creates incongruity. They evaluated vision-language models on their ability to detect such satirical content.
\citet{abdalla-etal-2024-comparison} investigated large language models on the satire detection task, showing that chain-of-thought prompting and structured reasoning were key to achieving high performance.

\section{Method}

\subsection{Dataset}

\begin{figure*}[!t]
\centering
\includegraphics[width=\textwidth]{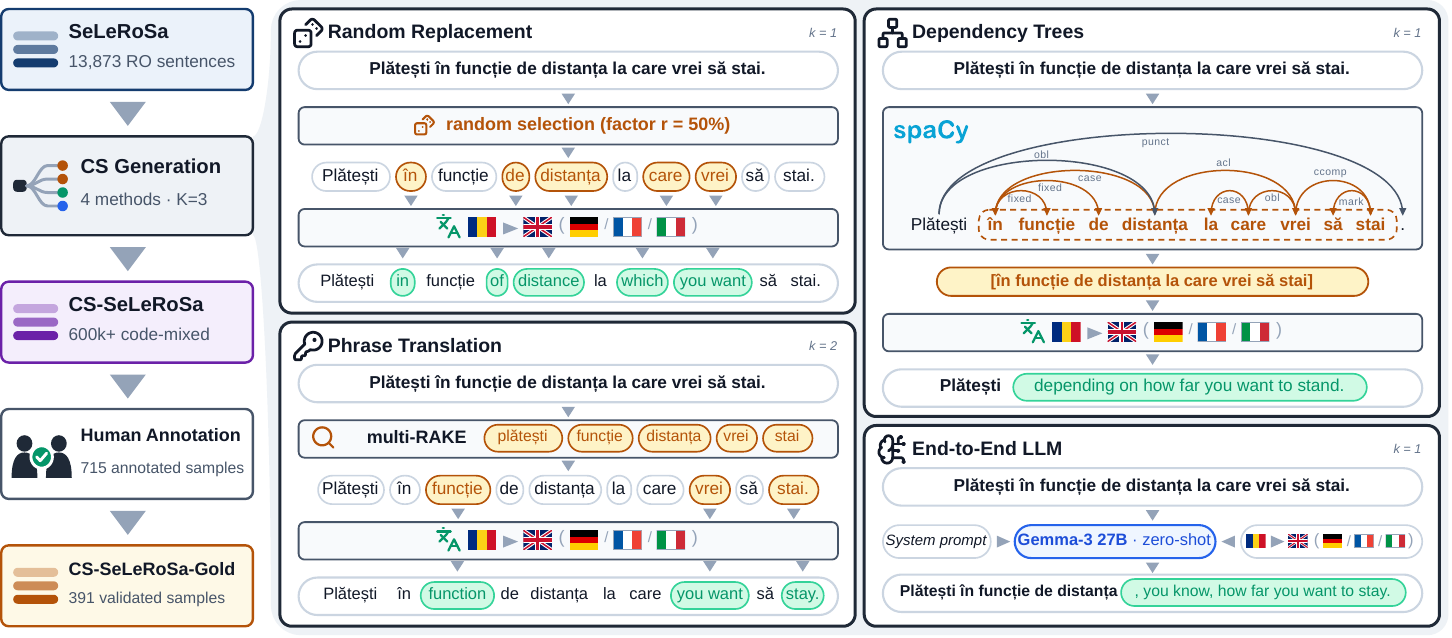}
\caption{End-to-end dataset generation for \textsc{CS-SeLeRoSa} and \textsc{CS-SeLeRoSa-Gold}. The data generation pipeline starts from the original Romanian monolingual \textsc{SeLeRoSa} corpus. Four code-switching (CS) generation strategies are applied to create the \textsc{CS-SeLeRoSa} dataset across $K=3$ iterations, as presented in \S\ref{sec:cs_methods}: random replacement, phrase translation, dependency tree-based, and LLM generation. From the resulting 600k+ code-switched sentences, we select 715 unique sentences which are then human-validated to produce \textsc{CS-SeLeRoSa-Gold}. The four methods are illustrated on a sentence showing generated variants (e.g., $k=1$ or $k=2$) for English.}
\label{fig:generation_pipeline}
\end{figure*}

To generate the synthetic code-mixed dataset, we employ the SeLeRoSa dataset \cite{smuadu2025selerosa}, which contains approximately 13.8k manually annotated sentences from Romanian satirical and non-satirical news outlets. The dataset was already anonymized by replacing named entities with generic tags (\texttt{<PERSON>}, \texttt{<GPE>}, \texttt{<ORG>}, and \texttt{<LOC>}). We maintain this anonymization in the generated code-mixed dataset to prevent models from learning entity-specific satire cues and reduce unwanted biases. The official train/validation/test split is 9,800/2,000/2,073. We follow the same split in our experiments. To generate the code-switched dataset, we apply four techniques which are presented in \S\ref{sec:cs_methods}. The resulting large-scale dataset is called CS-SeLeRoSa. In addition, we also provide a smaller human-annotated subset of this dataset, called CS-SeLeRoSa-Gold, meant for testing only. The full pipeline is illustrated in Figure~\ref{fig:generation_pipeline}.

\subsection{Translation}
\label{sec:translation}

To generate the synthetic data, we investigate several large language models (LLMs) for machine translation. Translation is a key component in generating the synthetic code-mixed dataset. The generation of code-switched sentences involves considering the sentence in the source language (i.e., the matrix language is Romanian), and applying a code-switching method that requires accessing a translation system to translate words or phrases to the target language. Then, the translated words or phrases are reintegrated into the original sentence, according to the code-switching method, to form a code-mixed sentence. Note that not all methods (e.g., end-to-end LLM generation) require a separate translation step.
For our code-mixing generation setup, we select Gemma 3 27B \cite{DBLP:journals/corr/abs-2503-19786} to translate the text using the machine translation prompt in Appendix \ref{app:prompts}. An evaluation of multiple candidate models for translation is provided in Appendix \ref{app:translation_evaluation}.

\subsection{Synthetic Code-Mix Generation}
\label{sec:cs_methods}

\paragraph{Random Replacement.}
Introduced by \citet{krishnan-etal-2021-multilingual}, the random replacement approach performs chunk-level code-switching. For each input sentence, we split it into tokens at word boundaries and then randomly select a subset of words to replace with their translations. The number of words selected is given by a factor $r$ which in our experiments was set to 50\%. As opposed to the original implementation, we use an LLM-based machine translation as described in \S\ref{sec:translation}, and ignore the anonymization tags during sampling. We acknowledge that the random-replacement strategy naturally yields degraded outputs. However, we applied the method exactly as proposed by \citet{krishnan-etal-2021-multilingual} without additional filtering to maintain consistency with prior work.

\paragraph{Phrase Translation.} The phrase translation approach \cite{salaam-etal-2022-offensive} is a three-stage method: phrase identification, phrase translation, and phrase reintegration. For phrase identification, as opposed to the original implementation, we employ multi-RAKE \cite{rose2010automatic} to identify keyphrases, using the \texttt{vgrabovets\/multi\_rake} library from GitHub\footnote{\url{https://github.com/vgrabovets/multi_rake}}, which supports the languages we investigate.
According to the documentation and \citet{rose2010automatic}, the algorithm is not limited to keywords (i.e., single words); it can also return multi-word keyphrases. Note that the algorithm does not make a difference between the specific kinds of keyphrases (e.g., noun-phrase, verb-phrase, etc). But in our case, this distinction is not relevant. Instead, it is relevant that the phrase must be self-contained, i.e., as a standalone, it should be complete and have a meaning.
RAKE does that by removing stop words and punctuation. \citet{salaam-etal-2022-offensive}, who proposed the phrase-based code-switching approach, used the UCPhrase unsupervised model \cite{10.1145/3447548.3467397}, which was trained on English data. However, this approach does not support other languages out of the box. The \texttt{multi\_rake} package already supports Romanian and the other target languages, so we opted for this solution instead.
Next, in the phrase translation step, we use the LLM-based machine translation system from \S\ref{sec:translation} to translate the identified phrases. Finally, in the phrase reintegration step, we replace the original phrases with their translations. To increase diversity in code-mixed sentences, we randomly select which phrases to replace using uniform sampling. The motivation for this approach is that the chosen groups of words should make sense on their own to be translated coherently.

\paragraph{Dependency Trees.}
\citet{gregorius-okadome-2022-generating} proposed code-switching text generation using a machine translation system and a syntactic dependency parser. This approach does not require training models on parallel corpora; its components are plug-and-play and already available. The idea relies on finding a code-switching point by leveraging the properties of the dependency graph
and then using the machine translation system (e.g., from \S\ref{sec:translation}) to translate the phrase spanned by the sub-tree to the other language.
This takes the phrase translation idea to the next level by considering sub-trees that are large enough to be self-contained, providing more context for the translator. However, it is therefore limited by the quality of the dependency parser and the machine translation system. For the implementation, we rely on the spaCy\footnote{\url{https://spacy.io/}} dependency parser for Romanian, \texttt{ro\_core\_news\_lg}.

\paragraph{Large Language Models.}
Large language models (LLMs) have demonstrated impressive capabilities in various natural language processing tasks, including text generation, translation, and code-mixing \cite{yang2025codemixbenchevaluatingcodemixingcapabilities}. We explore the use of LLMs to generate code-mixed sentences directly in the zero-shot setting. Therefore, we prompt the model, in our case, Gemma 3 27B with temperature set to 0.8, to generate a code-mixed version in the given languages, preserving the original meaning (see Appendix~\ref{app:prompts} for the prompts). Note that in this setting, the generation is performed end-to-end by the LLM, and therefore the translation does not follow \S\ref{sec:translation}.

\subsection{Dataset Statistics}

Table~\ref{tab:cs_dataset_statistics} summarizes the statistics of the generated code-mixed dataset CS-SeLeRoSa. We use 4 language pairs, i.e., Romanian (RO) to English (EN), German (DE), French (FR), and Italian (IT), and 4 code-mixing methods described in \S\ref{sec:cs_methods}. For each language pair and method, we generate at most $K=3$ code-mixed examples (with IDs from 1 to 3, 0 indicating the original sentence). We maintain the original dataset split into train, validation, and test sets, by the original source sentence.

The original Romanian dataset has 13,873 samples, which were processed preserving the original train/validation/test splits. For every sample, we apply 4 code-switch generation methods (random, phrase, dependency, and LLM) on 4 target languages (English, French, German, Italian), yielding at most 16 combinations per sample (i.e., $16 \times 13,873 = 221,968$ total). We repeat each combination $K=3$ times, resulting in at most 665,904 code-switched samples. However, some generation attempts fail to produce all 3 variants. Additionally, the dependency-based method failed to produce sub-trees from 3 single-word sentences; therefore, 12 such cases are ignored (3 sentences across 4 languages), resulting in a maximum of 221,956 samples for k=1. The counts decrease further for k=2 (221,716) and k=3 (219,306) as sentences with fewer than 3 variants are excluded at higher k levels. For example, for some short samples, we could not extract three different sub-trees for the dependency method. Further analysis regarding code-mixing quality, lexical diversity, and semantic preservation across all methods is presented in Appendix~\ref{sec:cs_evaluation}.

For the human-annotated gold set, CS-SeLeRoSa-Gold, we randomly sample 50 original sentences and extract their corresponding code-mixed variants across the four methods and four languages. After deduplication and removing failed translations, 715 samples remained for annotation. Following human adjudication, 391 high-quality samples were retained. Appendix~\ref{app:human_evaluation} presents the annotation protocol, data selection, and statistics across annotation and the final gold set.

\begin{table}[t!]
\centering
\setlength{\tabcolsep}{2pt}
\small
\begin{tabular*}{\columnwidth}{@{\extracolsep{\fill}}lcccc@{}}
\toprule
\multicolumn{5}{c}{\textbf{Language}} \\
\midrule
RO & EN & DE & FR & IT \\
13,873 & 165,725 & 165,725 & 165,725 & 165,803 \\

\bottomrule
\end{tabular*}

\vspace{0.1em}

\begin{tabular*}{\columnwidth}{@{\extracolsep{\fill}}lccc@{}}
\toprule
\multicolumn{4}{c}{\textbf{K (code-switch variation)}} \\
\midrule
k=0 & k=1 & k=2 & k=3 \\
13,873 & 221,956 & 221,716 & 219,306 \\
\bottomrule
\end{tabular*}

\vspace{0.1em}

\begin{tabular*}{\columnwidth}{@{\extracolsep{\fill}}lcccc@{}}
\toprule
\multicolumn{5}{c}{\textbf{Generation Method}} \\
\midrule
Original & Random & Phrase & Dependency & LLM \\
13,873 & 166,476 & 165,934 & 164,092 & 166,476 \\
\bottomrule
\end{tabular*}

\vspace{0.1em}

\begin{tabular*}{\columnwidth}{@{\extracolsep{\fill}}ccccc@{}}
\toprule
\multicolumn{5}{c}{\textbf{Split}} \\
\midrule
& Train & Test & Validation & \\
& 478,592 & 100,975 & 97,284 & \\
\bottomrule
\end{tabular*}

\caption{Summary of the entire CS-SeLeRoSa code-mixed dataset statistics.}
\label{tab:cs_dataset_statistics}
\end{table}

\section{Experimental Setup}
\label{sec:experimental_setup}

\subsection{Classification Task}
\label{sec:classification_task}

We address satire news detection as a binary classification task. Let $D = \{(x_i, y_i)\}_{i=1}^{N}$ be a dataset of size $N$, where $x_i = (x_i^1, x_i^2, \dots, x_i^{n_i})$ is a sentence of $n_i$ tokens, and $y_i \in \{0, 1\}$ represents the label of the sentence (0 for regular/factual, 1 for satire). In the monolingual setting, each word $x_i^j$ comes from the matrix language (e.g., Romanian). However, in the code-switching setting, $x_i^j$ originates from either the matrix language or the embedded target language. Importantly, the language assignment for each word must align with established code-switching linguistic theories. In our approach, we follow the Matrix Language Frame (MLF) model \cite{10.1093/oso/9780198240594.001.0001}, where each sentence preserves the grammatical structure of the original Matrix Language (ML) frame, i.e., Romanian in our case. The translated segments function as Embedded Language (EL) islands inserted into this frame. Following the constraints of the MLF model, these EL structures must be internally well-formed and semantically self-contained, meaning that if these translated parts are isolated, they remain inherently understandable. However, the random code-switching generation method often does not follow this principle, resulting in syntactically or semantically unnatural samples.

In our LLM-based evaluation protocol, the language model is given a prompt specifying the task, language pair, target classes, and input sentence. In the zero-shot and chain-of-thought settings, the model is provided with output formatting instructions, while chain-of-thought additionally requests a reasoning trace. The fine-tuning prompt, however, is minimal, containing only the task description, input sentence, and target answer. All classification prompt templates, along with their English translations, are presented in Appendix~\ref{app:prompts}.

The classes are discrete labels (i.e., \textit{``satire''} or \textit{``factual''}) which are replaced in the prompt templates. To avoid position bias \citep{DBLP:conf/nips/ZhengC00WZL0LXZ23}, we shuffle the positions of those labels in the text randomly (i.e., \textit{``satire or factual''} or \textit{``factual or satire''}). The models are tasked to produce XML-like structured responses, from which we extract the predicted labels. Reasoning traces are kept for further analysis, while the string labels are converted to numerical values, i.e., factual is mapped to 0 and satire is mapped to 1. Outputs that fail to parse are treated as classification errors (detailed in Appendix~\ref{app:instruction_following_error_rate}).

\subsection{Models}

On the CS-SeLeRoSa test set, we evaluate several LLMs in the zero-shot (ZS) setting, with and without chain-of-thought (CoT) prompting \cite{10.5555/3600270.3602070}, as well as fine-tuning (FT). We include open-weights and proprietary models in the evaluation. Specifically, for open-weights models, we evaluate Gemma 3 27B \citep{DBLP:journals/corr/abs-2503-19786}, Gemma 4 26B A4B and Gemma 4 31B \cite{DBLP:journals/corr/abs-2607-02770}, Llama 3.1 8B \cite{grattafiori2024llama3herdmodels}, and Romanian LLMs \citep{masala-etal-2024-vorbesti} such as RoMistral 7B, RoLlama 3.1 8B, and RoGemma 2 9B. For the closed-source models, we employ OpenAI's models, such as GPT-4.1 mini \cite{openai2024gpt4technicalreport}, GPT-5 mini \cite{singh2026openaigpt5card}, and GPT-6 Luna \cite{openai2026gpt6astra}. On the CS-SeLeRoSa-Gold test set, additionally, we also evaluate GPT-5 \cite{singh2026openaigpt5card}, GPT-6 Astra, GPT-6 Sol \cite{openai2026gpt6astra}, and GPT-6.1 Sol \cite{openai2026gpt6sol} as frontier models. The fine-tuning regime is exclusively performed on open-weights models, while proprietary reasoning models are evaluated only in the zero-shot setting, as they natively generate internal reasoning traces before responding.

The full configuration, including model checkpoints, hyperparameters, and computational costs, can be found in Appendix~\ref{app:experimental_setup}. The rest of the hyperparameters that are not mentioned are the default ones. We make public the inference code, including the full prompts, hyperparameters, seeds, and scripts to reproduce the experiments\footnote{\url{https://anonymous.4open.science/r/CS-SeLeRoSa-D85B/}}.

\subsection{Evaluation Metrics}

For the satirical code-switching classification, we focus our analysis on macro-averaged F1-score (F1). We also provide accuracy results in Appendix~\ref{app:breakdown_results}, which generally shows similar trends to F1-score. For open-weights models, we report the mean and standard deviation across $N=3$ independent runs with different random seeds. For proprietary models, we report single-run results, except for zero-shot evaluation on CS-SeLeRoSa-Gold where we report the mean and standard deviation over $N=3$ runs (proprietary CoT on Gold uses a single run). Statistical significance against the best-performing open-weights model (i.e., Gemma 4 31B) is indicated in the detailed appendix tables.

\section{Results}
\label{sec:results}

\begin{table*}[!t]
\centering
\small
\setlength{\tabcolsep}{2.5pt}
\begin{tabular}{lccccccccc}
\toprule
\multirow{2}{*}{\textbf{Model}} 
& \multicolumn{3}{c}{\textbf{SeLeRoSa}} 
& \multicolumn{3}{c}{\textbf{CS-SeLeRoSa}} 
& \multicolumn{3}{c}{\textbf{CS-SeLeRoSa-Gold}} \\
\cmidrule(lr){2-4} \cmidrule(lr){5-7} \cmidrule(lr){8-10}
& \textbf{ZS} & \textbf{CoT} & \textbf{FT} 
& \textbf{ZS} & \textbf{CoT} & \textbf{FT} 
& \textbf{ZS} & \textbf{CoT} & \textbf{FT} \\
\midrule
Llama 3.1 8B       & 46.3$_{\pm11.1}$ & 53.3$_{\pm1.9}$ & 69.0$_{\pm0.1}$ & 44.8$_{\pm15.2}$ & 47.9$_{\pm2.2}$ & 74.9$_{\pm0.8}$ & 43.0$_{\pm12.8}$ & 43.2$_{\pm2.2}$ & 65.7$_{\pm0.0}$ \\
Gemma 3 27B        & 49.9$_{\pm0.6}$  & 53.2$_{\pm0.4}$ & 74.3$_{\pm1.3}$ & 37.0$_{\pm0.3}$ & 41.3$_{\pm1.8}$ & 75.4$_{\pm0.8}$ & 36.1$_{\pm0.8}$ & 39.1$_{\pm3.2}$ & 67.2$_{\pm0.8}$ \\
Gemma 4 26B A4B    & 67.0$_{\pm0.0}$  & 68.1$_{\pm0.5}$ & 70.2$_{\pm0.5}$ & 63.7$_{\pm0.1}$ & 64.3$_{\pm0.1}$ & 70.2$_{\pm0.6}$ & 58.2$_{\pm0.3}$ & 61.5$_{\pm0.3}$ & 63.7$_{\pm0.8}$ \\
Gemma 4 31B        & 70.2$_{\pm0.1}$  & \textbf{69.4$_{\pm0.1}$} & \textbf{78.3$_{\pm0.8}$} & 68.1$_{\pm0.0}$ & \textbf{66.5$_{\pm0.0}$} & \textbf{77.5$_{\pm0.1}$} & \textbf{61.8$_{\pm0.2}$} & \textbf{63.0$_{\pm0.5}$} & \textbf{72.6$_{\pm0.4}$} \\
RoMistral 7B       & 41.2$_{\pm11.5}$ & 49.6$_{\pm0.6}$ & 72.5$_{\pm1.4}$ & 36.8$_{\pm14.4}$ & 48.3$_{\pm1.0}$ & 74.6$_{\pm0.6}$ & 35.7$_{\pm13.7}$ & 46.6$_{\pm1.5}$ & 70.4$_{\pm0.5}$ \\
RoLlama 3.1 8B     & 61.6$_{\pm0.8}$  & 50.4$_{\pm2.1}$ & 73.3$_{\pm0.8}$ & 52.6$_{\pm0.9}$ & 45.9$_{\pm1.0}$ & 73.4$_{\pm0.1}$ & 54.8$_{\pm1.5}$ & 46.9$_{\pm2.5}$ & 66.3$_{\pm0.9}$ \\
RoGemma 2 9B       & 60.3$_{\pm0.3}$  & 62.8$_{\pm1.4}$ & 73.1$_{\pm0.9}$ & 55.0$_{\pm0.7}$ & 60.0$_{\pm0.8}$ & 75.4$_{\pm0.0}$ & 56.8$_{\pm1.5}$ & 62.1$_{\pm0.9}$ & 68.9$_{\pm1.5}$ \\
\midrule
GPT-4.1 mini       & 63.3             & 68.0             & --              & 56.0          & 61.9          & --                      & 50.4$_{\pm0.8}$ & 59.8                     & -- \\
GPT-5 mini         & \textbf{71.3}    & --               & --              & \textbf{69.5} & --                      & --                      & 59.6$_{\pm2.0}$          & --                       & -- \\
GPT-6 Luna         & 68.1             & --               & --              & 67.0          & --                      & --                      & 58.1$_{\pm1.7}$          & --                       & -- \\
\midrule
GPT-5              & --               & --               & --              & --                      & --                      & --                      & 61.1$_{\pm1.6}$          & --                       & -- \\
GPT-6 Astra        & --               & --               & --              & --                      & --                      & --                      & 58.3$_{\pm0.3}$  & --                       & -- \\
GPT-6 Sol          & --               & --               & --              & --                      & --                      & --                      & 55.0$_{\pm1.0}$  & --                       & -- \\
GPT-6.1 Sol        & --               & --               & --              & --                      & --                      & --                      & 59.1$_{\pm1.7}$          & --                       & -- \\
\bottomrule
\end{tabular}
\caption{Macro-averaged F1-score (\%) results on the three evaluation sets from monolingual SeLeRoSa, CS-SeLeRoSa, and CS-SeLeRoSa-Gold. Models are evaluated in zero-shot (ZS), chain-of-thought (CoT), and fine-tuning (FT) settings. The first set of models are open-source models evaluated in all settings, the second set are closed-source models evaluated only in ZS or CoT settings, and the third set represents frontier models evaluated only on the CS-SeLeRoSa-Gold test set. Bold indicates the best performance per setting and dataset, across all models.}
\label{tab:overall_results}
\end{table*}

We present the overall results in Table~\ref{tab:overall_results} across the monolingual SeLeRoSa baseline, the synthetic CS-SeLeRoSa test split, and the human-adjudicated CS-SeLeRoSa-Gold benchmark. We organize our empirical findings around three core research questions: (RQ1) performance degradation under code-switching across model families, (RQ2) mitigating performance drops via Chain-of-Thought reasoning or fine-tuning, and (RQ3) failure modes across generation methods and target languages. Detailed per-setting breakdowns and full accuracy results are provided in Appendix~\ref{app:error_analysis}.

\textbf{RQ1: To what extent does code-switching degrade satire detection across open-weight and frontier models?}

Multilingual models are capable of processing diverse languages; however, code-switching introduces severe pragmatic and syntactic challenges. As shown in Table~\ref{tab:overall_results}, under the zero-shot setting, performance degradation is influenced by model scale and generation. Older-generation and smaller models collapse: Gemma 3 27B drops by 12.9 macro-F1 points (49.9 to 37.0), while Romanian-adapted models such as RoLlama 3.1 8B and RoMistral 7B drop by 9.0 and 4.4 points, respectively. For Llama 3.1 8B and RoMistral 7B, we also observe very high run-to-run variance ($\pm 15.2$ and $\pm 14.4$), which arises from parse and instruction-following failures under the zero-shot code-switching setting (see Table~\ref{tab:parse_failure_rates} from Appendix~\ref{app:instruction_following_error_rate}). In contrast, modern open-weights models prove far more resilient. Gemma 4 31B and 26B drop by only 2.1 and 3.3 F1 points, respectively, achieving among the highest overall scores and matching proprietary models such as GPT-5 mini ($-1.8$ drop) and GPT-6 Luna ($-1.1$ drop).

Evaluating models on the human-adjudicated CS-SeLeRoSa-Gold benchmark reveals an interesting interaction: weaker models remain largely consistent from synthetic to Gold ($\pm 2$ points; e.g., Gemma 3 27B changes by $-0.9$, Llama 3.1 8B by $-1.8$, RoGemma 2 9B by $+1.8$, and RoLlama by $+2.2$), whereas capable models drop noticeably (Gemma 4 26B by 5.5 points, Gemma 4 31B by 6.3 points, GPT-6 Luna by 8.9 points, and GPT-5 mini by 9.9 points). This indicates that these models partially overreact to certain patterns of the synthetic generation. When evaluating the frontier models on the gold set, models from the GPT-5 and GPT-6 families perform consistently in the 55.0\%--61.1\% macro-F1 band. Notably, Gemma 4 31B in zero-shot ($61.8 \pm 0.2$) is statistically on par with GPT-5 ($61.1 \pm 1.6$; $p \ge 0.05$, see Appendix~\ref{app:overall_cs_results}).

\textbf{RQ2: Can Chain-of-Thought reasoning or supervised fine-tuning mitigate code-switching degradation?}

To mitigate zero-shot performance drops, we evaluate models under chain-of-thought prompting and supervised fine-tuning settings. The effects of CoT are twofold. First, it improves performance for weaker models, yielding gains ranging from 3.1 points on Llama 3.1 8B (44.8 to 47.9) to 11.5 points on RoMistral 7B. CoT also stabilizes generation across seeds, drastically reducing variance (e.g., RoMistral variance drops from $\pm 14.4$ to $\pm 1.0$). Conversely, CoT provides no benefit for top-performing models, yielding a slight decline on Gemma 4 31B ($-1.6$ points on CS).

RoLlama 3.1 8B represents a negative outlier in the results, where chain-of-thought degraded performance by 6.7 points on CS-SeLeRoSa and 7.9 points on CS-SeLeRoSa-Gold. Appendix~\ref{app:rollama_baseline_degradation} shows that format failures do not represent the failure mode of the chain-of-thought regime, since the number of unparseable outputs overall does not increase proportionally. Instead, most errors come from label flips from factual to satirical, while chain-of-thought reasoning traces amplify the satire prediction bias, treating everyday colloquialisms as satire. This issue is resolved under supervised fine-tuning. Across the board, performance meaningfully improves, providing the highest scores on our benchmark. Specifically, Gemma 4 31B reaches a 77.5\% F1 score (with an increase of 9.4 points), while models such as RoMistral 7B and Gemma 3 27B achieve significant gains of 37.8 and 38.4 F1 points, respectively. In fact, fine-tuning models on code-switched data fully recovers performance to match or exceed the monolingual baseline (within $-0.8$ to $+2.3$ F1 points for most architectures, and outperforming by $+5.9$ points from 69.0\% to 74.9\% for Llama 3.1 8B). High performance gains are also observed on the gold set when compared with the zero-shot and chain-of-thought settings. The highest score is achieved, again, by Gemma 4 31B, outperforming zero-shot proprietary models by a significant margin of 11.5 points.

\textbf{RQ3: How do generation methods and target languages influence model failure modes?}

As mentioned earlier, a common reason for performance degradation is the high False Positive Rate (FPR), i.e., the models overpredict satire, as detailed in Appendix~\ref{sec:error_analysis}. Across settings, when code-switching is incorporated (mostly preserving the input prompts, while adding details regarding the code-switching regime), FPR raises to very high values, e.g., for RoLlama 3.1 8B it goes from 51.2\% to 82.5\% under random replacement, and for Gemma 3 27B it goes from 76.3\% to 97.9\% under the same setting. Meanwhile, GPT-5 mini remains robust against language variation and generation methods, with its FPR remaining generally invariant between 26.0\% and 29.6\% across methods ($p > 0.09$, see Table~\ref{tab:fpr_fnr_method}). Instead, this model suffers from an increase in False Negative Rate (FNR). In the supervised fine-tuning regime, the FPR collapses to 17.9\%--38.0\%.

When investigating the degradation per method (see Table~\ref{tab:degradation_method} from Appendix~\ref{app:degradation_analysis}), random replacement yields the highest degradation deltas against the monolingual baseline, with drops in performance across models between $-2.7$ and $-18.8$ ($\mathrm{mean} = -9.0$), acting as an input perturbation method. This is also corroborated by the t-SNE representation of the sentence embedding space (see Figure~\ref{fig:tsne} from Appendix~\ref{app:tsne}), where the random method produces the most separable representations between the original and code-switched samples. The most resilient method remains the dependency-based insertion, which produces more natural code-switched variants (see human evaluation from Appendix~\ref{app:human_evaluation}). Usually the degradation is under 3 points ($0.0$ points for Llama 3.1 8B, $+0.2$ points for GPT-6 Luna, $-0.7$ for Gemma 4 31B, and $-2.3$ for RoGemma 2 9B). 

When investigating the target language families in Appendix~\ref{sec:error_analysis} and~\ref{app:breakdown_results} (see Tables~\ref{tab:fpr_fnr_language} and~\ref{tab:cs_results}), we note that Romance languages (i.e., French and Italian) consistently produce lower FPR than Germanic ones (i.e., English and German) across most evaluated models. This result is consistent with the morphological and syntactic similarity between Romance languages and dissimilarities with Germanic insertions. However, we note that between languages, the code-mixing intensity varies, as indicated in Table~\ref{tab:cs_aggregated_metrics} from Appendix~\ref{sec:cs_evaluation}.

\section{Discussion}

Our results suggest that code-switching introduces significant challenges for satire detection, particularly for older and smaller models. Especially in zero-shot settings, these models struggle to disentangle satirical cues from code-switched input, leading to increasing false positive rates. As detailed in Appendix~\ref{app:breakdown_results} and evidenced by embedding representation in Appendix~\ref{app:tsne}, lexical substitutions act as perturbations that fine-tuning effectively mitigates, recovering performance to match or exceed the monolingual baseline. Other failure modes include instruction-following errors, which are more prevalent in smaller models (specifically, Llama 3.1 8B and RoMistral 7B). The models either fail to produce the XML format, translate tags, use synonyms like ironic, or output corrupted labels (e.g., ``\textit{satură}'', Engl. ``\textit{get full}''). 

\section{Conclusions}

We presented CS-SeLeRoSa, the first large-scale code-mixed Romanian satire benchmark, evaluating four generation methods across four target languages. Our analysis shows that code-mixing strategies strongly impact lexical naturalness and syntactic cohesion. Furthermore, our model evaluations reveal that while older open-weights models degrade under code-switching noise, modern open models such as Gemma 4 31B rival frontier systems zero-shot and, when fine-tuned, comfortably outperform them on human-adjudicated code-switched satire.

\section*{Limitations}

Our work has several limitations.

\textbf{Dataset Quality.} We only investigate synthetic code-mixing generation without assessing for human evaluation on the entire dataset. Some methods, such as random replacement, produce low-quality outputs. Even though those samples do not reflect real-world scenarios, we believe they can still be used to analyze robustness to noisy inputs. However, we mitigate the dataset evaluation by using various metrics, as indicated in Appendix~\ref{sec:cs_evaluation}, and by conducting a human evaluation on a small subset of the data in Appendix~\ref{app:human_evaluation}. We agree that synthetic generation through the random method creates degraded samples that may look unnatural to a native speaker. We mitigated this through the carefully crafted human-validated gold set, whose goal was to reduce this kind of noise and contain samples as close as possible to ``in the wild'' speech \cite{krishnan-etal-2021-multilingual}.

Additionally, the naturalness of the data is an inherent limitation of the task, since there are very few sources for satirical news that could also contain code-switching, and reliably identifying those situations is even more challenging. To mitigate this, we based our approach on linguistic theories such as the Matrix Language Frame model \cite{10.1093/oso/9780198240594.001.0001}, which guarantees that code-switched segments are grammatically and semantically coherent. We hope that our proposed resources will foster the development of new datasets.

\textbf{Romanian as Matrix Language.} We only consider the Romanian language as the base language; however, even if we added more source languages, the number of experiments and costs would increase exponentially. Instead, we focused on four target languages with more resources than Romanian, drawn from two language families. Using Romanian as the embedded language instead of the matrix language is an interesting direction for future work, as well as evaluating other code-switching language settings.

\textbf{Small LLM Evaluation.} Our evaluation is limited to several small LLMs due to the increased number of experiments and costs. We tried to mitigate this by selecting various models pre-trained on general and Romanian-specific corpora.

\section*{Ethical Considerations}

Our work used publicly available datasets and models. The SeLeRoSa dataset \cite{smuadu2025selerosa} is released under the CC BY-NC-SA 4.0 license, with the claim that the original copyright remains with the original authors of the news articles, under the existing copyright laws. The models are either publicly available on HuggingFace or proprietary, but provided using public APIs. Our usage of these resources complies with their intended use. Upon acceptance, we will release our generated code-mixed dataset and code for academic purposes only, under the same license.

Our dataset retains the original anonymization of the SeLeRoSa dataset, which replaces named entities with generic tags. First, original news sources may reflect specific biases or target particular entities (e.g., public figures, organisations) that should not be reflected in downstream applications. This is especially relevant in satire, which often ridicules those entities \citep{burfoot-baldwin-2009-automatic}. By preserving data anonymization, models are encouraged to predict satire from linguistic features and the context rather than reflecting memorised entity-level stereotypes. Moreover, from an ethical standpoint, models must not discriminate against or harm targeted groups. For transparency, the dataset retains the original source URLs to the news, allowing future work studying model biases to retrieve the original entity contexts if needed.

\bibliography{custom,anthology-1,anthology-2}

\begin{thebibliography}{72}
\providecommand{\natexlab}[1]{#1}

\bibitem[{Abd et~al.(2026)Abd, Aggarwal, Algayres, Andreev, Bachem, Ballantyne, Brick, Carbune, Casbon, Chaturvedi, Chawla, Cotruta, Coucke, Culliton, Dadashi, Dixon, Elhawaty, Evci, Farabet, Ferret, Galgani, Girgin, Grill, Grootendorst, Guo, Hardin, He, Hernandez, Homburger, Hussenot, yeong Ji, Joulin, Kamath, Kassraie, Lacombe, Lahoti, Liu, Martins, Martins, Matejovicova, Merhej, Momchev, Mondal, Mullins, Panyam, Pathak, Perrin, Pinto, Pot, Pouget, Ram{\'{e}}, Ramos, Reid, Rim, Rivi{\`{e}}re, Roth, Rouillard, Sanseviero, Sessa, Settle, Sinopalnikov, Smoot, Stanczyk, Steiner, Stewart, Tolstikhin, Tschannen, Tsitsulin, Vieillard, Wu, Xu, Yang, Yvinec, Zhang, Zhang, Zou, Aagnes, Abdelhamed, Ad{\'{a}}mek, Agrawal, Agrawal, Alabdulmohsin, Alayrac, Alon, Amarnath, Anand, Anastasiou, Ariafar, Aubet, Axiotis, Barbero, Barral, Bendebury, Bergmann, Bileschi, Black, Blondel, Borgeaud, and Brazinskas}]{DBLP:journals/corr/abs-2607-02770}
Sherif~El Abd, Vaibhav Aggarwal, Robin Algayres, Alek Andreev, Olivier Bachem, Ian Ballantyne, Cormac Brick, Victor Carbune, Michelle Casbon, Mayank Chaturvedi, Aditya Chawla, Victor Cotruta, Alice Coucke, Phil Culliton, Robert Dadashi, Lucas Dixon, Mohamed Elhawaty, Utku Evci, Cl{\'{e}}ment Farabet, and 80 others. 2026.
\newblock \href {https://doi.org/10.48550/ARXIV.2607.02770} {Gemma 4 technical report}.
\newblock \emph{CoRR}, abs/2607.02770.

\bibitem[{Abdalla et~al.(2024)Abdalla, Joshi, Masood, and Kanhere}]{abdalla-etal-2024-comparison}
Omar~W. Abdalla, Aditya Joshi, Rahat Masood, and Salil~S. Kanhere. 2024.
\newblock \href {https://aclanthology.org/2024.alta-1.14/} {Comparison of multilingual and bilingual models for satirical news detection of {A}rabic and {E}nglish}.
\newblock In \emph{Proceedings of the 22nd Annual Workshop of the Australasian Language Technology Association}, pages 173--178, Canberra, Australia. Association for Computational Linguistics.

\bibitem[{Aggarwal et~al.(2020)Aggarwal, Wadhawan, Chaudhary, and Maurya}]{aggarwal-etal-2020-really}
Akshita Aggarwal, Anshul Wadhawan, Anshima Chaudhary, and Kavita Maurya. 2020.
\newblock \href {https://doi.org/10.18653/v1/2020.wnut-1.2} {``did you really mean what you said?'' : Sarcasm detection in {H}indi-{E}nglish code-mixed data using bilingual word embeddings}.
\newblock In \emph{Proceedings of the Sixth Workshop on Noisy User-generated Text (W-NUT 2020)}, pages 7--15, Online. Association for Computational Linguistics.

\bibitem[{Aguilar et~al.(2020)Aguilar, Kar, and Solorio}]{aguilar-etal-2020-lince}
Gustavo Aguilar, Sudipta Kar, and Thamar Solorio. 2020.
\newblock \href {https://aclanthology.org/2020.lrec-1.223/} {{L}in{CE}: A centralized benchmark for linguistic code-switching evaluation}.
\newblock In \emph{Proceedings of the Twelfth Language Resources and Evaluation Conference}, pages 1803--1813, Marseille, France. European Language Resources Association.

\bibitem[{Bancu(2013)}]{bancu2013comparative}
Ariana Bancu. 2013.
\newblock A comparative analysis of romanian-english and romanian-spanish code-switching patterns.

\bibitem[{Bansal et~al.(2020)Bansal, Garimella, Suhane, Patro, and Mukherjee}]{bansal-etal-2020-code}
Srijan Bansal, Vishal Garimella, Ayush Suhane, Jasabanta Patro, and Animesh Mukherjee. 2020.
\newblock \href {https://doi.org/10.18653/v1/2020.acl-main.96} {Code-switching patterns can be an effective route to improve performance of downstream {NLP} applications: A case study of humour, sarcasm and hate speech detection}.
\newblock In \emph{Proceedings of the 58th Annual Meeting of the Association for Computational Linguistics}, pages 1018--1023, Online. Association for Computational Linguistics.

\bibitem[{Barnett et~al.(2000)Barnett, Codó, Eppler, Forcadell, Gardner-Chloros, van Hout, Moyer, Torras, Turell, Sebba, Starren, and Wensing}]{doi:10.1177/13670069000040020101}
Ruthanna Barnett, Eva Codó, Eva Eppler, Montse Forcadell, Penelope Gardner-Chloros, Roeland van Hout, Melissa Moyer, Maria~Carme Torras, Maria~Teresa Turell, Mark Sebba, Marianne Starren, and Sietse Wensing. 2000.
\newblock \href {https://doi.org/10.1177/13670069000040020101} {The lides coding manual: A document for preparing and analyzing language interaction data version 1.1—july, 1999}.
\newblock \emph{International Journal of Bilingualism}, 4(2):131--132.

\bibitem[{Belani and Flanigan(2023)}]{belani-flanigan-2023-automatic}
Ritu Belani and Jeffrey Flanigan. 2023.
\newblock \href {https://doi.org/10.18653/v1/2023.findings-acl.469} {Automatic identification of code-switching functions in speech transcripts}.
\newblock In \emph{Findings of the Association for Computational Linguistics: ACL 2023}, pages 7438--7448, Toronto, Canada. Association for Computational Linguistics.

\bibitem[{Boia(2001)}]{boia_romania_2001}
Lucian Boia. 2001.
\newblock \emph{Romania: Borderland of Europe}.
\newblock Reaktion Books, London.

\bibitem[{Burfoot and Baldwin(2009)}]{burfoot-baldwin-2009-automatic}
Clint Burfoot and Timothy Baldwin. 2009.
\newblock \href {https://aclanthology.org/P09-2041/} {Automatic satire detection: Are you having a laugh?}
\newblock In \emph{Proceedings of the {ACL}-{IJCNLP} 2009 Conference Short Papers}, pages 161--164, Suntec, Singapore. Association for Computational Linguistics.

\bibitem[{Buzilă(2025)}]{buzila2025}
Paul Buzilă. 2025.
\newblock \href {https://doi.org/10.1163/19552629-bja10094} {Recent developments of romanian in contact with other languages in europe}.
\newblock \emph{Journal of Language Contact}, 18(2):1 -- 9.

\bibitem[{Chim et~al.(2025)Chim, Ive, and Liakata}]{chim-etal-2025-evaluating}
Jenny Chim, Julia Ive, and Maria Liakata. 2025.
\newblock \href {https://doi.org/10.1162/coli_a_00540} {Evaluating synthetic data generation from user generated text}.
\newblock \emph{Computational Linguistics}, 51(1):191--233.

\bibitem[{Conneau et~al.(2020)Conneau, Khandelwal, Goyal, Chaudhary, Wenzek, Guzm{\'a}n, Grave, Ott, Zettlemoyer, and Stoyanov}]{conneau-etal-2020-unsupervised}
Alexis Conneau, Kartikay Khandelwal, Naman Goyal, Vishrav Chaudhary, Guillaume Wenzek, Francisco Guzm{\'a}n, Edouard Grave, Myle Ott, Luke Zettlemoyer, and Veselin Stoyanov. 2020.
\newblock \href {https://doi.org/10.18653/v1/2020.acl-main.747} {Unsupervised cross-lingual representation learning at scale}.
\newblock In \emph{Proceedings of the 58th Annual Meeting of the Association for Computational Linguistics}, pages 8440--8451, Online. Association for Computational Linguistics.

\bibitem[{Das and Gamb{\"a}ck(2014)}]{das-gamback-2014-identifying}
Amitava Das and Bj{\"o}rn Gamb{\"a}ck. 2014.
\newblock \href {https://aclanthology.org/W14-5152/} {Identifying languages at the word level in code-mixed {I}ndian social media text}.
\newblock In \emph{Proceedings of the 11th International Conference on Natural Language Processing}, pages 378--387, Goa, India. NLP Association of India.

\bibitem[{Dettmers et~al.(2023)Dettmers, Pagnoni, Holtzman, and Zettlemoyer}]{dettmers2023qlora}
Tim Dettmers, Artidoro Pagnoni, Ari Holtzman, and Luke Zettlemoyer. 2023.
\newblock Qlora: efficient finetuning of quantized llms.
\newblock In \emph{Proceedings of the 37th International Conference on Neural Information Processing Systems}, NIPS '23, Red Hook, NY, USA. Curran Associates Inc.

\bibitem[{Do{\u{g}}ru{\"o}z et~al.(2021)Do{\u{g}}ru{\"o}z, Sitaram, Bullock, and Toribio}]{dogruoz-etal-2021-survey}
A.~Seza Do{\u{g}}ru{\"o}z, Sunayana Sitaram, Barbara~E. Bullock, and Almeida~Jacqueline Toribio. 2021.
\newblock \href {https://doi.org/10.18653/v1/2021.acl-long.131} {A survey of code-switching: Linguistic and social perspectives for language technologies}.
\newblock In \emph{Proceedings of the 59th Annual Meeting of the Association for Computational Linguistics and the 11th International Joint Conference on Natural Language Processing (Volume 1: Long Papers)}, pages 1654--1666, Online. Association for Computational Linguistics.

\bibitem[{Ghanem et~al.(2020)Ghanem, Karoui, Benamara, Rosso, and Moriceau}]{10.1007/978-3-030-45442-5_18}
Bilal Ghanem, Jihen Karoui, Farah Benamara, Paolo Rosso, and V\'{e}ronique Moriceau. 2020.
\newblock \href {https://doi.org/10.1007/978-3-030-45442-5_18} {Irony detection in a multilingual context}.
\newblock In \emph{Advances in Information Retrieval: 42nd European Conference on IR Research, ECIR 2020, Lisbon, Portugal, April 14–17, 2020, Proceedings, Part II}, page 141–149, Berlin, Heidelberg. Springer-Verlag.

\bibitem[{Goldwasser and Zhang(2016)}]{goldwasser-zhang-2016-understanding}
Dan Goldwasser and Xiao Zhang. 2016.
\newblock \href {https://doi.org/10.1162/tacl_a_00116} {Understanding satirical articles using common-sense}.
\newblock \emph{Transactions of the Association for Computational Linguistics}, 4:537--549.

\bibitem[{Grattafiori et~al.(2024)Grattafiori, Dubey, Jauhri, Pandey, Kadian, Al-Dahle, Letman, Mathur, Schelten, Vaughan, Yang, Fan, Goyal, Hartshorn, Yang, Mitra, Sravankumar, Korenev, Hinsvark, Rao, Zhang, Rodriguez, Gregerson, Spataru, Roziere, Biron, Tang, Chern, Caucheteux, Nayak, Bi, Marra, McConnell, Keller, Touret, Wu, Wong, Ferrer, Nikolaidis, Allonsius, Song, Pintz, Livshits, Wyatt, Esiobu, Choudhary, Mahajan, Garcia-Olano, Perino, Hupkes, Lakomkin, AlBadawy, Lobanova, Dinan, Smith, Radenovic, Guzmán, Zhang, Synnaeve, Lee, Anderson, Thattai, Nail, Mialon, Pang, Cucurell, Nguyen, Korevaar, Xu, Touvron, Zarov, Ibarra, Kloumann, Misra, Evtimov, Zhang, Copet, Lee, Geffert, Vranes, Park, Mahadeokar, Shah, van~der Linde, Billock, Hong, Lee, Fu, Chi, Huang, Liu, Wang, Yu, Bitton, Spisak, Park, Rocca, Johnstun, Saxe, Jia, Alwala, Prasad, Upasani, Plawiak, Li, Heafield, Stone, El-Arini, Iyer, Malik, Chiu, Bhalla, Lakhotia, Rantala-Yeary, van~der Maaten, Chen, Tan, Jenkins, Martin, Madaan, Malo, Blecher,
  Landzaat, de~Oliveira, Muzzi, Pasupuleti, Singh, Paluri, Kardas, Tsimpoukelli, Oldham, Rita, Pavlova, Kambadur, Lewis, Si, Singh, Hassan, Goyal, Torabi, Bashlykov, Bogoychev, Chatterji, Zhang, Duchenne, Çelebi, Alrassy, Zhang, Li, Vasic, Weng, Bhargava, Dubal, Krishnan, Koura, Xu, He, Dong, Srinivasan, Ganapathy, Calderer, Cabral, Stojnic, Raileanu, Maheswari, Girdhar, Patel, Sauvestre, Polidoro, Sumbaly, Taylor, Silva, Hou, Wang, Hosseini, Chennabasappa, Singh, Bell, Kim, Edunov, Nie, Narang, Raparthy, Shen, Wan, Bhosale, Zhang, Vandenhende, Batra, Whitman, Sootla, Collot, Gururangan, Borodinsky, Herman, Fowler, Sheasha, Georgiou, Scialom, Speckbacher, Mihaylov, Xiao, Karn, Goswami, Gupta, Ramanathan, Kerkez, Gonguet, Do, Vogeti, Albiero, Petrovic, Chu, Xiong, Fu, Meers, Martinet, Wang, Wang, Tan, Xia, Xie, Jia, Wang, Goldschlag, Gaur, Babaei, Wen, Song, Zhang, Li, Mao, Coudert, Yan, Chen, Papakipos, Singh, Srivastava, Jain, Kelsey, Shajnfeld, Gangidi, Victoria, Goldstand, Menon, Sharma, Boesenberg,
  Baevski, Feinstein, Kallet, Sangani, Teo, Yunus, Lupu, Alvarado, Caples, Gu, Ho, Poulton, Ryan, Ramchandani, Dong, Franco, Goyal, Saraf, Chowdhury, Gabriel, Bharambe, Eisenman, Yazdan, James, Maurer, Leonhardi, Huang, Loyd, Paola, Paranjape, Liu, Wu, Ni, Hancock, Wasti, Spence, Stojkovic, Gamido, Montalvo, Parker, Burton, Mejia, Liu, Wang, Kim, Zhou, Hu, Chu, Cai, Tindal, Feichtenhofer, Gao, Civin, Beaty, Kreymer, Li, Adkins, Xu, Testuggine, David, Parikh, Liskovich, Foss, Wang, Le, Holland, Dowling, Jamil, Montgomery, Presani, Hahn, Wood, Le, Brinkman, Arcaute, Dunbar, Smothers, Sun, Kreuk, Tian, Kokkinos, Ozgenel, Caggioni, Kanayet, Seide, Florez, Schwarz, Badeer, Swee, Halpern, Herman, Sizov, Guangyi, Zhang, Lakshminarayanan, Inan, Shojanazeri, Zou, Wang, Zha, Habeeb, Rudolph, Suk, Aspegren, Goldman, Zhan, Damlaj, Molybog, Tufanov, Leontiadis, Veliche, Gat, Weissman, Geboski, Kohli, Lam, Asher, Gaya, Marcus, Tang, Chan, Zhen, Reizenstein, Teboul, Zhong, Jin, Yang, Cummings, Carvill, Shepard, McPhie,
  Torres, Ginsburg, Wang, Wu, U, Saxena, Khandelwal, Zand, Matosich, Veeraraghavan, Michelena, Li, Jagadeesh, Huang, Chawla, Huang, Chen, Garg, A, Silva, Bell, Zhang, Guo, Yu, Moshkovich, Wehrstedt, Khabsa, Avalani, Bhatt, Mankus, Hasson, Lennie, Reso, Groshev, Naumov, Lathi, Keneally, Liu, Seltzer, Valko, Restrepo, Patel, Vyatskov, Samvelyan, Clark, Macey, Wang, Hermoso, Metanat, Rastegari, Bansal, Santhanam, Parks, White, Bawa, Singhal, Egebo, Usunier, Mehta, Laptev, Dong, Cheng, Chernoguz, Hart, Salpekar, Kalinli, Kent, Parekh, Saab, Balaji, Rittner, Bontrager, Roux, Dollar, Zvyagina, Ratanchandani, Yuvraj, Liang, Alao, Rodriguez, Ayub, Murthy, Nayani, Mitra, Parthasarathy, Li, Hogan, Battey, Wang, Howes, Rinott, Mehta, Siby, Bondu, Datta, Chugh, Hunt, Dhillon, Sidorov, Pan, Mahajan, Verma, Yamamoto, Ramaswamy, Lindsay, Lindsay, Feng, Lin, Zha, Patil, Shankar, Zhang, Zhang, Wang, Agarwal, Sajuyigbe, Chintala, Max, Chen, Kehoe, Satterfield, Govindaprasad, Gupta, Deng, Cho, Virk, Subramanian, Choudhury,
  Goldman, Remez, Glaser, Best, Koehler, Robinson, Li, Zhang, Matthews, Chou, Shaked, Vontimitta, Ajayi, Montanez, Mohan, Kumar, Mangla, Ionescu, Poenaru, Mihailescu, Ivanov, Li, Wang, Jiang, Bouaziz, Constable, Tang, Wu, Wang, Wu, Gao, Kleinman, Chen, Hu, Jia, Qi, Li, Zhang, Zhang, Adi, Nam, Yu, Wang, Zhao, Hao, Qian, Li, He, Rait, DeVito, Rosnbrick, Wen, Yang, Zhao, and Ma}]{grattafiori2024llama3herdmodels}
Aaron Grattafiori, Abhimanyu Dubey, Abhinav Jauhri, Abhinav Pandey, Abhishek Kadian, Ahmad Al-Dahle, Aiesha Letman, Akhil Mathur, Alan Schelten, Alex Vaughan, Amy Yang, Angela Fan, Anirudh Goyal, Anthony Hartshorn, Aobo Yang, Archi Mitra, Archie Sravankumar, Artem Korenev, Arthur Hinsvark, and 542 others. 2024.
\newblock \href {https://arxiv.org/abs/2407.21783} {The llama 3 herd of models}.
\newblock \emph{Preprint}, arXiv:2407.21783.

\bibitem[{Greavu(2022)}]{Greavu_2022}
Arina Greavu. 2022.
\newblock \href {https://doi.org/10.35824/sjrs.v5i2.23888} {Morphological and syntactical aspects of romanian/ english codeswitching}.
\newblock \emph{Swedish Journal of Romanian Studies}, 5(2):191–220.

\bibitem[{Gregorius and Okadome(2022)}]{gregorius-okadome-2022-generating}
Bryan Gregorius and Takeshi Okadome. 2022.
\newblock \href {https://aclanthology.org/2022.alta-1.12/} {Generating code-switched text from monolingual text with dependency tree}.
\newblock In \emph{Proceedings of the 20th Annual Workshop of the Australasian Language Technology Association}, pages 90--97, Adelaide, Australia. Australasian Language Technology Association.

\bibitem[{Grootendorst(2020)}]{grootendorst2020keybert}
Maarten Grootendorst. 2020.
\newblock \href {https://doi.org/10.5281/zenodo.4461265} {Keybert: Minimal keyword extraction with bert.}

\bibitem[{Grootendorst(2022)}]{DBLP:journals/corr/abs-2203-05794}
Maarten Grootendorst. 2022.
\newblock \href {https://doi.org/10.48550/ARXIV.2203.05794} {Bertopic: Neural topic modeling with a class-based {TF-IDF} procedure}.
\newblock \emph{CoRR}, abs/2203.05794.

\bibitem[{Gu et~al.(2021)Gu, Wang, Bi, Meng, Liu, Han, and Shang}]{10.1145/3447548.3467397}
Xiaotao Gu, Zihan Wang, Zhenyu Bi, Yu~Meng, Liyuan Liu, Jiawei Han, and Jingbo Shang. 2021.
\newblock \href {https://doi.org/10.1145/3447548.3467397} {Ucphrase: Unsupervised context-aware quality phrase tagging}.
\newblock In \emph{Proceedings of the 27th ACM SIGKDD Conference on Knowledge Discovery \& Data Mining}, KDD '21, page 478–486, New York, NY, USA. Association for Computing Machinery.

\bibitem[{Guzmán et~al.(2017)Guzmán, Ricard, Serigos, Bullock, and Toribio}]{guzman17_interspeech}
Gualberto Guzmán, Joseph Ricard, Jacqueline Serigos, Barbara~E. Bullock, and Almeida~Jacqueline Toribio. 2017.
\newblock \href {https://doi.org/10.21437/Interspeech.2017-1429} {Metrics for modeling code-switching across corpora}.
\newblock In \emph{Interspeech 2017}, pages 67--71.

\bibitem[{Hamed et~al.(2025)Hamed, Sabty, Abdennadher, Vu, Solorio, and Habash}]{hamed-etal-2025-survey}
Injy Hamed, Caroline Sabty, Slim Abdennadher, Ngoc~Thang Vu, Thamar Solorio, and Nizar Habash. 2025.
\newblock \href {https://aclanthology.org/2025.coling-main.307/} {A survey of code-switched {A}rabic {NLP}: Progress, challenges, and future directions}.
\newblock In \emph{Proceedings of the 31st International Conference on Computational Linguistics}, pages 4561--4585, Abu Dhabi, UAE. Association for Computational Linguistics.

\bibitem[{Hochreiter and Schmidhuber(1997)}]{DBLP:journals/neco/HochreiterS97}
Sepp Hochreiter and J{\"{u}}rgen Schmidhuber. 1997.
\newblock \href {https://doi.org/10.1162/NECO.1997.9.8.1735} {Long short-term memory}.
\newblock \emph{Neural Comput.}, 9(8):1735--1780.

\bibitem[{Jayanthi et~al.(2021)Jayanthi, Nerella, Chandu, and Black}]{jayanthi-etal-2021-codemixednlp}
Sai~Muralidhar Jayanthi, Kavya Nerella, Khyathi~Raghavi Chandu, and Alan~W Black. 2021.
\newblock \href {https://doi.org/10.18653/v1/2021.calcs-1.14} {{C}odemixed{NLP}: An extensible and open {NLP} toolkit for code-mixing}.
\newblock In \emph{Proceedings of the Fifth Workshop on Computational Approaches to Linguistic Code-Switching}, pages 113--118, Online. Association for Computational Linguistics.

\bibitem[{Kamath et~al.(2025)Kamath, Ferret, Pathak, Vieillard, Merhej, Perrin, Matejovicova, Ram{\'{e}}, Rivi{\`{e}}re, Rouillard, Mesnard, Cideron, Grill, Ramos, Yvinec, Casbon, Pot, Penchev, Liu, Visin, Kenealy, Beyer, Zhai, Tsitsulin, Busa{-}Fekete, Feng, Sachdeva, Coleman, Gao, Mustafa, Barr, Parisotto, Tian, Eyal, Cherry, Peter, Sinopalnikov, Bhupatiraju, Agarwal, Kazemi, Malkin, Kumar, Vilar, Brusilovsky, Luo, Steiner, Friesen, Sharma, Sharma, Gilady, Goedeckemeyer, Saade, Kolesnikov, Bendebury, Abdagic, Vadi, Gy{\"{o}}rgy, Pinto, Das, Bapna, Miech, Yang, Paterson, Shenoy, Chakrabarti, Piot, Wu, Shahriari, Petrini, Chen, Lan, Choquette{-}Choo, Carey, Brick, Deutsch, Eisenbud, Cattle, Cheng, Paparas, Sreepathihalli, Reid, Tran, Zelle, Noland, Huizenga, Kharitonov, Liu, Amirkhanyan, Cameron, Hashemi, Klimczak{-}Plucinska, Singh, Mehta, Lehri, Hazimeh, Ballantyne, Szpektor, Nardini, Pouget{-}Abadie, Chan, Stanton, Wieting, Lai, Orbay, Fernandez, Newlan, Ji, Singh, Black, Yu, Hui, Vodrahalli, Greff, Qiu,
  Valentine, Coelho, Ritter, Hoffman, Watson, Chaturvedi, Moynihan, Ma, Babar, Noy, Byrd, Roy, Momchev, Chauhan, Bunyan, Botarda, Caron, Rubenstein, Culliton, Schmid, Sessa, Xu, Stanczyk, Tafti, Shivanna, Wu, Pan, Rokni, Willoughby, Vallu, Mullins, Jerome, Smoot, Girgin, Iqbal, Reddy, Sheth, der, Bhatnagar, Panyam, Eiger, Zhang, Liu, Yacovone, Liechty, Kalra, Evci, Misra, Roseberry, Feinberg, Kolesnikov, Han, Kwon, Chen, Chow, Zhu, Wei, Egyed, Cotruta, Giang, Kirk, Rao, Lo, Moreira, Martins, Sanseviero, Gonzalez, Gleicher, Warkentin, Mirrokni, Senter, Collins, Barral, Ghahramani, Hadsell, Matias, Sculley, Petrov, Fiedel, Shazeer, Vinyals, Dean, Hassabis, Kavukcuoglu, Farabet, Buchatskaya, Alayrac, Anil, Lepikhin, Borgeaud, Bachem, Joulin, Andreev, Hardin, Dadashi, and Hussenot}]{DBLP:journals/corr/abs-2503-19786}
Aishwarya Kamath, Johan Ferret, Shreya Pathak, Nino Vieillard, Ramona Merhej, Sarah Perrin, Tatiana Matejovicova, Alexandre Ram{\'{e}}, Morgane Rivi{\`{e}}re, Louis Rouillard, Thomas Mesnard, Geoffrey Cideron, Jean{-}~Bastien Grill, Sabela Ramos, Edouard Yvinec, Michelle Casbon, Etienne Pot, Ivo Penchev, Ga{\"{e}}l Liu, and 191 others. 2025.
\newblock \href {https://doi.org/10.48550/ARXIV.2503.19786} {Gemma 3 technical report}.
\newblock \emph{CoRR}, abs/2503.19786.

\bibitem[{Kingma and Ba(2015)}]{KingmaB14}
Diederik~P. Kingma and Jimmy Ba. 2015.
\newblock \href {http://arxiv.org/abs/1412.6980} {Adam: {A} method for stochastic optimization}.
\newblock In \emph{3rd International Conference on Learning Representations, {ICLR } 2015, San Diego, CA, USA, May 7-9, 2015, Conference Track Proceedings}.

\bibitem[{Krishnan et~al.(2021)Krishnan, Anastasopoulos, Purohit, and Rangwala}]{krishnan-etal-2021-multilingual}
Jitin Krishnan, Antonios Anastasopoulos, Hemant Purohit, and Huzefa Rangwala. 2021.
\newblock \href {https://doi.org/10.18653/v1/2021.mrl-1.18} {Multilingual code-switching for zero-shot cross-lingual intent prediction and slot filling}.
\newblock In \emph{Proceedings of the 1st Workshop on Multilingual Representation Learning}, pages 211--223, Punta Cana, Dominican Republic. Association for Computational Linguistics.

\bibitem[{Lee et~al.(2024)Lee, Jung, and Hwang}]{lee-etal-2024-commit}
Jaeseong Lee, YeonJoon Jung, and Seung-won Hwang. 2024.
\newblock \href {https://doi.org/10.18653/v1/2024.findings-naacl.198} {{COMMIT}: Code-mixing {E}nglish-centric large language model for multilingual instruction tuning}.
\newblock In \emph{Findings of the Association for Computational Linguistics: NAACL 2024}, pages 3130--3137, Mexico City, Mexico. Association for Computational Linguistics.

\bibitem[{Levi et~al.(2019)Levi, Hosseini, Diab, and Broniatowski}]{levi-etal-2019-identifying}
Or~Levi, Pedram Hosseini, Mona Diab, and David Broniatowski. 2019.
\newblock \href {https://doi.org/10.18653/v1/D19-5004} {Identifying nuances in fake news vs. satire: Using semantic and linguistic cues}.
\newblock In \emph{Proceedings of the Second Workshop on Natural Language Processing for Internet Freedom: Censorship, Disinformation, and Propaganda}, pages 31--35, Hong Kong, China. Association for Computational Linguistics.

\bibitem[{Li et~al.(2020)Li, Levi, Hosseini, and Broniatowski}]{li-etal-2020-multi-modal}
Lily Li, Or~Levi, Pedram Hosseini, and David Broniatowski. 2020.
\newblock \href {https://aclanthology.org/2020.nlp4if-1.4/} {A multi-modal method for satire detection using textual and visual cues}.
\newblock In \emph{Proceedings of the 3rd NLP4IF Workshop on NLP for Internet Freedom: Censorship, Disinformation, and Propaganda}, pages 33--38, Barcelona, Spain (Online). International Committee on Computational Linguistics (ICCL).

\bibitem[{Masala et~al.(2024)Masala, Ilie-Ablachim, Dima, Corlatescu, Zavelca, Olaru, Terian, Terian, Leordeanu, Velicu, Popescu, Dascalu, and Rebedea}]{masala-etal-2024-vorbesti}
Mihai Masala, Denis Ilie-Ablachim, Alexandru Dima, Dragos~Georgian Corlatescu, Miruna-Andreea Zavelca, Ovio Olaru, Simina-Maria Terian, Andrei Terian, Marius Leordeanu, Horia Velicu, Marius Popescu, Mihai Dascalu, and Traian Rebedea. 2024.
\newblock \href {https://doi.org/10.18653/v1/2024.findings-emnlp.681} {``vorbești rom{\^a}nește?'' a recipe to train powerful {R}omanian {LLM}s with {E}nglish instructions}.
\newblock In \emph{Findings of the Association for Computational Linguistics: EMNLP 2024}, pages 11632--11647, Miami, Florida, USA. Association for Computational Linguistics.

\bibitem[{Mazumder et~al.(2025)Mazumder, Kumar, and Patro}]{mazumder-etal-2025-revealing}
Debajyoti Mazumder, Aakash Kumar, and Jasabanta Patro. 2025.
\newblock \href {https://doi.org/10.18653/v1/2025.findings-emnlp.1308} {Revealing the impact of synthetic native samples and multi-tasking strategies in {H}indi-{E}nglish code-mixed humour and sarcasm detection}.
\newblock In \emph{Findings of the Association for Computational Linguistics: EMNLP 2025}, pages 24077--24107, Suzhou, China. Association for Computational Linguistics.

\bibitem[{McHardy et~al.(2019)McHardy, Adel, and Klinger}]{mchardy-etal-2019-adversarial}
Robert McHardy, Heike Adel, and Roman Klinger. 2019.
\newblock \href {https://doi.org/10.18653/v1/N19-1069} {Adversarial training for satire detection: Controlling for confounding variables}.
\newblock In \emph{Proceedings of the 2019 Conference of the North {A}merican Chapter of the Association for Computational Linguistics: Human Language Technologies, Volume 1 (Long and Short Papers)}, pages 660--665, Minneapolis, Minnesota. Association for Computational Linguistics.

\bibitem[{McInnes et~al.(2018)McInnes, Healy, Saul, and Großberger}]{mcinnes2018umap}
Leland McInnes, John Healy, Nathaniel Saul, and Lukas Großberger. 2018.
\newblock \href {https://doi.org/10.21105/joss.00861} {Umap: Uniform manifold approximation and projection}.
\newblock \emph{Journal of Open Source Software}, 3(29):861.

\bibitem[{Myers-Scotton(1993)}]{10.1093/oso/9780198240594.001.0001}
Carol Myers-Scotton. 1993.
\newblock \href {https://doi.org/10.1093/oso/9780198240594.001.0001} {\emph{Duelling Languages: Grammatical Structure in Codeswitching}}.
\newblock Oxford University Press.

\bibitem[{Nandy et~al.(2024)Nandy, Agarwal, Patwa, Das, Bansal, Raj, Goyal, and Ganguly}]{nandy-etal-2024-yesbut}
Abhilash Nandy, Yash Agarwal, Ashish Patwa, Millon~Madhur Das, Aman Bansal, Ankit Raj, Pawan Goyal, and Niloy Ganguly. 2024.
\newblock \href {https://doi.org/10.18653/v1/2024.emnlp-main.937} {***{Y}es{B}ut***: A high-quality annotated multimodal dataset for evaluating satire comprehension capability of vision-language models}.
\newblock In \emph{Proceedings of the 2024 Conference on Empirical Methods in Natural Language Processing}, pages 16878--16895, Miami, Florida, USA. Association for Computational Linguistics.

\bibitem[{{NLLB Team} et~al.(2024){NLLB Team}, Costa-juss{\`a}, Cross, {\c {C}}elebi, Elbayad, Heafield, Heffernan, Kalbassi, Lam, Licht, Maillard, Sun, Wang, Wenzek, Youngblood, Akula, Barrault, Gonzalez, Hansanti, Hoffman, Jarrett, Sadagopan, Rowe, Spruit, Tran, Andrews, Ayan, Bhosale, Edunov, Fan, Gao, Goswami, Guzm{\'a}n, Koehn, Mourachko, Ropers, Saleem, Schwenk, and Wang}]{nllb-24}
{NLLB Team}, Marta~R. Costa-juss{\`a}, James Cross, Onur {\c {C}}elebi, Maha Elbayad, Kenneth Heafield, Kevin Heffernan, Elahe Kalbassi, Janice Lam, Daniel Licht, Jean Maillard, Anna Sun, Skyler Wang, Guillaume Wenzek, Al~Youngblood, Bapi Akula, Loic Barrault, Gabriel~Mejia Gonzalez, Prangthip Hansanti, and 20 others. 2024.
\newblock \href {https://doi.org/10.1038/s41586-024-07335-x} {Scaling neural machine translation to 200 languages}.
\newblock \emph{Nature}, 630(8018):841--846.

\bibitem[{{OECD}(2019)}]{oecd_talent_2019}
{OECD}. 2019.
\newblock \href {https://doi.org/10.1787/bac53150-en} {\emph{Talent Abroad: A Review of Romanian Emigrants}}.
\newblock Talent Abroad. OECD Publishing, Paris.

\bibitem[{Olaleye et~al.(2025)Olaleye, Oncevay, Sibue, Zondi, Terblanche, Mapikitla, Lastrucci, Smiley, and Marivate}]{olaleye-etal-2025-afrocs}
Kayode Olaleye, Arturo Oncevay, Mathieu Sibue, Nombuyiselo Zondi, Michelle Terblanche, Sibongile Mapikitla, Richard Lastrucci, Charese Smiley, and Vukosi Marivate. 2025.
\newblock \href {https://doi.org/10.18653/v1/2025.acl-long.1601} {{A}fro{CS}-xs: Creating a compact, high-quality, human-validated code-switched dataset for {A}frican languages}.
\newblock In \emph{Proceedings of the 63rd Annual Meeting of the Association for Computational Linguistics (Volume 1: Long Papers)}, pages 33391--33410, Vienna, Austria. Association for Computational Linguistics.

\bibitem[{{OpenAI}(2026{\natexlab{a}})}]{openai2026gpt6sol}
{OpenAI}. 2026{\natexlab{a}}.
\newblock Addendum to {GPT-6} astra system card: {GPT-6.1} sol.
\newblock \url{https://deploymentsafety.openai.com/gpt-6-1-sol}.
\newblock Accessed: 2026-10-06.

\bibitem[{{OpenAI}(2026{\natexlab{b}})}]{openai2026gpt6astra}
{OpenAI}. 2026{\natexlab{b}}.
\newblock {GPT-6} astra system card.
\newblock \url{ https://deploymentsafety.openai.com/gpt-6-astra/gpt-6-astra.pdf }.
\newblock Accessed: 2026-10-06.

\bibitem[{OpenAI et~al.(2024)OpenAI, Achiam, Adler, Agarwal, Ahmad, Akkaya, Aleman, Almeida, Altenschmidt, Altman, Anadkat, Avila, Babuschkin, Balaji, Balcom, Baltescu, Bao, Bavarian, Belgum, Bello, Berdine, Bernadett-Shapiro, Berner, Bogdonoff, Boiko, Boyd, Brakman, Brockman, Brooks, Brundage, Button, Cai, Campbell, Cann, Carey, Carlson, Carmichael, Chan, Chang, Chantzis, Chen, Chen, Chen, Chen, Chen, Chess, Cho, Chu, Chung, Cummings, Currier, Dai, Decareaux, Degry, Deutsch, Deville, Dhar, Dohan, Dowling, Dunning, Ecoffet, Eleti, Eloundou, Farhi, Fedus, Felix, Fishman, Forte, Fulford, Gao, Georges, Gibson, Goel, Gogineni, Goh, Gontijo-Lopes, Gordon, Grafstein, Gray, Greene, Gross, Gu, Guo, Hallacy, Han, Harris, He, Heaton, Heidecke, Hesse, Hickey, Hickey, Hoeschele, Houghton, Hsu, Hu, Hu, Huizinga, Jain, Jain, Jang, Jiang, Jiang, Jin, Jin, Jomoto, Jonn, Jun, Kaftan, Łukasz Kaiser, Kamali, Kanitscheider, Keskar, Khan, Kilpatrick, Kim, Kim, Kim, Kirchner, Kiros, Knight, Kokotajlo, Łukasz Kondraciuk,
  Kondrich, Konstantinidis, Kosic, Krueger, Kuo, Lampe, Lan, Lee, Leike, Leung, Levy, Li, Lim, Lin, Lin, Litwin, Lopez, Lowe, Lue, Makanju, Malfacini, Manning, Markov, Markovski, Martin, Mayer, Mayne, McGrew, McKinney, McLeavey, McMillan, McNeil, Medina, Mehta, Menick, Metz, Mishchenko, Mishkin, Monaco, Morikawa, Mossing, Mu, Murati, Murk, Mély, Nair, Nakano, Nayak, Neelakantan, Ngo, Noh, Ouyang, O'Keefe, Pachocki, Paino, Palermo, Pantuliano, Parascandolo, Parish, Parparita, Passos, Pavlov, Peng, Perelman, de~Avila Belbute~Peres, Petrov, de~Oliveira~Pinto, Michael, Pokorny, Pokrass, Pong, Powell, Power, Power, Proehl, Puri, Radford, Rae, Ramesh, Raymond, Real, Rimbach, Ross, Rotsted, Roussez, Ryder, Saltarelli, Sanders, Santurkar, Sastry, Schmidt, Schnurr, Schulman, Selsam, Sheppard, Sherbakov, Shieh, Shoker, Shyam, Sidor, Sigler, Simens, Sitkin, Slama, Sohl, Sokolowsky, Song, Staudacher, Such, Summers, Sutskever, Tang, Tezak, Thompson, Tillet, Tootoonchian, Tseng, Tuggle, Turley, Tworek, Uribe, Vallone,
  Vijayvergiya, Voss, Wainwright, Wang, Wang, Wang, Ward, Wei, Weinmann, Welihinda, Welinder, Weng, Weng, Wiethoff, Willner, Winter, Wolrich, Wong, Workman, Wu, Wu, Wu, Xiao, Xu, Yoo, Yu, Yuan, Zaremba, Zellers, Zhang, Zhang, Zhao, Zheng, Zhuang, Zhuk, and Zoph}]{openai2024gpt4technicalreport}
OpenAI, Josh Achiam, Steven Adler, Sandhini Agarwal, Lama Ahmad, Ilge Akkaya, Florencia~Leoni Aleman, Diogo Almeida, Janko Altenschmidt, Sam Altman, Shyamal Anadkat, Red Avila, Igor Babuschkin, Suchir Balaji, Valerie Balcom, Paul Baltescu, Haiming Bao, Mohammad Bavarian, Jeff Belgum, and 262 others. 2024.
\newblock \href {https://arxiv.org/abs/2303.08774} {Gpt-4 technical report}.
\newblock \emph{Preprint}, arXiv:2303.08774.

\bibitem[{Pant and Dadu(2020)}]{pant-dadu-2020-towards}
Kartikey Pant and Tanvi Dadu. 2020.
\newblock \href {https://doi.org/10.18653/v1/2020.aacl-srw.6} {Towards code-switched classification exploiting constituent language resources}.
\newblock In \emph{Proceedings of the 1st Conference of the Asia-Pacific Chapter of the Association for Computational Linguistics and the 10th International Joint Conference on Natural Language Processing: Student Research Workshop}, pages 37--43, Suzhou, China. Association for Computational Linguistics.

\bibitem[{Papineni et~al.(2002)Papineni, Roukos, Ward, and Zhu}]{papineni-etal-2002-bleu}
Kishore Papineni, Salim Roukos, Todd Ward, and Wei-Jing Zhu. 2002.
\newblock \href {https://doi.org/10.3115/1073083.1073135} {{B}leu: a method for automatic evaluation of machine translation}.
\newblock In \emph{Proceedings of the 40th Annual Meeting of the Association for Computational Linguistics}, pages 311--318, Philadelphia, Pennsylvania, USA. Association for Computational Linguistics.

\bibitem[{Popovi{\'c}(2015)}]{popovic-2015-chrf}
Maja Popovi{\'c}. 2015.
\newblock \href {https://doi.org/10.18653/v1/W15-3049} {chr{F}: character n-gram {F}-score for automatic {MT} evaluation}.
\newblock In \emph{Proceedings of the Tenth Workshop on Statistical Machine Translation}, pages 392--395, Lisbon, Portugal. Association for Computational Linguistics.

\bibitem[{Pratapa et~al.(2018)Pratapa, Bhat, Choudhury, Sitaram, Dandapat, and Bali}]{pratapa-etal-2018-language}
Adithya Pratapa, Gayatri Bhat, Monojit Choudhury, Sunayana Sitaram, Sandipan Dandapat, and Kalika Bali. 2018.
\newblock \href {https://doi.org/10.18653/v1/P18-1143} {Language modeling for code-mixing: The role of linguistic theory based synthetic data}.
\newblock In \emph{Proceedings of the 56th Annual Meeting of the Association for Computational Linguistics (Volume 1: Long Papers)}, pages 1543--1553, Melbourne, Australia. Association for Computational Linguistics.

\bibitem[{Rei et~al.(2020)Rei, Stewart, Farinha, and Lavie}]{rei-etal-2020-comet}
Ricardo Rei, Craig Stewart, Ana~C Farinha, and Alon Lavie. 2020.
\newblock \href {https://doi.org/10.18653/v1/2020.emnlp-main.213} {{COMET}: A neural framework for {MT} evaluation}.
\newblock In \emph{Proceedings of the 2020 Conference on Empirical Methods in Natural Language Processing (EMNLP)}, pages 2685--2702, Online. Association for Computational Linguistics.

\bibitem[{Reimers and Gurevych(2019)}]{reimers-2019-sentence-bert}
Nils Reimers and Iryna Gurevych. 2019.
\newblock \href {http://arxiv.org/abs/1908.10084} {Sentence-bert: Sentence embeddings using siamese bert-networks}.
\newblock In \emph{Proceedings of the 2019 Conference on Empirical Methods in Natural Language Processing}. Association for Computational Linguistics.

\bibitem[{Rogoz et~al.(2021)Rogoz, G{\u{a}}man, and Ionescu}]{rogoz-etal-2021-saroco}
Ana-Cristina Rogoz, Mihaela G{\u{a}}man, and Radu~Tudor Ionescu. 2021.
\newblock \href {https://doi.org/10.18653/v1/2021.acl-short.136} {{S}a{R}o{C}o: Detecting satire in a novel {R}omanian corpus of news articles}.
\newblock In \emph{Proceedings of the 59th Annual Meeting of the Association for Computational Linguistics and the 11th International Joint Conference on Natural Language Processing (Volume 2: Short Papers)}, pages 1073--1079, Online. Association for Computational Linguistics.

\bibitem[{Rose et~al.(2010)Rose, Engel, Cramer, and Cowley}]{rose2010automatic}
Stuart Rose, Dave Engel, Nick Cramer, and Wendy Cowley. 2010.
\newblock Automatic keyword extraction from individual documents.
\newblock \emph{Text mining: applications and theory}, pages 1--20.

\bibitem[{Salaam et~al.(2022)Salaam, Dernoncourt, Bui, Rawat, and Yoon}]{salaam-etal-2022-offensive}
Cesa Salaam, Franck Dernoncourt, Trung Bui, Danda Rawat, and Seunghyun Yoon. 2022.
\newblock \href {https://aclanthology.org/2022.coling-1.575/} {Offensive content detection via synthetic code-switched text}.
\newblock In \emph{Proceedings of the 29th International Conference on Computational Linguistics}, pages 6617--6624, Gyeongju, Republic of Korea. International Committee on Computational Linguistics.

\bibitem[{Shah and Maurya(2021)}]{shah-maurya-2021-effective}
Aditya Shah and Chandresh Maurya. 2021.
\newblock \href {https://aclanthology.org/2021.icon-main.32/} {How effective is incongruity? implications for code-mixed sarcasm detection}.
\newblock In \emph{Proceedings of the 18th International Conference on Natural Language Processing (ICON)}, pages 271--276, National Institute of Technology Silchar, Silchar, India. NLP Association of India (NLPAI).

\bibitem[{Sheth et~al.(2025)Sheth, Beniwal, and Singh}]{sheth-etal-2025-comi}
Rajvee Sheth, Himanshu Beniwal, and Mayank Singh. 2025.
\newblock \href {https://doi.org/10.18653/v1/2025.findings-emnlp.422} {{COMI}-{LINGUA}: Expert annotated large-scale dataset for multitask {NLP} in {H}indi-{E}nglish code-mixing}.
\newblock In \emph{Findings of the Association for Computational Linguistics: EMNLP 2025}, pages 7973--7992, Suzhou, China. Association for Computational Linguistics.

\bibitem[{Singh et~al.(2026)Singh, Fry, Perelman, Tart, Ganesh, El-Kishky, McLaughlin, Low, Ostrow, Ananthram, Nathan, Luo, Helyar, Madry, Efremov, Spyra, Baker-Whitcomb, Beutel, Karpenko, Makelov, Neitz, Wei, Barr, Kirchmeyer, Ivanov, Christakis, Gillespie, Tam, Bennett, Wan, Huang, Sandjideh, Yang, Kumar, Saraiva, Vallone, Gheorghe, Garcia, Braunstein, Liu, Schmidt, Mereskin, Mishchenko, Applebaum, Rogerson, Rajan, Wei, Kotha, Srivastava, Agrawal, Vijayvergiya, Tyra, Nair, Nayak, Eggers, Ji, Hoover, Chen, Chen, Barak, Minaiev, Hao, Baker, Lightcap, McKinzie, Wang, Quinn, Fioca, Hsu, Yang, Yu, Zhang, Brenner, Zetino, Raymond, Lugaresi, Paz, Hudson, Whitney, Li, Chen, Cole, Voss, Ding, Shen, Huang, Colby, Hallacy, Koch, Lu, Kaplan, Kim, Minott-Henriques, Frey, Yu, Czarnecki, Reid, Wei, Decareaux, Scheau, Zhang, Forbes, Tang, Goldberg, Roberts, Palmie, Kappler, Levine, Wright, Leo, Lin, Robinson, Grabb, Chen, Lim, Salama, Bhattacharjee, Tsipras, Li, Yu, Strouse, Williams, Hunn, Bayes, Arbus, Akyurek, Le,
  Widmann, Yani, Proehl, Sert, Cheung, Schwartz, Han, Jiang, Mitchell, Sigler, Wallace, Ritter, Kavanaugh, Mays, Nikishin, Li, Such, de~Avila Belbute~Peres, Raso, Bekerman, Tsimpourlas, Chantzis, Song, Zhang, Raila, McGrath, Briggs, Yang, Parascandolo, Chabot, Kim, Zhao, Valiant, Leclerc, Salman, Wang, Sheng, Jiang, Wang, Jin, Sikchi, Schmidt, Aspegren, Chen, Qiu, Lightman, Covert, Kivlichan, Silber, Sohl, Hammoud, Clavera, Lan, Akkaya, Kostrikov, Kofman, Etinger, Singal, Hehir, Huh, Pan, Wilczynski, Pachocki, Lee, Quinn, Kiros, Kalra, Samaroo, Wang, Wolfe, Chen, Wang, Harb, Han, Wang, Zhao, Chen, Yang, Tworek, Chand, Landon, Liang, Lin, Liu, Wang, Tang, Yin, Jang, Morris, Flynn, Ferstad, Heidecke, Fishbein, Hallman, Grant, Chien, Gordon, Park, Liss, Kraaijeveld, Guay, Mo, Lawson, McGrath, Vendrow, Jiao, Lee, Steele, Wang, Mao, Chen, Hayashi, Xiao, Salahi, Wu, Sekhri, Sharma, Singhal, Li, Nguyen, Gu-Lemberg, King, Liu, Stone, Yu, Ying, Georgiev, Lim, Tirumala, Miller, Ahmad, Lv, Clare, Fauconnet, Itow, Yang,
  Romaniuk, Anise, Byron, Pathak, Maksin, Lo, Ho, Jing, Wu, Xiong, Mamitsuka, Yang, McCallum, Held, Bourgeois, Engstrom, Kuhn, Feuvrier, Zhang, Switzer, Kondraciuk, Kaiser, Joglekar, Singh, Shah, Stratta, Williams, Chen, Sun, Cayton, Li, Zhang, Aljubeh, Nichols, Haines, Schwarzer, Gupta, Shah, Guan, Huang, Dong, Wang, Glaese, Carroll, Lampe, Malek, Sharman, Zhang, Wang, Pokrass, Florian, Pavlov, Wang, Chen, Wang, Feng, Bavarian, Lin, Abdool, Rohaninejad, Soto, Staudacher, LaFontaine, Marwell, Liu, Preston, Turley, Ansman, Blades, Pancha, Mikhaylin, Felix, Handa, Rai, Keskar, Brown, Nachum, Boiko, Murk, Watkins, Gleeson, Mishkin, Lesiewicz, Baltescu, Belov, Zhokhov, Pronin, Guo, Thacker, Liu, Yuan, Liu, Dias, Puckett, Arora, Mullapudi, Gaon, Miyara, Song, Aggarwal, Marsan, Yemiru, Xiong, Kshirsagar, Nuttall, Tsiupa, Eldan, Wang, James, Ziv, Shu, Nigmatullin, Jain, Talaie, Altman, Arnesen, Toizer, Toyer, Miserendino, Agarwal, Yoo, Heon, Ethersmith, Grove, Taylor, Bubeck, Banesiu, Amdo, Zhao, Wu, Santurkar,
  Zhao, Chaudhuri, Krishnaswamy, Shuaiqi, Xia, Cheng, Anadkat, Fishman, Tobin, Fu, Jain, Mei, Egoian, Kim, Golden, Mah, Lin, Imm, Sharpe, Yadlowsky, Choudhry, Eum, Sanjeev, Khan, Stramer, Wang, Xin, Gogineni, Christianson, Sanders, Patwardhan, Degry, Shadwell, Fu, Gao, Garipov, Sriskandarajah, Sherbakov, Korbak, Kaftan, Hiratsuka, Wang, Song, Zhao, Peterson, Kharitonov, Chernova, Kosaraju, Kuo, Pong, Verma, Petrov, Jiang, Zhang, Zhou, Xie, Zhan, McCabe, DePue, Ellsworth, Bain, Thompson, Chen, Qi, Xiang, Shi, Dubois, Yu, Khakbaz, Wu, Qian, Lee, Chen, Zhang, Xiong, Tian, Cha, Bai, Yang, Yuan, Li, Zhang, Yang, Jin, Jiang, Wang, Wang, Liu, Stubenvoll, Dou, Wu, and Wang}]{singh2026openaigpt5card}
Aaditya Singh, Adam Fry, Adam Perelman, Adam Tart, Adi Ganesh, Ahmed El-Kishky, Aidan McLaughlin, Aiden Low, AJ~Ostrow, Akhila Ananthram, Akshay Nathan, Alan Luo, Alec Helyar, Aleksander Madry, Aleksandr Efremov, Aleksandra Spyra, Alex Baker-Whitcomb, Alex Beutel, Alex Karpenko, and 467 others. 2026.
\newblock \href {https://arxiv.org/abs/2601.03267} {Openai gpt-5 system card}.
\newblock \emph{Preprint}, arXiv:2601.03267.

\bibitem[{Sm{\u{a}}du et~al.(2025)Sm{\u{a}}du, Iuga, Cercel, and Pop}]{smuadu2025selerosa}
R{\u{a}}zvan-Alexandru Sm{\u{a}}du, Andreea Iuga, Dumitru-Clementin Cercel, and Florin Pop. 2025.
\newblock Selerosa: Sentence-level romanian satire detection dataset.
\newblock \emph{arXiv preprint arXiv:2509.00893}.

\bibitem[{Snover et~al.(2006)Snover, Dorr, Schwartz, Micciulla, and Makhoul}]{snover-etal-2006-study}
Matthew Snover, Bonnie Dorr, Rich Schwartz, Linnea Micciulla, and John Makhoul. 2006.
\newblock \href {https://aclanthology.org/2006.amta-papers.25/} {A study of translation edit rate with targeted human annotation}.
\newblock In \emph{Proceedings of the 7th Conference of the Association for Machine Translation in the Americas: Technical Papers}, pages 223--231, Cambridge, Massachusetts, USA. Association for Machine Translation in the Americas.

\bibitem[{Srivastava and Singh(2021)}]{srivastava-singh-2021-challenges}
Vivek Srivastava and Mayank Singh. 2021.
\newblock \href {https://doi.org/10.18653/v1/2021.calcs-1.2} {Challenges and limitations with the metrics measuring the complexity of code-mixed text}.
\newblock In \emph{Proceedings of the Fifth Workshop on Computational Approaches to Linguistic Code-Switching}, pages 6--14, Online. Association for Computational Linguistics.

\bibitem[{Sterner and Teufel(2025{\natexlab{a}})}]{sterner-teufel-2025-code}
Igor Sterner and Simone Teufel. 2025{\natexlab{a}}.
\newblock \href {https://doi.org/10.18653/v1/2025.findings-acl.600} {Code-switching and syntax: A large-scale experiment}.
\newblock In \emph{Findings of the Association for Computational Linguistics: ACL 2025}, pages 11526--11533, Vienna, Austria. Association for Computational Linguistics.

\bibitem[{Sterner and Teufel(2025{\natexlab{b}})}]{sterner-teufel-2025-minimal}
Igor Sterner and Simone Teufel. 2025{\natexlab{b}}.
\newblock \href {https://doi.org/10.18653/v1/2025.acl-long.910} {Minimal pair-based evaluation of code-switching}.
\newblock In \emph{Proceedings of the 63rd Annual Meeting of the Association for Computational Linguistics (Volume 1: Long Papers)}, pages 18575--18598, Vienna, Austria. Association for Computational Linguistics.

\bibitem[{Tarunesh et~al.(2021)Tarunesh, Kumar, and Jyothi}]{tarunesh-etal-2021-machine}
Ishan Tarunesh, Syamantak Kumar, and Preethi Jyothi. 2021.
\newblock \href {https://doi.org/10.18653/v1/2021.acl-long.245} {From machine translation to code-switching: Generating high-quality code-switched text}.
\newblock In \emph{Proceedings of the 59th Annual Meeting of the Association for Computational Linguistics and the 11th International Joint Conference on Natural Language Processing (Volume 1: Long Papers)}, pages 3154--3169, Online. Association for Computational Linguistics.

\bibitem[{Wei et~al.(2022)Wei, Wang, Schuurmans, Bosma, Ichter, Xia, Chi, Le, and Zhou}]{10.5555/3600270.3602070}
Jason Wei, Xuezhi Wang, Dale Schuurmans, Maarten Bosma, Brian Ichter, Fei Xia, Ed~H. Chi, Quoc~V. Le, and Denny Zhou. 2022.
\newblock Chain-of-thought prompting elicits reasoning in large language models.
\newblock In \emph{Proceedings of the 36th International Conference on Neural Information Processing Systems}, NIPS '22, Red Hook, NY, USA. Curran Associates Inc.

\bibitem[{Wilhelm(2017)}]{wilhelm2017}
Csilla-Anna Wilhelm. 2017.
\newblock \href {https://doi.org/10.1163/19552629-01001004} {Between simplification and complexification. german, hungarian, romanian noun and adjective morphologies in contact}.
\newblock \emph{Journal of Language Contact}, 10(1):56 -- 75.

\bibitem[{Xie et~al.(2026)Xie, Liu, Bie, Chan, Song, Wang, Chen, and Chen}]{xie2026switchlingua}
Peng Xie, Xingyuan Liu, Yequan Bie, Tsz~Wai Chan, Yangqiu Song, Yang Wang, Hao Chen, and Kani Chen. 2026.
\newblock Switchlingua: The first large-scale multilingual and multi-ethnic code-switching dataset.
\newblock \emph{Advances in Neural Information Processing Systems}, 38.

\bibitem[{Yang et~al.(2017)Yang, Mukherjee, and Dragut}]{yang-etal-2017-satirical}
Fan Yang, Arjun Mukherjee, and Eduard Dragut. 2017.
\newblock \href {https://doi.org/10.18653/v1/D17-1211} {Satirical news detection and analysis using attention mechanism and linguistic features}.
\newblock In \emph{Proceedings of the 2017 Conference on Empirical Methods in Natural Language Processing}, pages 1979--1989, Copenhagen, Denmark. Association for Computational Linguistics.

\bibitem[{Yang and Chai(2025)}]{yang2025codemixbenchevaluatingcodemixingcapabilities}
Yilun Yang and Yekun Chai. 2025.
\newblock \href {https://arxiv.org/abs/2507.18791} {Codemixbench: Evaluating code-mixing capabilities of llms across 18 languages}.
\newblock \emph{Preprint}, arXiv:2507.18791.

\bibitem[{Zhang et~al.(2020)Zhang, Kishore, Wu, Weinberger, and Artzi}]{DBLP:conf/iclr/ZhangKWWA20}
Tianyi Zhang, Varsha Kishore, Felix Wu, Kilian~Q. Weinberger, and Yoav Artzi. 2020.
\newblock \href {https://openreview.net/forum?id=SkeHuCVFDr} {Bertscore: Evaluating text generation with {BERT}}.
\newblock In \emph{8th International Conference on Learning Representations, {ICLR } 2020, Addis Ababa, Ethiopia, April 26-30, 2020}. OpenReview.net.

\bibitem[{Zhang et~al.(2025)Zhang, Majumdar, and Yadav}]{zhang2025chaillmsimprovingcodemixed}
Wenbo Zhang, Aditya Majumdar, and Amulya Yadav. 2025.
\newblock \href {https://arxiv.org/abs/2411.09073} {Chai for llms: Improving code-mixed translation in large language models through reinforcement learning with ai feedback}.
\newblock \emph{Preprint}, arXiv:2411.09073.

\bibitem[{Zheng et~al.(2023)Zheng, Chiang, Sheng, Zhuang, Wu, Zhuang, Lin, Li, Li, Xing, Zhang, Gonzalez, and Stoica}]{DBLP:conf/nips/ZhengC00WZL0LXZ23}
Lianmin Zheng, Wei{-}Lin Chiang, Ying Sheng, Siyuan Zhuang, Zhanghao Wu, Yonghao Zhuang, Zi~Lin, Zhuohan Li, Dacheng Li, Eric~P. Xing, Hao Zhang, Joseph~E. Gonzalez, and Ion Stoica. 2023.
\newblock \href {http://papers.nips.cc/paper\_ files/paper/2023/hash/91f18a1287b398d378ef22505bf41832-Abstract-Datasets \_and\_Benchmarks.html} {Judging llm-as-a-judge with mt-bench and chatbot arena}.
\newblock In \emph{Advances in Neural Information Processing Systems 36: Annual Conference on Neural Information Processing Systems 2023, NeurIPS 2023, New Orleans, LA, USA, December 10 - 16, 2023}.

\end{thebibliography}

\clearpage

\appendix

\section{Motivation}
\label{app:motivation}

The motivation and ultimate goal of our work, including the dataset construction and extensive evaluation, is multifold:

\textbf{First}, it evaluates the current state of models' understanding of satire at a granular level. Satire often occurs sporadically rather than persistently throughout an entire text. Sentence-level labels capture this nuance, which is crucial for analyzing long-form texts or instances of sporadic satire.

\textbf{Second}, code-switching is a phenomenon accentuated by globalization. As presented in \S\ref{sec:introduction}, Romania has one of the largest diasporas in the European Union, particularly in the countries we investigate for code-switching. Given the growing influence of LLMs through chat-based applications in daily life, which is likely to increase further, we believe evaluating the models' performance under such conditions is sound. Mixing languages (as suggested by \citet{grattafiori2024llama3herdmodels}) could affect performance, and our results indicate this (see Table \ref{tab:cs_results} vs Table \ref{tab:baseline_results}; in zero-shot and chain-of-thought settings, baseline performance is consistently superior to code-mixed performance).

\textbf{Third}, in Romania, German communities lived in enclaves (particularly in Transylvania), although the number of speakers has declined considerably over the last century. The French influence, however, was significant at the beginning of the 19th century, as many words entered the language from French. Although this is not considered code-mixing, since some words lacked Romanian equivalents, it suggests that historical multilingualism was common, creating an environment in which foreign terms could be readily adopted \cite{boia_romania_2001}. Studies on Romanian-English bilingual speech \cite{bancu2013comparative,Greavu_2022} indicate that the structural integrity of Romanian grammar is strictly maintained as the matrix language, thereby forcing embedded foreign elements to adapt to its complex inflectional and syntactic rules.

\textbf{Fourth}, for satire itself, as described above, there are very few public sources (satirical expressions usually occur in informal settings \cite{pant-dadu-2020-towards}), making it even more challenging to find code-switching in this setting. Therefore, naturally, we developed a code-switched dataset to address this shortcoming until more reliable resources could be generated. Note that this is less of an issue with other languages, such as Hindi.

\section{Translation Evaluation}
\label{app:translation_evaluation}

To ensure the quality of machine translations in our code-switching pipelines, we evaluate the translation capabilities of large language models using a parallel benchmark. This step is important as most machine-translation evaluations are English-centric, and lack pairs such as Ro $\to$ X (where X is French, German, Italian).
Specifically, we evaluate two sizes of the Gemma 3 model (27B and 12B) \cite{DBLP:journals/corr/abs-2503-19786} and OpenAI's GPT-5 mini \cite{singh2026openaigpt5card}, a state-of-the-art proprietary model. We select the 27B model as our primary option, with the 12B model as a fallback if it performs reasonably well with faster inference. Both models were trained on over 140 languages.
To select the best translation backbone for our code-switching pipeline, we evaluate candidate models on the parallel FLORES+ benchmark\footnote{\url{https://huggingface.co/datasets/openlanguagedata/flores_plus}} \cite{nllb-24} to assess the models' performance. It contains high-quality translations for over 200 languages.
We use Romanian as the source language and select two Germanic languages (i.e., English and German) and two Romance languages (i.e., French and Italian) as target languages. The temperature was set to 0.2 for Gemma 3 models to increase determinism, and left at the default for GPT-5 mini. In addition, we set the reasoning effort to low for OpenAI's model as a reasonable trade-off between performance and cost.

\begin{table}[ht!]
\centering
\setlength{\tabcolsep}{2pt}
\small
\begin{tabular}{lcccc}
\toprule
\textbf{Model} & \textbf{BLEU} $\uparrow$ & \textbf{chrF} $\uparrow$ & \textbf{TER} $\downarrow$ & \textbf{COMET} $\uparrow$ \\
\midrule
\multicolumn{5}{c}{\textit{RO $\rightarrow$ EN}} \\
\midrule
Gemma 3 12B & 41.04 & 67.30 & 46.23 & 0.89 \\
Gemma 3 27B & \textbf{42.80} & \textbf{69.37} & \textbf{43.31} & \textbf{0.90} \\
GPT-5 mini & 40.07 & 67.60 & 46.13 & 0.89 \\
\midrule
\multicolumn{5}{c}{\textit{RO $\rightarrow$ DE}} \\
\midrule
Gemma 3 12B & 23.61 & 52.45 & 70.19 & 0.82 \\
Gemma 3 27B & \textbf{30.42} & \textbf{60.51} & \textbf{59.59} & \textbf{0.87} \\
GPT-5 mini & 26.41 & 57.72 & 64.28 & 0.85 \\
\midrule
\multicolumn{5}{c}{\textit{RO $\rightarrow$ FR}} \\
\midrule
Gemma 3 12B & 28.67 & 54.76 & 62.80 & 0.82 \\
Gemma 3 27B & \textbf{38.78} & \textbf{64.72} & \textbf{50.53} & \textbf{0.87} \\
GPT-5 mini & 33.84 & 61.31 & 55.53 & 0.85 \\
\midrule
\multicolumn{5}{c}{\textit{RO $\rightarrow$ IT}} \\
\midrule
Gemma 3 12B & 21.82 & 52.73 & 68.53 & 0.86 \\
Gemma 3 27B & \textbf{27.11} & \textbf{57.36} & \textbf{62.22} & \textbf{0.89} \\
GPT-5 mini & 23.82 & 54.94 & 66.06 & 0.87 \\
\bottomrule
\end{tabular}
\caption{Translation performance across models and language pairs RO $\rightarrow$ X (EN, DE, FR, IT) on the FLORES+ dataset.}
\label{tab:translation_results}
\end{table}

In Table~\ref{tab:translation_results}, we report the BLEU score \cite{papineni-etal-2002-bleu}, character n-gram F-score (chrF) \cite{popovic-2015-chrf}, Translation Edit Rate (TER) \cite{snover-etal-2006-study}, and COMET \cite{rei-etal-2020-comet} for each model and language pair. Gemma 3 27B outperforms the other models on all language pairs and metrics. The best scores are obtained for English and French, which are high-resource languages where more translation resources are available. Considering these results, we select Gemma 3 27B as the translation model for generating the code-mixed dataset.

\section{Code-Mixing Evaluation}
\label{sec:cs_evaluation}

\textbf{Automatic Evaluation.}
Due to the large size of the synthetic dataset, it is challenging to assess its quality manually. Therefore, we employ several automated metrics that capture diversity, semantic similarity, and readability \cite{srivastava-singh-2021-challenges,guzman17_interspeech}. Code-mixing Index (CMI) \cite{das-gamback-2014-identifying} indicates the degree of code-mixing in a text, considering the number of words from each language. The Multilingual Index (M-Index) \cite{doi:10.1177/13670069000040020101} measures the language distribution from the text, where $1$ indicates an equal distribution of languages. The Integration Index (I-Index) \cite{guzman17_interspeech} quantifies the probability of language switching in the text. The Language Entropy (LE) \cite{guzman17_interspeech} computes the number of bits required to encode the language distribution from the text. Finally, the Switch Point Average (SP-Avg) \cite{jayanthi-etal-2021-codemixednlp} measures the probability of the language being switched in the text. To compute these metrics, we use the CodeMixedNLP toolkit\footnote{\url{https://github.com/murali1996/CodemixedNLP}} \cite{jayanthi-etal-2021-codemixednlp}.
For token-level language identification, we employ a series of heuristics based on language-specific special characters. If the word cannot be identified in this way, we use the lingua toolkit\footnote{\url{https://github.com/pemistahl/lingua-py}}, which supports the languages evaluated. Table~\ref{tab:cs_aggregated_metrics} shows that using the random replacement method achieves much more frequent switching between languages, while the other methods are more balanced.

\begin{table}[th!]
\centering
\setlength{\tabcolsep}{2pt}
\resizebox{0.8\columnwidth}{!}{%
\begin{tabular}{lccccc}
\toprule
\textbf{Method} & \textbf{CMI} & \textbf{M-Index} & \textbf{I-Index} & \textbf{LE} & \textbf{SP-Avg} \\
\midrule
\multicolumn{6}{c}{\textit{RO $\rightarrow$ EN}} \\
\midrule
Random & 39.29 & 0.44 & 0.47 & 0.94 & 0.46 \\
Phrase & 15.93 & 0.18 & 0.22 & 0.55 & 0.21 \\
Dependency & 18.75 & 0.21 & 0.16 & 0.57 & 0.16 \\
LLM & 18.64 & 0.21 & 0.19 & 0.62 & 0.21 \\
\midrule
\multicolumn{6}{c}{\textit{RO $\rightarrow$ DE}} \\
\midrule
Random & 36.34 & 0.41 & 0.45 & 0.91 & 0.44 \\
Phrase & 26.66 & 0.31 & 0.31 & 0.76 & 0.30 \\
Dependency & 25.05 & 0.29 & 0.22 & 0.73 & 0.23 \\
LLM & 21.17 & 0.24 & 0.24 & 0.69 & 0.25 \\
\midrule
\multicolumn{6}{c}{\textit{RO $\rightarrow$ FR}} \\
\midrule
Random & 40.12 & 0.44 & 0.47 & 0.94 & 0.46 \\
Phrase & 20.52 & 0.24 & 0.27 & 0.65 & 0.26 \\
Dependency & 21.61 & 0.25 & 0.20 & 0.64 & 0.21 \\
LLM & 22.40 & 0.26 & 0.24 & 0.71 & 0.25 \\
\midrule
\multicolumn{6}{c}{\textit{RO $\rightarrow$ IT}} \\
\midrule
Random & 40.93 & 0.43 & 0.49 & 1.00 & 0.48 \\
Phrase & 22.48 & 0.25 & 0.30 & 0.72 & 0.29 \\
Dependency & 23.54 & 0.26 & 0.23 & 0.72 & 0.23 \\
LLM & 25.55 & 0.28 & 0.27 & 0.80 & 0.29 \\
\bottomrule
\end{tabular}}
\caption{Aggregated code-switching metrics by language and code-switching method.}
\label{tab:cs_aggregated_metrics}
\end{table}

In addition, we measure the dissimilarity between generations at the token level (i.e., derived from the Jaccard index) and the semantic level (i.e., derived from BERTScore \cite{DBLP:conf/iclr/ZhangKWWA20}). Given two sentences $S_1$ and $S_2$ provided as a list of tokens, the Jaccard similarity is computed as:

\begin{equation}
J(S_1, S_2) = \frac{|S_1 \cap S_2|}{|S_1 \cup S_2|}
\end{equation}

We define the Jaccard dissimilarity as $1 - J(S_1, S_2)$. On the other hand, BERTScore measures the semantic similarity between two texts based on the pairwise cosine similarity of their embeddings. We measure the dissimilarity as $1 - \texttt{BERTScore}(S_1, S_2)$. We employ the multilingual XLM-RoBERTa-large pretrained model \cite{conneau-etal-2020-unsupervised} to embed the sentences.
In Table~\ref{tab:cs_semantic_jaccard_scores}, we report the average dissimilarity scores (scaled to $[0, 100]$) between all pairs of generated sentences for each method and language pair. As observed before, the random replacement method generates more dissimilar sentences due to the lack of context when translating individual words. Additionally, for most language pairs (EN, FR, IT), LLM-based code-mixing produces slightly more dissimilar sentences than the two rule-based methods (phrase and dependency), likely reflecting prompt-driven lexical variety, though for German the rule-based methods yield slightly higher semantic dissimilarity.

\begin{table}[!t]
\centering
\setlength{\tabcolsep}{2pt}
\resizebox{\columnwidth}{!}{
\begin{tabular}{lcccccc}
\toprule
\multicolumn{1}{c}{\multirow{2}{*}{\textbf{Method}}} & 
\multicolumn{3}{c}{\textbf{BERTScore}} & 
\multicolumn{3}{c}{\textbf{Jaccard}} \\
& \textbf{Mean} & \textbf{Std} & \textbf{Median} & \textbf{Mean} & \textbf{Std} & \textbf{Median} \\
\midrule
\multicolumn{7}{c}{\textit{RO $\rightarrow$ EN}} \\
\midrule
Random & 8.03 & 1.91 & 7.90 & 73.28 & 7.85 & 72.77 \\
Phrase & 3.33 & 1.51 & 3.37 & 29.80 & 13.23 & 30.68 \\
Dependency & 3.31 & 0.95 & 3.22 & 37.37 & 6.53 & 37.89 \\
LLM & 3.61 & 2.05 & 3.08 & 39.06 & 19.66 & 34.41 \\
\midrule
\multicolumn{7}{c}{\textit{RO $\rightarrow$ DE}} \\
\midrule
Random & 10.24 & 1.87 & 10.18 & 73.61 & 7.62 & 73.12 \\
Phrase & 4.22 & 1.86 & 4.27 & 30.30 & 13.65 & 30.83 \\
Dependency & 4.23 & 1.07 & 4.26 & 36.67 & 7.25 & 37.46 \\
LLM & 3.49 & 1.92 & 3.03 & 28.35 & 16.80 & 24.34 \\
\midrule
\multicolumn{7}{c}{\textit{RO $\rightarrow$ FR}} \\
\midrule
Random & 8.11 & 1.92 & 7.94 & 67.73 & 9.01 & 67.68 \\
Phrase & 3.57 & 1.65 & 3.59 & 28.35 & 13.02 & 29.28 \\
Dependency & 3.66 & 1.13 & 3.67 & 34.57 & 8.05 & 35.70 \\
LLM & 3.87 & 1.97 & 3.37 & 36.83 & 16.79 & 33.88 \\
\midrule
\multicolumn{7}{c}{\textit{RO $\rightarrow$ IT}} \\
\midrule
Random & 7.86 & 1.80 & 7.75 & 68.77 & 8.91 & 68.76 \\
Phrase & 3.58 & 1.61 & 3.65 & 28.37 & 13.07 & 28.70 \\
Dependency & 3.57 & 1.01 & 3.54 & 35.11 & 7.41 & 35.86 \\
LLM & 4.04 & 1.88 & 3.61 & 38.03 & 17.72 & 34.81 \\
\bottomrule
\end{tabular}}
\caption{Semantic distance and Jaccard dissimilarity for code-switched sentences by language and method.}
\label{tab:cs_semantic_jaccard_scores}
\end{table}

\textbf{Manual Evaluation.} We select a subset of sentences from the generated dataset for human evaluation. Specifically, we randomly sample 50 non-code-mixed sentences and extract the corresponding generated code-mixed sentences for each of the four methods and four languages, resulting in a total of 715 samples. These samples were then filtered to remove instances of unnatural code-switching or significant grammatical errors, resulting in a final set of 391 high-quality code-mixed sentences. We call this subset CS-SeLeRoSa-Gold, which, due to its data selection strategy, is meant only for model evaluation.
More details on the annotation protocol and human evaluation statistics are provided in Appendix~\ref{app:human_evaluation}.

\section{Dataset Structure}
\label{app:dataset_structure}

For each sample in our dataset, we record the generation index; this allows researchers to isolate specific generation variants or evaluate deduplicated splits without repeated base sentences. More specifically, in addition to the columns from the SeLeRoSa dataset \cite{smuadu2025selerosa}, our code-mixed dataset adds the following fields for each sample:
\begin{itemize}
    \item \texttt{code\_switch\_language}: the target language for code switching (the source is by default Romanian);
    \item \texttt{k}: the generation index, $k \in \{0, 1, 2, 3\}$, where $k=0$ indicates the original sentence;
    \item \texttt{cs\_method}: the code-switching generation method (\texttt{none}, \texttt{random}, \texttt{dependency}, \texttt{llm}, \texttt{phrase}), where \texttt{none} corresponds to the original sentence.
\end{itemize}

Table~\ref{tab:cs_examples} presents concrete examples of these generated code-switched sentences under the different methods, target languages, and label classes.

\begin{table*}[ht!]
\centering
\scriptsize
\setlength{\tabcolsep}{3pt}
\resizebox{\textwidth}{!}{%
\begin{tabular}{llp{3.3cm}p{3.3cm}p{3.3cm}}
\toprule
\textbf{Label} & \textbf{Method} & \textbf{Original (RO)} & \textbf{Code-Switched} & \textbf{Translated (EN)} \\
\midrule
\multicolumn{5}{c}{\textit{RO $\rightarrow$ EN}} \\
\midrule
\multirow{2}{*}{\textbf{Satire}} 
 & \textbf{Phrase} & Ăla poate să distrugă şi şalele transplantate, şi pe ale tale.
 & Ăla poate să destroy şi şalele transplantate, şi pe ale tale.
 & That one can destroy both the transplanted loins and your own. \\
\cmidrule{2-5}
 & \textbf{LLM} & Deși sunt pe cartelă, după ce consumam net, primeam altul.
 & Deși sunt pe cartelă, after I consumam net, primeam another one.
 & Although I am on a prepaid card, after I consumed the internet, I would receive another one. \\
\midrule
\multirow{2}{*}{\textbf{Regular}} 
 & \textbf{Random} & Totul din lemn, calitate norvegiană.
 & Everything din wood, quality Norwegian.
 & Everything made of wood, Norwegian quality. \\
\cmidrule{2-5}
 & \textbf{Dependency} & Cât să mai răbdăm, că nici noi nu suntem de fier!
 & Cât să mai răbdăm, that we are not made of iron either!
 & How much more must we endure, since we are not made of iron either! \\
\midrule
\multicolumn{5}{c}{\textit{RO $\rightarrow$ DE}} \\
\midrule
\multirow{2}{*}{\textbf{Satire}} 
 & \textbf{Random} & \textless{}PERSON\textgreater{} contemplativ .
 & \textless{}PERSON\textgreater{} kontemplativ.
 & \textless{}PERSON\textgreater{} contemplative. \\
\cmidrule{2-5}
 & \textbf{Dependency} & A montat un cep și de atunci nu s-a mai despărțit de stup.
 & A montat un cep und seitdem hat er sich nicht mehr vom Bienenstock getrennt.
 & He mounted a spigot and since then he hasn't separated from the beehive. \\
\midrule
\multirow{2}{*}{\textbf{Regular}} 
 & \textbf{Phrase} & Iar \textless{}PERSON\textgreater{} ştie exact câţi bani are fiecare.
 & Iar \textless{}PERSON\textgreater{} ştie exact câţi Geld are fiecare.
 & And \textless{}PERSON\textgreater{} knows exactly how much money everyone has. \\
\cmidrule{2-5}
 & \textbf{LLM} & Nevesti-mii i s-a făcut rău pe loc, am dus-o la spital.
 & Nevesti-mii i s-a făcut rău pe loc, am dus-o ins Krankenhaus.
 & My wife became ill on the spot, I took her to the hospital. \\
\midrule
\multicolumn{5}{c}{\textit{RO $\rightarrow$ FR}} \\
\midrule
\multirow{2}{*}{\textbf{Satire}} 
 & \textbf{Random} & Pentru că blocajele sunt tot mai numeroase, \textless{}PERSON\textgreater{}, împreună cu o firmă specializată în ridicarea vehicolelor plutitoare, au început să ridice hidrobicicletele care încurcă traficul.
 & Pentru que les blocages sunt tot mai nombreux, \textless{}PERSON\textgreater{}, împreună avec un firme spécialisée în l'élévation des véhicules flottantes, ont început să lève hidrobicicletele qui embrouille le trafic.
 & Because blockages are increasingly numerous, \textless{}PERSON\textgreater{}, together with a company specialized in towing floating vehicles, started towing paddle boats that obstruct traffic. \\
\cmidrule{2-5}
 & \textbf{Dependency} & Doi \textless{}PERSON\textgreater{} beţi au spart o conductă de gaze de şist, şi un excavator a căzut în magmă.
 & Doi \textless{}PERSON\textgreater{} beţi au spart un pipeline de gaz de schiste, şi un excavator a căzut în magmă.
 & Two drunk \textless{}PERSON\textgreater{} broke a shale gas pipeline, and an excavator fell into the magma. \\
\midrule
\multirow{2}{*}{\textbf{Regular}} 
 & \textbf{Phrase} & Ce să facem şi noi, decât să dăm în cap la lume pe stradă
 & Ce să nous faisons şi noi, decât să dăm în cap la lume pe rue
 & What else should we do, other than hit people on the head on the street? \\
\cmidrule{2-5}
 & \textbf{LLM} & Nu întâmplător evenimentul este prezentat pe afişe drept Târgul care te învaţă cum să combaţi viaţa!.
 & Nu întâmplător, l'événement este prezentat pe afişe drept Târgul care te învaţă cum să combaţi la vie!.
 & It is not by chance that the event is presented on posters as the Fair that teaches you how to fight life!. \\
\midrule
\multicolumn{5}{c}{\textit{RO $\rightarrow$ IT}} \\
\midrule
\multirow{2}{*}{\textbf{Satire}} 
 & \textbf{Random} & Nemţii sunt în stare de şoc, după ce un post de televiziune a anunţat că armata statului \textless{}GPE\textgreater{} a invadat \textless{}GPE\textgreater{}.
 & I tedeschi sunt în stare di shock, dopo ce un post di televisione a annunciato că esercito dello stato \textless{}GPE\textgreater{} a invaso \textless{}GPE\textgreater{}.
 & The Germans are in a state of shock after a television station announced that the army of state \textless{}GPE\textgreater{} invaded \textless{}GPE\textgreater{}. \\
\cmidrule{2-5}
 & \textbf{Phrase} & În acelaşi timp, procesul generează temeri cum că \textless{}ORG\textgreater{} ar putea invada viaţa privată a \textless{}PERSON\textgreater{}.
 & În acelaşi timp, il processo genera timori cum că \textless{}ORG\textgreater{} ar putea invada viaţa privată a \textless{}PERSON\textgreater{}.
 & At the same time, the process generates fears that \textless{}ORG\textgreater{} could invade the private life of \textless{}PERSON\textgreater{}. \\
\midrule
\multirow{2}{*}{\textbf{Regular}} 
 & \textbf{Dependency} & Iar \textless{}PERSON\textgreater{} ştie exact câţi bani are fiecare.
 & Iar \textless{}PERSON\textgreater{} ştie  esattamente quanti soldi ha ciascuno.
 & And \textless{}PERSON\textgreater{} knows exactly how much money everyone has. \\
\cmidrule{2-5}
 & \textbf{LLM} & Nevesti-mii i s-a făcut rău pe loc, am dus-o la spital.
 & Nevesti-mii i s-a făcut rău pe loc, am dus-o in ospedale.
 & My wife became ill on the spot, I took her to the hospital. \\
\bottomrule
\end{tabular}%
}
\caption{Examples of original Romanian sentences, their synthetically generated code-switched counterparts, and their English translations from the CS-SeLeRoSa dataset. The selection is grouped by target languages (EN, DE, FR, IT), covering all four generation methods and annotation labels (Satire vs. Regular) with non-trivial examples.}
\label{tab:cs_examples}
\end{table*}

\section{Additional Experimental Setup}
\label{app:experimental_setup}

\subsection{Models}
\label{app:models}

In Table \ref{tab:checkpoints}, we present the model checkpoints used in our experiments, along with their sizes. All models used Romanian in their pre-training data. In what follows, we summarize the training data and its importance to each model's performance.

Llama 3.1 \cite{grattafiori2024llama3herdmodels} was trained on a mix of publicly available data totaling more than 15T tokens. The model was optimized for several languages, including English, German, French, and Italian; however, the training data included more languages, such as Romanian.

Gemma 3 models \cite{DBLP:journals/corr/abs-2503-19786} were trained on various data, including web documents (from over 140 languages), code, and images, totaling 14T tokens for the 27B parameters model. However, we could not find exact details about the data.

Gemma 4 models \cite{DBLP:journals/corr/abs-2607-02770} represent the next generation of Gemma models, which were trained on a large-scale multilingual dataset, following a similar pretraining setup as Gemma 3. The family introduces a more efficient 26B A4B (i.e., with 3.8B activated parameters) mixture-of-experts model alongside the 31B dense variant.

The Romanian language models \cite{masala-etal-2024-vorbesti} were instruction-fine-tuned on 2.7M instructions and conversations translated from various existing datasets. Because all base models used the same instruction-fine-tuning data, the observed performance disparities stem from differences in their pretraining strategies, data, and architectures.

Regarding the closed-source models (GPT-4.1 mini, GPT-5 mini, and GPT-6 Luna), as with previous models, the data is a mix of publicly available and proprietary sources. The exact size and nature of the datasets are unknown to the public. Additionally, on the smaller gold set, due to cost considerations, we also evaluate GPT-5, GPT-6 Astra, GPT-6 Sol, and GPT-6.1 Sol.

The training data used to generate these checkpoints influences the models' performance, as suggested in \S\ref{sec:experimental_setup}. In zero-shot settings, RoGemma 2 and RoLlama 3.1 consistently outperform their base counterparts on the gold-standard dataset. While closed-source models and newer open systems (such as Gemma 4 31B) achieve the highest scores, Romanian-pretrained models remain highly competitive against their base counterparts. This likely reflects the relative volume and composition of their pre-training data. Additionally, the Llama 3 model card \cite{grattafiori2024llama3herdmodels} notes that mixing languages could yield unexpected results, a finding we validated by observing poor performance.

\begin{table*}[t!]
\centering
\begin{tabular}{llc}
\toprule
\textbf{Model Name} & \textbf{Checkpoint} & \textbf{Size} \\
\midrule
Gemma 3 12B & \texttt{google/gemma-3-12b-it} & 12B \\
Gemma 3 27B & \texttt{google/gemma-3-27b-it} & 27B \\
Gemma 4 26B A4B & \texttt{google/gemma-4-26B-A4B-it} & 26B \\
Gemma 4 31B & \texttt{google/gemma-4-31B-it} & 31B \\
Llama 3.1 8B & \texttt{meta-llama/Llama-3.1-8B-Instruct} & 8B \\
RoGemma 2 9B & \texttt{OpenLLM-Ro/RoGemma2-9b-Instruct-DPO} & 9B \\
RoLlama 3.1 8B & \texttt{OpenLLM-Ro/RoLlama3.1-8b-Instruct-DPO} & 8B \\
RoMistral 7B & \texttt{OpenLLM-Ro/RoMistral-7b-Instruct-DPO} & 7B \\
\midrule
GPT-4.1 mini & \texttt{gpt-4.1-mini-2025-04-14} & -- \\
GPT-5 mini & \texttt{gpt-5-mini-2025-08-07} & -- \\
GPT-5 & \texttt{gpt-5-2025-08-07} & -- \\
GPT-6 Luna & \texttt{gpt-6-luna} & -- \\
GPT-6 Astra & \texttt{gpt-6-astra} & -- \\
GPT-6 Sol & \texttt{gpt-6-sol} & -- \\
GPT-6.1 Sol & \texttt{gpt-6.1-sol} & -- \\
\bottomrule
\end{tabular}
\caption{Model checkpoints used in our experiments.}
\label{tab:checkpoints}
\end{table*}

\subsection{Hyperparameters}
\label{app:hyperparameters}

\textbf{Inference.} The evaluated models (excluding the sample generation stage) use the following hyperparameters during inference. For Gemma models, we follow the official recommendations: the temperature was set to $1.0$, top-$p$ to $0.95$, and top-$k$ to $64$. For the other open-source models, the temperature was set to $0.8$ and to the default for the proprietary models. The maximum number of tokens was set to $2048$ for all models. The reasoning effort was set to low for the proprietary models in the translation setting and medium in the satirical code-switching classification setting. The number of runs per prompt was set to $3$ for the open-source models and to $1$ for the proprietary models (due to budget constraints) on the full CS-SeLeRoSa test set, and $3$ for the gold set. We used the unquantized \texttt{bfloat16} versions of the models.

\textbf{Fine-tuning.} We also include fine-tuning results of the open-source models. We use an efficient 4-bit quantization with QLoRA \cite{dettmers2023qlora} to fine-tune the open-source models. The models were fine-tuned for 3 epochs, with a linear learning-rate scheduler starting at $1\times 10^{-4}$ with a 10\% warm-up, a batch size of 32, and a context length of 1024 tokens. We used LoRA rank of 16, $\alpha=32$, and dropout of 0.05 for all attention and MLP projection layers (\texttt{up\_proj}, \texttt{down\_proj}, \texttt{gate\_proj}, \texttt{k\_proj}, \texttt{q\_proj}, \texttt{v\_proj}, and \texttt{o\_proj}). For optimization, we used Adam \cite{KingmaB14}. The remaining parameters are used by the HuggingFace Trainer by default. We did not fine-tune the proprietary models due to budget constraints.

\subsection{Computational Resources and Costs}
\label{app:hardware_resources}

For inference and fine-tuning, we used NVIDIA A100 80GB and NVIDIA H100 80GB GPUs, employing vLLM\footnote{\url{https://github.com/vllm-project/vllm}} to improve inference speed, with an OpenAI-compatible endpoint. For the proprietary models, we use the OpenAI API\footnote{\url{https://platform.openai.com/docs/models}}. The total cost of the experiments on the proprietary models is approximately 120 USD.

\section{Human Evaluation}
\label{app:human_evaluation}

\begin{figure*}[htbp]
\centering
\includegraphics[width=\textwidth]{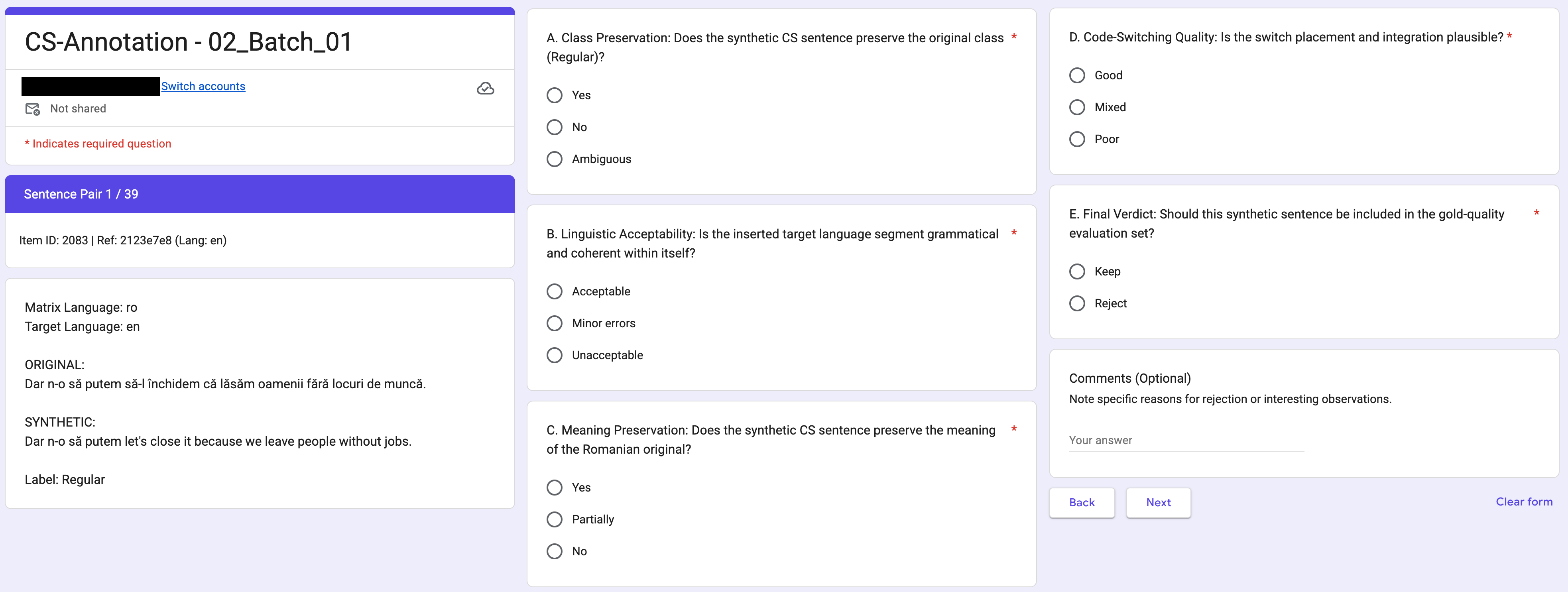}
\caption{Screenshot of an example from the Google Form used during the annotation process. It indicates the batch identifier, progress through the batch, sample identifier, matrix, and target languages, the original and synthetic sentences, and the original label. After this, there are the five mandatory questions and an optional comments field. The original screenshot was anonymized and split into three columns for presentation purposes.}
\label{fig:annotation_form_example}
\end{figure*}

The annotation process involved four annotators proficient in Romanian and knowledgeable of the target languages (i.e., English, German, French, and Italian). The annotators were provided with detailed guidelines to ensure consistency in their evaluations.

\subsection{Annotation Protocol}
\label{app:annotation_protocol}

The annotation campaign was done using Google Forms\footnote{\url{forms.google.com}}. 
Each annotator was presented with the original Romanian sentences, the target language segments used for code-switching, and the original satirical label. Inspired by \citet{chim-etal-2025-evaluating}, the annotators were provided guidelines with instructions and examples to evaluate each synthetic sentence based on the following criteria:
\begin{itemize}
    \item \textbf{Class Preservation}: Does the synthetic CS sentence preserve the original class (satire or regular)? Yes/No/Ambiguous
    \item \textbf{Linguistic Acceptability}: Is the inserted target language segment grammatical and coherent within itself? Acceptable/Minor errors/Unacceptable
    \item \textbf{Meaning Preservation}: Does the synthetic CS sentence preserve the meaning of the original Romanian sentence? Yes/Partially/No
    \item \textbf{Code-Switching Quality}: Is the switch placement and integration plausible? Good/Mixed/Poor
    \item \textbf{Final Verdict}: Should the synthetic sentence be kept or rejected? Keep/Reject
\end{itemize}

The annotators were instructed to reject any synthetic sentence that introduces hate speech, targeted insults, or violent content not present in the original. The full annotation guidelines are provided in the Supplemental Materials.

The annotation process was performed in two phases. First, the \textbf{Calibration Phase}, where annotators labeled a small set of sentences to align their understanding of the guidelines. Second, the \textbf{Batch Annotation Phase}, where the remaining workload was distributed in manageable batches. To make the process easier to leverage, we provided the data in batches of approx. 40 examples, containing code-mixed sentences whose target languages were known by the annotator. An example is presented in Figure \ref{fig:annotation_form_example}.
At the end, we checked the agreement among annotators, and in case of disagreements, they discussed to reach a consensus. A final human verification was then performed on all samples to ensure response quality and identify any potential issues.

In addition, to the best of our knowledge, we note that this is the first work to include such a rich set of categories labeled by human annotators for code-switched sentences. Most works limit themselves to fewer criteria (e.g., naturalness and acceptability \cite{sterner-teufel-2025-minimal}) or employ humans in the loop to validate and fix the samples \cite{olaleye-etal-2025-afrocs}.

Since the samples are synthetically generated, we provide this minimal set of categories to label to reduce annotation bias. The questions are designed to identify potential misinterpretations of the guidelines and to fix them in a timely manner. That is, if we found contradictions during annotation (e.g., the annotator considered the linguistic acceptability unacceptable; however, the final verdict was to keep the sample), we addressed them in the adjudication and final verification stages. The number of options per question was kept low (i.e., 2-3 categorical options).

We also note that throughout the annotation process, the human experts identified additional aspects that could be interesting to investigate. For example, the phonetic/pronunciation aspect affects how easily integrated a code-switched phrase is in the matrix sentence. Additionally, some words may be pronounced similarly despite having different meanings or parts of speech (e.g., ``joc'' in Romanian can mean the noun ``game'' or the first-person singular verb ``to play''). However, we leave this investigation as future work.

\subsection{Data Selection}
\label{app:data_selection}

To obtain a representative subset, we select sentences from the generated dataset for human evaluation. Specifically, we randomly sampled 50 non-code-mixed Romanian sentences across the test set. We extracted the corresponding generated code-mixed sentences for each of the four methods and four languages.
We deduplicated the generated samples and removed those that were not translated before sampling. 
In total, we annotated 715 samples. The base sentences contained 37 satirical (i.e., 74\%) and 13 regular (i.e., 26\%) samples, distributed among social (i.e., 78\%) and technology and science (i.e., 22\%) domains. For each language, between 150 and 200 samples were manually labeled (see also Table \ref{tab:language_distribution}).

\subsection{Annotation Statistics}

\begin{figure}[!t]
\centering
\includegraphics[width=\columnwidth,trim={1cm 0 0 0},clip]{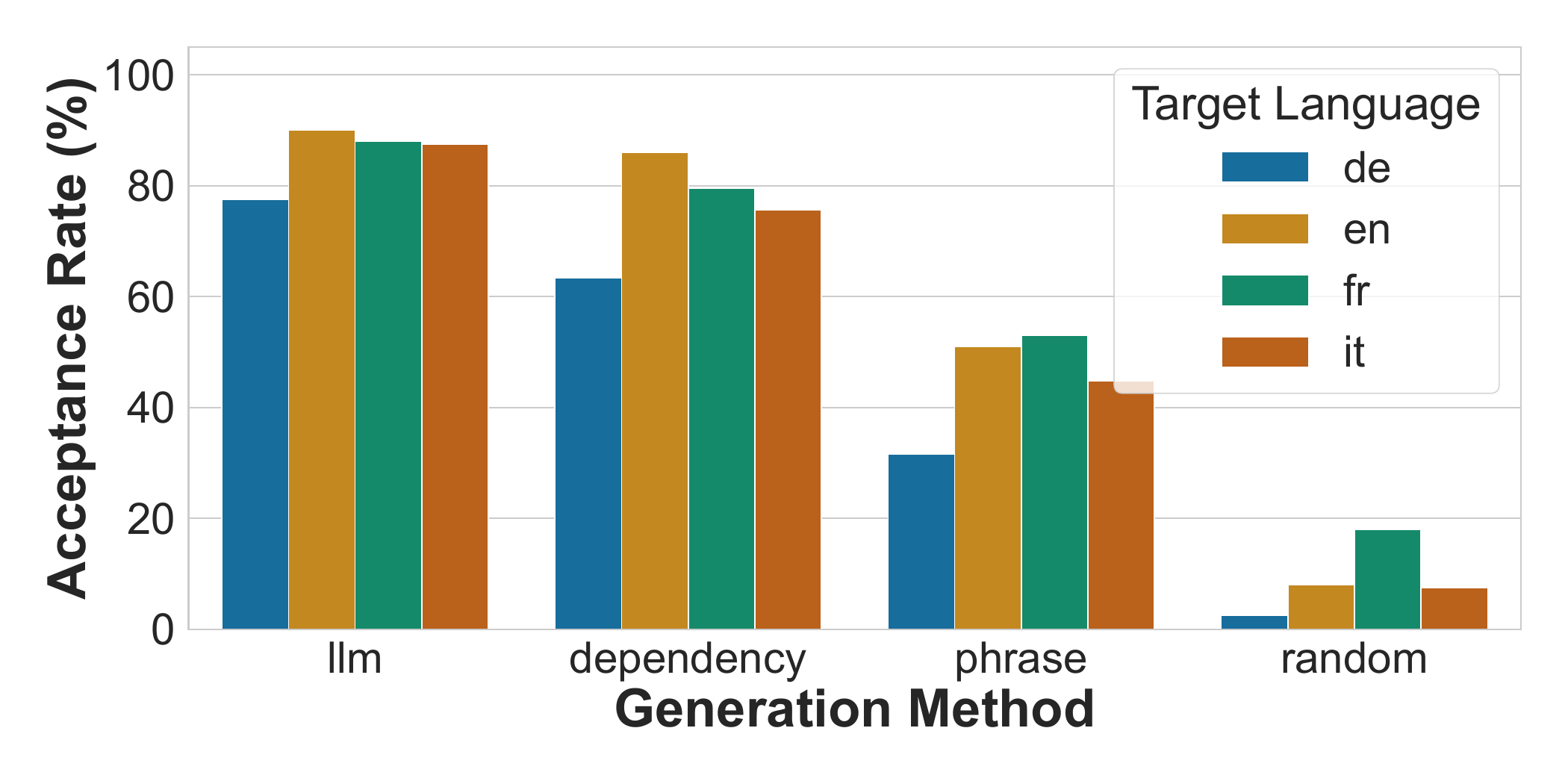}
\caption{The acceptance rate by method and target language.}
\label{fig:language_comparison}
\end{figure}

\begin{figure}[!t]
\centering
\includegraphics[width=\columnwidth,trim={1cm 0 0 0},clip]{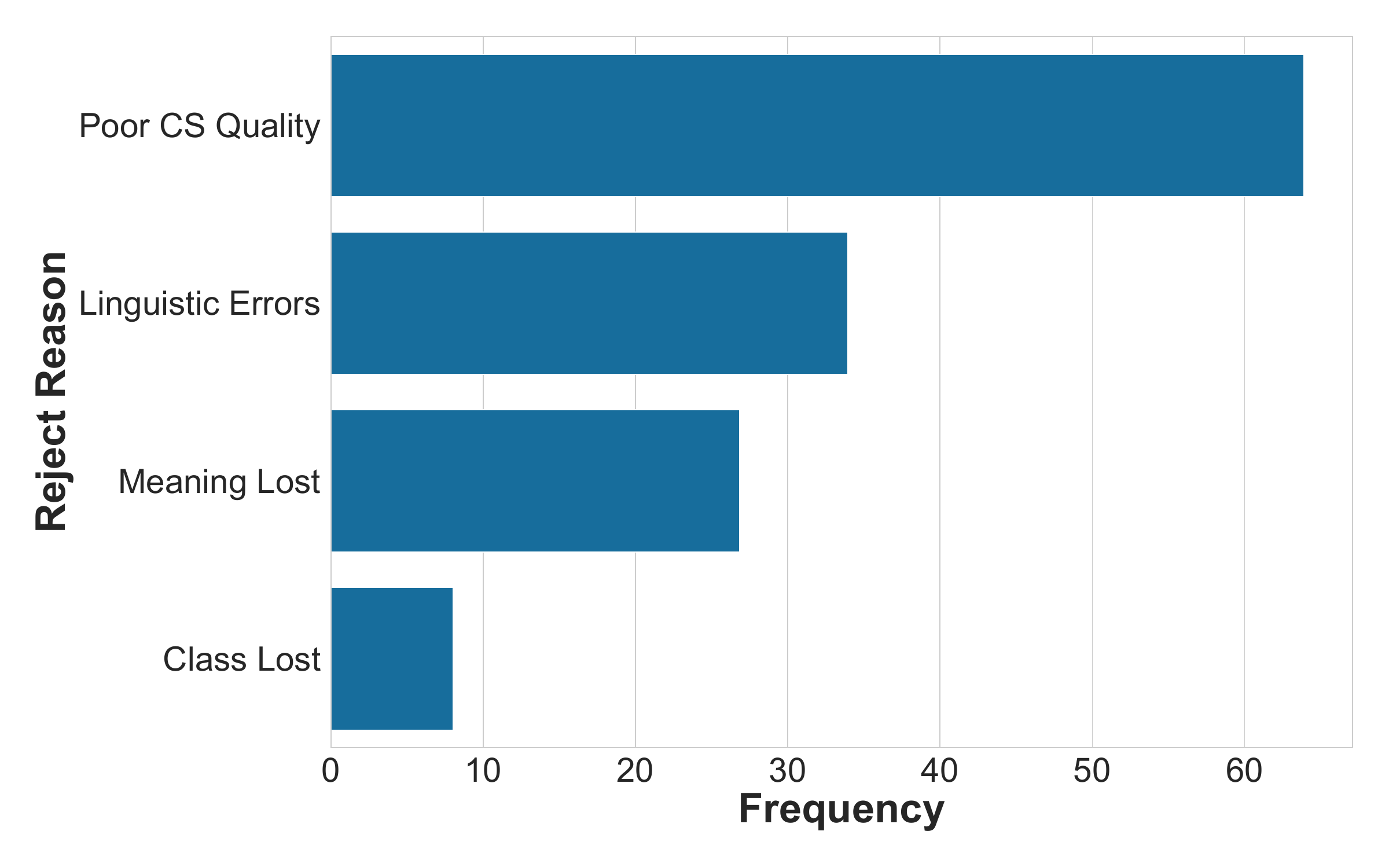}
\caption{The main reasons the samples were rejected, as a percentage of the rejected samples. Note that multiple reasons could lead to a rejected outcome.}
\label{fig:reject_reasons}
\end{figure}

\begin{figure}[!ht]
\centering
\includegraphics[width=0.7\columnwidth,trim={0 0 0 4cm},clip]{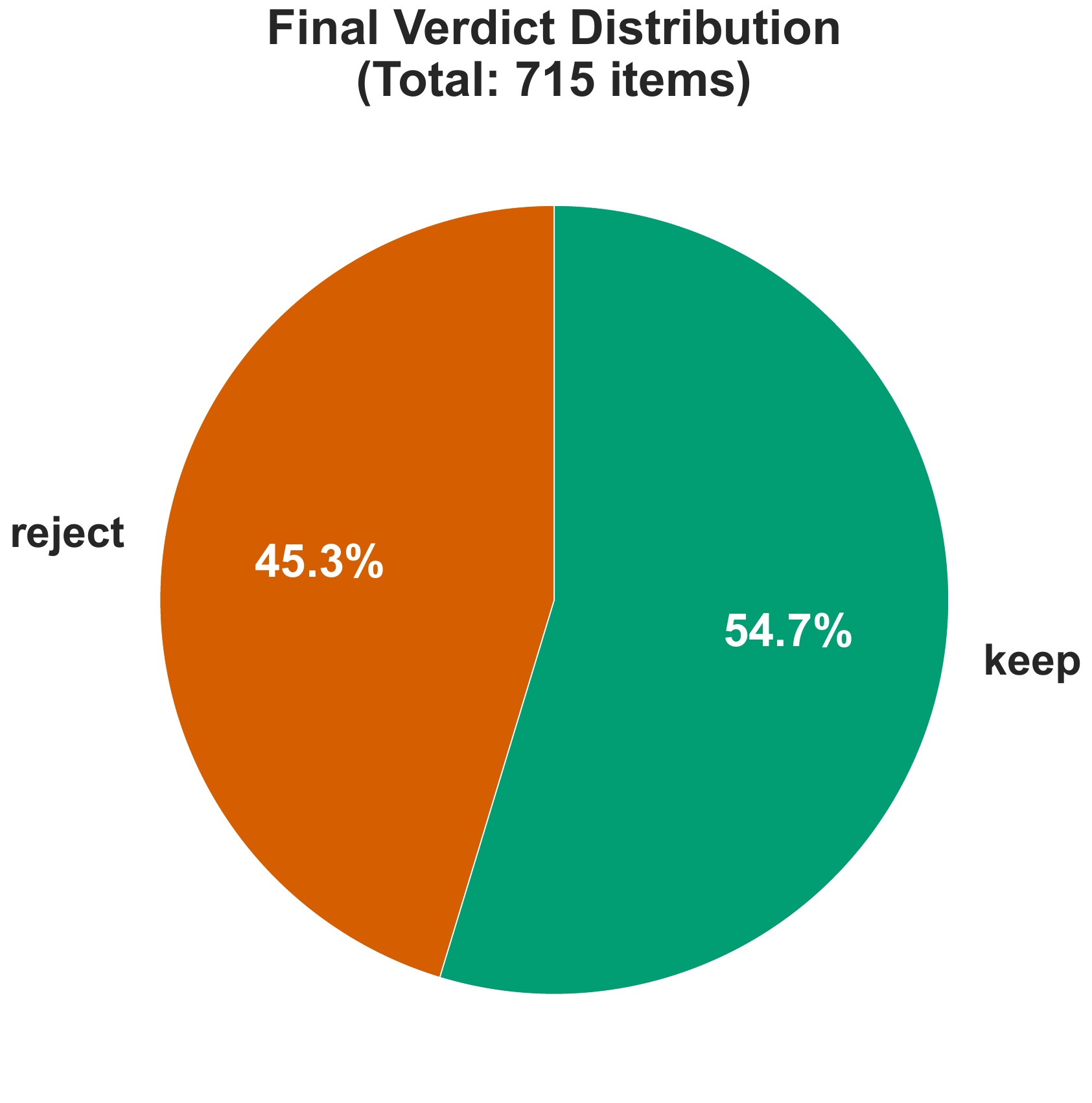}
\caption{Final verdict distribution after the adjudication process.}
\label{fig:adjudicated_all_verdict_pie}
\end{figure}

\textbf{Label Distribution.} Following human-led adjudication on the initially selected data, we derived a final high-quality benchmark of 391 samples (54.7\% of the total annotated set, see Figure~\ref{fig:adjudicated_all_verdict_pie}), as presented in \S\ref{app:annotation_protocol}. The per-language acceptance distribution is presented in Table \ref{tab:language_distribution}. Investigating the acceptance rate at a more granular level in Figure \ref{fig:language_comparison}, the LLM-generation and dependency-based approaches achieved the highest rates, over 60\% and, in some cases, 80\%. Also, as expected, the random replacement method generates low-quality samples (i.e., acceptance rate is under 20\%), mainly due to violations of linguistic rules or mistranslations caused by a lack of context. The most common reason for rejection was the poor quality of code-switching, with 63.9\% of the samples being annotated as poor code-switching quality (see Figure \ref{fig:reject_reasons}). This is followed by a significant margin by linguistic errors (34\%) and meaning loss (26.9\%). Only 8\% of the rejection cases involved a loss of class label, usually due to ambiguity or clarity reasons.

\begin{table}[!t]
\centering
\small
\begin{tabular}{lccc}
\toprule
\textbf{Language} & \textbf{Adjudicated} & \textbf{Keep} & \textbf{AR (\%)} \\
\midrule
English & 199 & 117 & 58.8\% \\
German & 159 & 70  & 44.0\% \\
French & 198 & 118 & 59.6\% \\
Italian & 159 & 86  & 54.1\% \\

\midrule
\textbf{Total} & \textbf{715} & \textbf{391} & \textbf{54.7\%} \\
\bottomrule
\end{tabular}
\caption{Language distribution of the original annotated set and human verification acceptance rates (AR) after the annotation process.}
\label{tab:language_distribution}
\end{table}

\begin{figure*}[!ht]
\centering
\includegraphics[width=\textwidth]{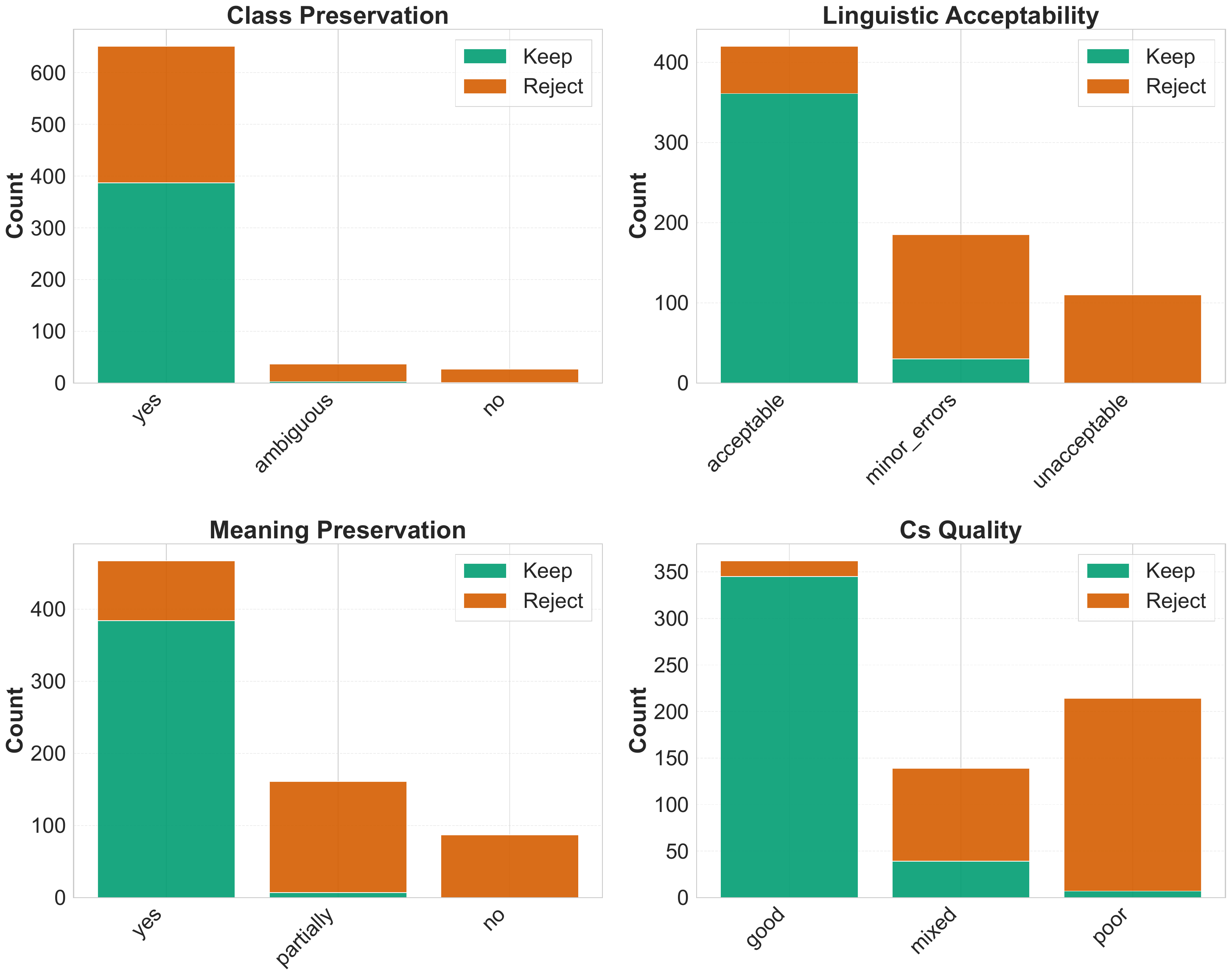}
\caption{Per-criteria distribution of the human labels after the adjudication process, on class preservation, linguistic acceptability, meaning preservation, and code-switching quality.}
\label{fig:adjudicated_all_verdict_by_criteria}
\end{figure*}

Investigating the label distribution per criterion (Figure \ref{fig:adjudicated_all_verdict_by_criteria}), we note a skew towards the \textit{yes} label for the \textit{class preservation} criterion. Comparing the keep/reject verdicts, we see that the positive class mostly yields a favorable outcome. However, for class preservation, the keep rate is lower for the \textit{yes} label than for similar labels in other categories, indicating that this metric is a weaker signal for the final verdict. 

\textbf{Criterion Importance.} To investigate which criteria most strongly influence annotators' final verdicts, we employ logistic regression modeling. We train a separate model for each language as well as a global one. In each model, the final verdict (keep/reject) is the binary dependent variable, while the four other criteria are independent variables. This allows us to compute regression coefficients that quantify the relative importance of each criterion in determining the final verdict. Formally, we model:

\begin{equation}
\begin{split}
    \textit{FinalVerdict} = {} & \beta_0 + \beta_1 \cdot CP + \beta_2 \cdot LA \\
    & + \beta_3 \cdot MP + \beta_4 \cdot CQ + \epsilon
\end{split}
\end{equation}
where $CP$ is class preservation, $LA$ is linguistic acceptability, $MP$ is meaning preservation, $CQ$ is code-switching quality, and $\epsilon$ is the error term representing unexplained variance in the model. The magnitude and sign of each coefficient ($\beta_i$) indicate the strength and direction of influence for each criterion. Larger absolute values suggest that the criterion has a stronger effect on the final verdict decision.

\begin{table}[!ht]
\centering
\small
\begin{tabular}{lccccc}
\toprule
\textbf{Lang.} & \textbf{CP} & \textbf{LA} & \textbf{MP} & \textbf{CQ}  & \textbf{Acc.} \\
\midrule
English & -0.001 & 1.373 & 2.070 & 1.844 & 94.0 \\
German  & 0.770 & 1.883 & 1.004 & 1.333  & 90.6\\
French  & 0.209 & 1.232 & 2.127 & 1.833  & 93.4\\
Italian & 0.156 & 1.759 & 1.468 & 1.884  & 91.8\\
\midrule
\textbf{All} & 0.556 & 1.858 & 2.089 & 1.858 & 92.9 \\
\bottomrule
\end{tabular}
\caption{Logistic regression coefficients indicating criterion influence on final verdict, along with the model accuracy (Acc.) as a percentage. Higher values indicate a stronger influence on accepting the sample.}
\label{tab:global_regression_table}
\end{table}

\begin{figure*}[!htbp]
\centering
\includegraphics[width=0.8\textwidth]{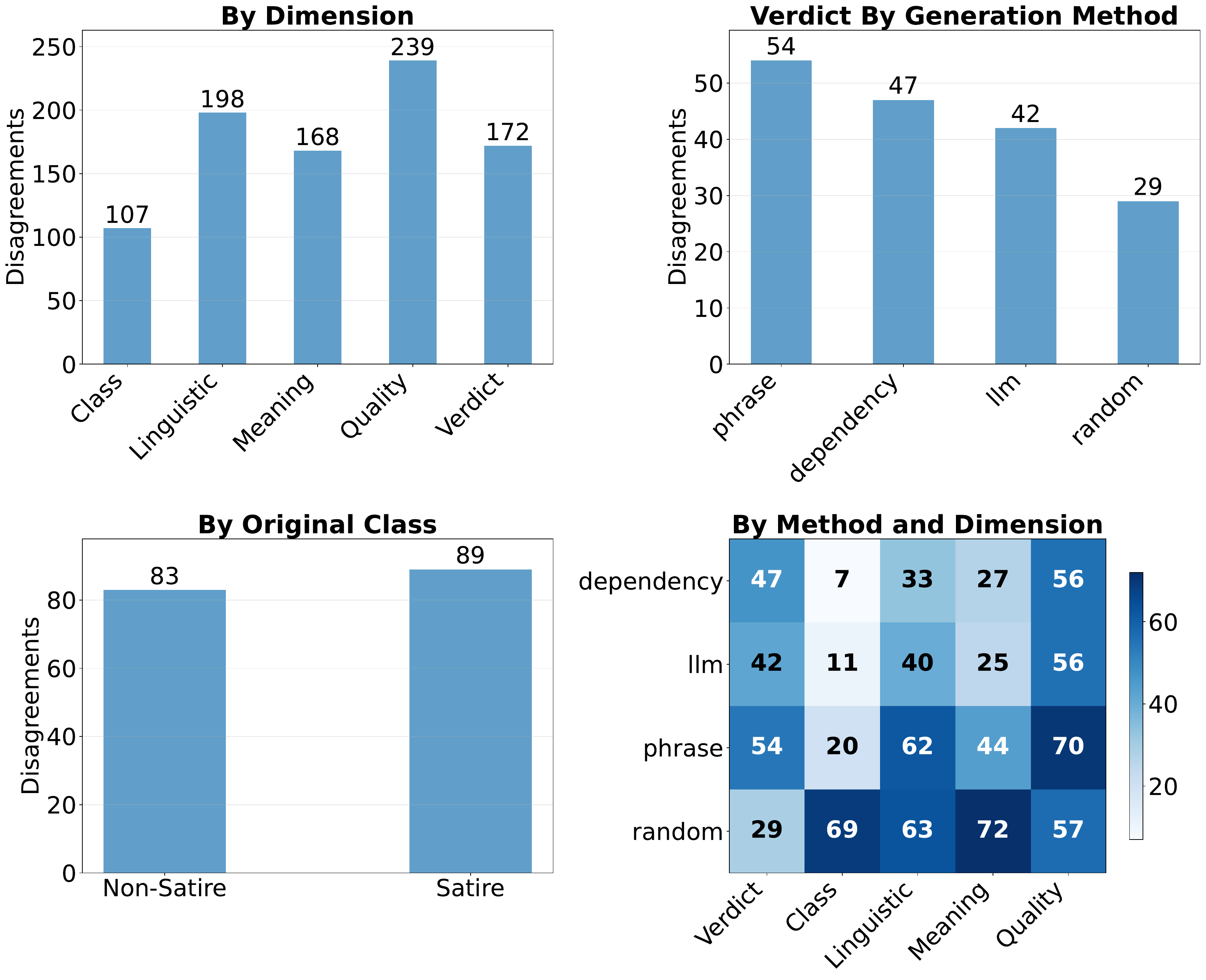}
\caption{Main disagreements between annotators considering category, generation method, and original class.}
\label{fig:global_patterns}
\end{figure*}

We train the logistic regression for at most 1,000 epochs or until convergence. Table~\ref{tab:global_regression_table} presents the logistic regression coefficients learned across target languages and globally. The regression models achieve over 90\% in-sample accuracy in predicting the final verdict from the constituent criteria. Upon investigating the regression coefficients, we confirm that class preservation yields the lowest signal. On the other hand, meaning preservation represents the most important feature, while language acceptability and code-switching quality are, to a lesser degree, similarly important.

\textbf{Inter-Annotator Agreement.} Table \ref{tab:global_iaa} shows the overall inter-annotator agreement across criteria. The agreement between annotators for the final verdict reaches 75.9\% raw agreement with a Krippendorff's alpha of $\alpha = 0.518$, indicating moderate agreement. Linguistic acceptability ($\alpha = 0.603$), meaning preservation ($\alpha = 0.609$), and code-switching quality ($\alpha = 0.655$) achieve moderate to substantial agreement. For \textit{class preservation}, raw agreement is high (85.0\%), but $\alpha = 0.178$ due to the prevalence paradox caused by the extreme skew towards the \textit{yes} label.
In Table \ref{tab:iaa_by_language}, we show the breakdown by language and criteria. As noted before, class preservation presents the lowest alpha across languages due to the label skew, while other criteria exhibit moderate to substantial agreement.

\begin{table}[ht]
\centering
\small
\begin{tabular}{lcc}
\toprule
\textbf{Criterion} & \textbf{$\alpha$} & \textbf{Agreement (\%)} \\
\midrule
Class Preservation  & 0.178 & 85.0\% \\
Linguistic Acceptability  & 0.603 & 72.3\% \\
Meaning Preservation  & 0.609 & 76.5\% \\
CS Quality  & 0.655 & 66.6\% \\
Final Verdict  & 0.518 & 75.9\% \\
\bottomrule
\end{tabular}
\caption{Inter-annotator agreement in terms of Krippendorff's alpha ($\alpha$) and agreement rate across all criteria for all N=715 samples.}
\label{tab:global_iaa}
\end{table}
\begin{table}[ht]
\centering
\small
\setlength{\tabcolsep}{4pt}
\begin{tabular}{lccccc}
\toprule
\textbf{Language} & \textbf{CP} & \textbf{LA} & \textbf{MP} & \textbf{CSQ} & \textbf{Verdict}\\
\midrule
English & 0.205  & 0.476 & 0.629 & 0.679 & 0.438 \\
German  & 0.315  & 0.723 & 0.503 & 0.731 & 0.634 \\
French  & -0.024 & 0.364 & 0.644 & 0.514 & 0.416 \\
Italian & 0.220  & 0.672 & 0.675 & 0.670 & 0.617 \\
\bottomrule
\end{tabular}
\caption{Inter-annotator agreement (Krippendorff's $\alpha$) across target languages by verdict, linguistic acceptability (LA), code-switching quality (CSQ), class preservation (CP), and meaning preservation (MP).}
\label{tab:iaa_by_language}
\end{table}

\textbf{Disagreement Assessment.} We investigate the main reasons for disagreement during the annotation process in Figure \ref{fig:global_patterns}. The main categories where we noticed disagreements were code-switching quality assessment, especially at the phrase level. Additionally, random generation posed greater challenges when considering all characteristics except the verdict, for which annotators generally agreed that those same characteristics should be rejected. The fewest challenges were raised by class preservation and meaning preservation across methods, except for random.
The original class label of the sentence (i.e., whether satire or non-satire) does not provide any meaningful influence on the results.


\subsection{Gold Standard Dataset Statistics}
\label{app:gold_standard_dataset_statistics}

The final dataset resulted from the adjudication process; 391 samples remained in the gold set with ``keep'' as the final verdict (i.e., 54.7\%).

In Table \ref{tab:gold_standard_distribution}, we present the distribution of samples in the gold standard dataset across code-switching methods and target languages. Most samples are from LLM and dependency-based generation, while the fewest are from random-based generation. Considering the language distribution, there are approximately 120 samples mixed in English and French, and 70-86 mixed in German and Italian. In Table \ref{tab:gold_standard_summary}, we provide summary statistics of the gold standard dataset used for evaluating code-switching generation methods. The final distribution of the data is 55.8\% satirical samples and 44.2\% regular samples (218 satire vs.\ 173 regular). The average sentence length is 21 words, with overall coverage of method-language-$k$ combinations at 89.6\% (43 of 48 cells covered, with German-random having 1 sample). 

Table \ref{tab:cs_metrics_breakdown} presents the code-switching statistics over each target language and generation method. Overall, we notice that for methods such as random, the human annotators preferred those with fewer switching points. Other metrics, such as those for phrase, dependency, and LLM-based generation, remained generally unchanged.

\begin{table}[!t]
\centering
\small
\begin{tabular}{lrrrrr}
\toprule
\textbf{Method} & \textbf{EN} & \textbf{DE} & \textbf{FR} & \textbf{IT} & \textbf{Total} \\
\midrule
Random & 4 & 1 & 9 & 3 & 17 \\
Phrase & 25 & 12 & 26 & 17 & 80 \\
Dependency & 43 & 26 & 39 & 31 & 139 \\
LLM & 45 & 31 & 44 & 35 & 155 \\
\midrule
\textbf{Total} & \textbf{117} & \textbf{70} & \textbf{118} & \textbf{86} & \textbf{391} \\
\bottomrule
\end{tabular}
\caption{Distribution of the gold standard dataset across code-switching generation methods and target languages.}
\label{tab:gold_standard_distribution}
\end{table}

\begin{table}[!t]
\centering
\small
\begin{tabular}{lr}
\toprule
\textbf{Statistic} & \textbf{Value} \\
\midrule
Total samples & 391 \\
Unique source sentences & 50 \\
CS methods & 4 \\
Target languages & 4 \\
K values & 3 \\
\midrule
Satire samples & 218 (55.8\%) \\
Non-satire samples & 173 (44.2\%) \\
\midrule
Avg sentence length & 21.0 words \\
Method-language pair coverage & 89.6\% \\
\bottomrule
\end{tabular}
\caption{Summary statistics of the gold standard dataset used for evaluating code-switching generation methods.}
\label{tab:gold_standard_summary}
\end{table}

\begin{table}[!t]
\centering
\setlength{\tabcolsep}{2pt}
\resizebox{0.8\columnwidth}{!}{%
\begin{tabular}{lccccc}
\toprule
\textbf{Method} & \textbf{CMI} & \textbf{M-Index} & \textbf{I-Index} & \textbf{LE} & \textbf{SP-Avg} \\
\midrule
\multicolumn{6}{c}{\textit{RO $\rightarrow$ EN}} \\
\midrule
Random & 24.60 & 0.28 & 0.34 & 0.65 & 0.34 \\
Phrase & 15.51 & 0.18 & 0.21 & 0.56 & 0.21 \\
Dependency & 23.47 & 0.26 & 0.19 & 0.68 & 0.19 \\
LLM & 17.14 & 0.20 & 0.20 & 0.59 & 0.20 \\
\midrule
\multicolumn{6}{c}{\textit{RO $\rightarrow$ DE}} \\
\midrule
Random & - & - & - & - & - \\
Phrase & 27.71 & 0.32 & 0.29 & 0.81 & 0.29 \\
Dependency & 25.77 & 0.29 & 0.20 & 0.75 & 0.20 \\
LLM & 21.00 & 0.25 & 0.26 & 0.70 & 0.26 \\
\midrule
\multicolumn{6}{c}{\textit{RO $\rightarrow$ FR}} \\
\midrule
Random & 36.52 & 0.41 & 0.40 & 0.86 & 0.40 \\
Phrase & 23.18 & 0.27 & 0.29 & 0.75 & 0.29 \\
Dependency & 22.38 & 0.26 & 0.22 & 0.70 & 0.22 \\
LLM & 22.88 & 0.27 & 0.27 & 0.73 & 0.27 \\
\midrule
\multicolumn{6}{c}{\textit{RO $\rightarrow$ IT}} \\
\midrule
Random & 14.29 & 0.16 & 0.19 & 0.33 & 0.19 \\
Phrase & 23.80 & 0.27 & 0.29 & 0.74 & 0.29 \\
Dependency & 23.13 & 0.25 & 0.24 & 0.74 & 0.24 \\
LLM & 25.94 & 0.28 & 0.28 & 0.80 & 0.28 \\
\bottomrule
\end{tabular}}
\caption{Aggregated code-switching metrics by language and code-switching method on the gold set (i.e., keep set, N=391).}
\label{tab:cs_metrics_breakdown}
\end{table}

\section{Extended Results and Error Analysis}
\label{app:error_analysis}

\subsection{Monolingual Baseline Evaluation}
\label{app:monolingual_baseline}

\begin{table*}[!t]
\centering
\small
\setlength{\tabcolsep}{3pt}
\begin{tabular}{lcccccc}
\toprule
\multicolumn{1}{c}{\multirow{2}{*}{\textbf{Model}}}
    & \multicolumn{2}{c}{\textbf{Zero-Shot}}
    & \multicolumn{2}{c}{\textbf{Chain-of-Thought}}
    & \multicolumn{2}{c}{\textbf{Fine-Tuning}} \\
    & \textbf{Acc (\%)} & \textbf{F1 (\%)}
    & \textbf{Acc (\%)} & \textbf{F1 (\%)}
    & \textbf{Acc (\%)} & \textbf{F1 (\%)} \\
\midrule
Llama 3.1 8B & 43.9$_{\scriptstyle\pm14.9}$ & 46.3$_{\scriptstyle\pm11.1}$ & 52.2$_{\scriptstyle\pm2.3}$ & 53.3$_{\scriptstyle\pm1.9}$ & 69.0$_{\scriptstyle\pm0.1}$ & 69.0$_{\scriptstyle\pm0.1}$ \\
Gemma 3 27B & 53.0$_{\scriptstyle\pm0.4}$ & 49.9$_{\scriptstyle\pm0.6}$ & 55.5$_{\scriptstyle\pm0.2}$ & 53.2$_{\scriptstyle\pm0.4}$ & 74.9$_{\scriptstyle\pm1.4}$ & 74.3$_{\scriptstyle\pm1.3}$ \\
Gemma 4 26B A4B & 66.9$_{\scriptstyle\pm0.0}$ & 67.0$_{\scriptstyle\pm0.0}$ & 68.0$_{\scriptstyle\pm0.5}$ & 68.1$_{\scriptstyle\pm0.5}$ & 70.1$_{\scriptstyle\pm0.5}$ & 70.2$_{\scriptstyle\pm0.5}$ \\
Gemma 4 31B & 70.2$_{\scriptstyle\pm0.1}$ & 70.2$_{\scriptstyle\pm0.1}$ & \textbf{69.4$_{\scriptstyle\pm0.1}$} & \textbf{69.4$_{\scriptstyle\pm0.1}$} & \textbf{78.5$_{\scriptstyle\pm0.8}$} & \textbf{78.3$_{\scriptstyle\pm0.8}$} \\
RoMistral 7B & 39.7$_{\scriptstyle\pm14.2}$ & 41.2$_{\scriptstyle\pm11.5}$ & 48.6$_{\scriptstyle\pm0.2}$ & 49.6$_{\scriptstyle\pm0.6}$ & 73.1$_{\scriptstyle\pm1.7}$ & 72.5$_{\scriptstyle\pm1.4}$ \\
RoLlama 3.1 8B & 61.0$_{\scriptstyle\pm0.3}$ & 61.6$_{\scriptstyle\pm0.8}$ & 51.9$_{\scriptstyle\pm1.5}$ & 50.4$_{\scriptstyle\pm2.1}$ & 73.9$_{\scriptstyle\pm0.7}$ & 73.3$_{\scriptstyle\pm0.8}$ \\
RoGemma 2 9B & 60.4$_{\scriptstyle\pm0.3}$ & 60.3$_{\scriptstyle\pm0.3}$ & 62.5$_{\scriptstyle\pm1.5}$ & 62.8$_{\scriptstyle\pm1.4}$ & 73.7$_{\scriptstyle\pm0.7}$ & 73.1$_{\scriptstyle\pm0.9}$ \\
GPT-4.1 mini & 63.4 & 63.3 & 68.0 & 68.0 & - & - \\
GPT-5 mini & \textbf{71.9} & \textbf{71.3} & - & - & - & - \\
GPT-6 Luna & 68.2 & 68.1 & - & - & - & - \\
\bottomrule
\end{tabular}
\caption{Monolingual baseline results on the Romanian-only (non-code-switched) test split across zero-shot (ZS), chain-of-thought (CoT), and fine-tuning (FT) settings. Bold indicates the best performance.}
\label{tab:baseline_results}
\end{table*}

\begin{table*}[!t]
\centering
\small
\setlength{\tabcolsep}{3pt}
\begin{tabular}{lcccccc}
\toprule
\multicolumn{1}{c}{\multirow{2}{*}{\textbf{Model}}}
    & \multicolumn{2}{c}{\textbf{Zero-Shot}}
    & \multicolumn{2}{c}{\textbf{Chain-of-Thought}}
    & \multicolumn{2}{c}{\textbf{Fine-Tuning}} \\
    & \textbf{Acc (\%)} & \textbf{F1 (\%)}
    & \textbf{Acc (\%)} & \textbf{F1 (\%)}
    & \textbf{Acc (\%)} & \textbf{F1 (\%)} \\
\midrule
Llama 3.1 8B & 46.0$_{\scriptstyle\pm24.2}$ & 47.5$_{\scriptstyle\pm21.3}$ & 48.7$_{\scriptstyle\pm2.3}$ & 49.8$_{\scriptstyle\pm2.2}$ & 60.7$_{\scriptstyle\pm3.1}$ & 60.6$_{\scriptstyle\pm3.1}$ \\
Gemma 3 27B & 42.7$_{\scriptstyle\pm2.3}$ & 39.5$_{\scriptstyle\pm2.8}$ & 46.7$_{\scriptstyle\pm7.0}$ & 44.3$_{\scriptstyle\pm8.1}$ & 68.0$_{\scriptstyle\pm5.3}$ & 66.7$_{\scriptstyle\pm5.3}$ \\
Gemma 4 26B A4B & 52.0$_{\scriptstyle\pm0.0}$ & 51.9$_{\scriptstyle\pm0.0}$ & 60.0$_{\scriptstyle\pm0.0}$ & 60.0$_{\scriptstyle\pm0.8}$ & 57.3$_{\scriptstyle\pm2.3}$ & 56.9$_{\scriptstyle\pm2.6}$ \\
Gemma 4 31B & \textbf{64.7$_{\scriptstyle\pm1.2}$} & 63.1$_{\scriptstyle\pm1.0}$ & \textbf{66.0$_{\scriptstyle\pm2.0}$} & 64.7$_{\scriptstyle\pm2.1}$ & \textbf{72.0$_{\scriptstyle\pm2.0}$} & \textbf{70.8$_{\scriptstyle\pm1.8}$} \\
RoMistral 7B & 35.3$_{\scriptstyle\pm16.8}$ & 35.8$_{\scriptstyle\pm15.7}$ & 40.7$_{\scriptstyle\pm1.2}$ & 41.8$_{\scriptstyle\pm1.9}$ & 65.3$_{\scriptstyle\pm3.1}$ & 64.5$_{\scriptstyle\pm3.5}$ \\
RoLlama 3.1 8B & 56.0$_{\scriptstyle\pm2.0}$ & 56.9$_{\scriptstyle\pm1.0}$ & 46.7$_{\scriptstyle\pm4.2}$ & 46.4$_{\scriptstyle\pm4.4}$ & 62.7$_{\scriptstyle\pm3.1}$ & 60.6$_{\scriptstyle\pm3.0}$ \\
RoGemma 2 9B & 59.3$_{\scriptstyle\pm5.0}$ & 59.1$_{\scriptstyle\pm4.8}$ & \textbf{66.0$_{\scriptstyle\pm2.0}$} & \textbf{65.8$_{\scriptstyle\pm2.4}$} & 69.3$_{\scriptstyle\pm4.6}$ & 67.6$_{\scriptstyle\pm5.1}$ \\
GPT-4.1 mini & 50.0 & 50.0 & 62.0 & 61.9 & - & - \\
GPT-5 mini & 62.0 & 60.1 & - & - & - & - \\
GPT-6 Luna & 64.0 & \textbf{63.9} & - & - & - & - \\
\bottomrule
\end{tabular}
\caption{Monolingual baseline results on the original 50 Romanian sentences used to derive CS-SeLeRoSa-Gold test set across zero-shot (ZS), chain-of-thought (CoT), and fine-tuning (FT) settings. Bold indicates the best performance.}
\label{tab:baseline_results_gs}
\end{table*}

In Table~\ref{tab:baseline_results}, we present baseline results on the original Romanian-only SeLeRoSa dataset across the evaluated models. The more recent Gemma 4 family models outperform other open-weights models by a considerable margin across all settings. The best results are achieved by the 31B dense model, achieving 70.2\% in the zero-shot setting, and over 78\% in the fine-tuning setting. Romanian adaptation provides a clear advantage for smaller architectures in the zero-shot regime; for instance, RoLlama 3.1 8B outperforms base Llama 3.1 8B by 15.3 F1 points on the monolingual baseline (61.6\% vs 46.3\%), a gap that narrows under fine-tuning. This shows that Romanian pretraining improves models' ability to detect satire in text through zero-shot transfer, despite not being trained on satirical data \cite{masala-etal-2024-vorbesti}. In our setup, chain-of-thought shows mixed results. For models such as Gemma 3 27B, Gemma 4 26B, RoMistral 7B, and RoGemma 2 9B, the performance improvement varies between 1 and 10 percentage points. Fine-tuning brings most models over 70\% accuracy and F1-scores, largely outperforming the in-context-learning setup. Without fine-tuning, proprietary models like GPT-5 mini (71.3\% F1) and GPT-6 Luna (68.1\% F1) outperform older open-weights models, though newer models like Gemma 4 31B remain highly competitive (70.2\% F1). Among the closed-source models, GPT-4.1 mini underperforms in the zero-shot setting, being the only proprietary model evaluated without any reasoning trace. Incorporating chain-of-thought, its performance increases to 68\% across metrics. For Llama 3.1 8B and RoMistral 7B, we notice higher variance in the zero-shot setting, which, as detailed in Appendix~\ref{app:instruction_following_error_rate}, is a result of the models failing to follow instructions and generating invalid predictions. Analysis regarding why RoLlama 3.1 8B underperforms when employing chain-of-thought compared with the zero-shot setting is presented in Appendix~\ref{app:rollama_baseline_degradation}. Table~\ref{tab:baseline_results_gs} presents the results on the 50 original Romanian sentences used to derive the gold set. In this setting, besides Gemma 4 31B, we also note other models such as GPT-6 Luna in the zero-shot setting, and RoGemma 2 9B under chain-of-thought achieving competitive performance. In the fine-tuning regime, Gemma 4 31B remains the top-performing model.

\subsection{Code-Switching Benchmark Results}
\label{app:overall_cs_results}

\begin{table*}[!t]
\small
\centering
\setlength{\tabcolsep}{2pt}
\begin{tabular}{lcccccc}
\toprule
\multicolumn{1}{c}{\multirow{2}{*}{\textbf{Model}}}
    & \multicolumn{2}{c}{\textbf{Zero-Shot}}
    & \multicolumn{2}{c}{\textbf{Chain-of-Thought}}
    & \multicolumn{2}{c}{\textbf{Fine-Tuning}} \\
    & \textbf{Acc (\%)} & \textbf{F1 (\%)}
    & \textbf{Acc (\%)} & \textbf{F1 (\%)}
    & \textbf{Acc (\%)} & \textbf{F1 (\%)} \\
\midrule
Llama 3.1 8B & 42.0$_{\scriptstyle\pm19.6}$ & 44.8$_{\scriptstyle\pm15.2}$ & 47.8$_{\scriptstyle\pm2.6}$$^{\dagger}$ & 47.9$_{\scriptstyle\pm2.2}$$^{\dagger}$ & 75.8$_{\scriptstyle\pm0.2}$$^{\dagger}$ & 74.9$_{\scriptstyle\pm0.8}$$^{\dagger}$ \\
Gemma 3 27B & 44.9$_{\scriptstyle\pm0.2}$$^{\dagger}$ & 37.0$_{\scriptstyle\pm0.3}$$^{\dagger}$ & 47.3$_{\scriptstyle\pm1.0}$$^{\dagger}$ & 41.3$_{\scriptstyle\pm1.8}$$^{\dagger}$ & 75.7$_{\scriptstyle\pm0.9}$$^{\dagger}$ & 75.4$_{\scriptstyle\pm0.8}$$^{\dagger}$ \\
Gemma 4 26B A4B & 63.8$_{\scriptstyle\pm0.0}$$^{\dagger}$ & 63.7$_{\scriptstyle\pm0.1}$$^{\dagger}$ & 64.3$_{\scriptstyle\pm0.1}$$^{\dagger}$ & 64.3$_{\scriptstyle\pm0.1}$$^{\dagger}$ & 70.2$_{\scriptstyle\pm0.6}$$^{\dagger}$ & 70.2$_{\scriptstyle\pm0.6}$$^{\dagger}$ \\
Gemma 4 31B & 68.1$_{\scriptstyle\pm0.0}$ & 68.1$_{\scriptstyle\pm0.0}$ & \textbf{66.5$_{\scriptstyle\pm0.0}$} & \textbf{66.5$_{\scriptstyle\pm0.0}$} & \textbf{78.0$_{\scriptstyle\pm0.1}$} & \textbf{77.5$_{\scriptstyle\pm0.1}$} \\
RoMistral 7B & 35.7$_{\scriptstyle\pm15.2}$ & 36.8$_{\scriptstyle\pm14.4}$ & 46.3$_{\scriptstyle\pm0.7}$$^{\dagger}$ & 48.3$_{\scriptstyle\pm1.0}$$^{\dagger}$ & 75.8$_{\scriptstyle\pm0.2}$$^{\dagger}$ & 74.6$_{\scriptstyle\pm0.6}$$^{\dagger}$ \\
RoLlama 3.1 8B & 53.7$_{\scriptstyle\pm0.4}$$^{\dagger}$ & 52.6$_{\scriptstyle\pm0.9}$$^{\dagger}$ & 48.4$_{\scriptstyle\pm0.8}$$^{\dagger}$ & 45.9$_{\scriptstyle\pm1.0}$$^{\dagger}$ & 73.9$_{\scriptstyle\pm0.1}$$^{\dagger}$ & 73.4$_{\scriptstyle\pm0.1}$$^{\dagger}$ \\
RoGemma 2 9B & 55.6$_{\scriptstyle\pm0.5}$$^{\dagger}$ & 55.0$_{\scriptstyle\pm0.7}$$^{\dagger}$ & 59.9$_{\scriptstyle\pm0.8}$$^{\dagger}$ & 60.0$_{\scriptstyle\pm0.8}$$^{\dagger}$ & 76.1$_{\scriptstyle\pm0.1}$$^{\dagger}$ & 75.4$_{\scriptstyle\pm0.0}$$^{\dagger}$ \\
GPT-4.1 mini & 57.0$^{\dagger}$ & 56.0$^{\dagger}$ & 62.0$^{\dagger}$ & 61.9$^{\dagger}$ & - & - \\
GPT-5 mini & \textbf{70.3}$^{\dagger}$ & \textbf{69.5}$^{\dagger}$ & - & - & - & - \\
GPT-6 Luna & 67.1$^{\dagger}$ & 67.0$^{\dagger}$ & - & - & - & - \\
\bottomrule
\end{tabular}
\caption{Results on the code-switched dataset across zero-shot (ZS), chain-of-thought (CoT), and fine-tuning (FT) settings. Bold indicates the best performance. $^{\dagger}$ indicates statistically significant difference compared to Gemma 4 31B, or its fine-tuned variant in the FT setting ($p < 0.05$; for $N>1$, Welch's $t$-test; for $N=1$, paired McNemar's test for accuracy and a paired bootstrap for F1).}
\label{tab:results}
\end{table*}

\begin{table*}[!t]
\small
\centering
\setlength{\tabcolsep}{2pt}
\begin{tabular}{lcccccc}
\toprule
\multicolumn{1}{c}{\multirow{2}{*}{\textbf{Model}}}
    & \multicolumn{2}{c}{\textbf{Zero-Shot}}
    & \multicolumn{2}{c}{\textbf{Chain-of-Thought}}
    & \multicolumn{2}{c}{\textbf{Fine-Tuning}} \\
    & \textbf{Acc (\%)} & \textbf{F1 (\%)}
    & \textbf{Acc (\%)} & \textbf{F1 (\%)}
    & \textbf{Acc (\%)} & \textbf{F1 (\%)} \\
\midrule
Llama 3.1 8B & 40.2$_{\scriptstyle\pm17.2}$ & 43.0$_{\scriptstyle\pm12.8}$ & 43.0$_{\scriptstyle\pm2.3}$$^{\dagger}$ & 43.2$_{\scriptstyle\pm2.2}$$^{\dagger}$ & 67.7$_{\scriptstyle\pm1.0}$$^{\dagger}$ & 65.7$_{\scriptstyle\pm0.0}$$^{\dagger}$ \\
Gemma 3 27B & 41.9$_{\scriptstyle\pm0.7}$$^{\dagger}$ & 36.1$_{\scriptstyle\pm0.8}$$^{\dagger}$ & 43.3$_{\scriptstyle\pm2.4}$$^{\dagger}$ & 39.1$_{\scriptstyle\pm3.2}$$^{\dagger}$ & 68.7$_{\scriptstyle\pm1.2}$$^{\dagger}$ & 67.2$_{\scriptstyle\pm0.8}$$^{\dagger}$ \\
Gemma 4 26B A4B & 58.2$_{\scriptstyle\pm0.3}$$^{\dagger}$ & 58.2$_{\scriptstyle\pm0.3}$$^{\dagger}$ & 61.5$_{\scriptstyle\pm0.4}$$^{\dagger}$ & 61.5$_{\scriptstyle\pm0.3}$$^{\dagger}$ & 63.9$_{\scriptstyle\pm0.8}$$^{\dagger}$ & 63.7$_{\scriptstyle\pm0.8}$$^{\dagger}$ \\
Gemma 4 31B & \textbf{62.4$_{\scriptstyle\pm0.3}$} & \textbf{61.8$_{\scriptstyle\pm0.2}$} & \textbf{63.6$_{\scriptstyle\pm0.4}$} & \textbf{63.0$_{\scriptstyle\pm0.5}$} & \textbf{73.9$_{\scriptstyle\pm0.3}$} & \textbf{72.6$_{\scriptstyle\pm0.4}$} \\
RoMistral 7B & 35.0$_{\scriptstyle\pm14.0}$ & 35.7$_{\scriptstyle\pm13.7}$ & 44.2$_{\scriptstyle\pm1.8}$$^{\dagger}$ & 46.6$_{\scriptstyle\pm1.5}$$^{\dagger}$ & 73.0$_{\scriptstyle\pm0.6}$ & 70.4$_{\scriptstyle\pm0.5}$$^{\dagger}$ \\
RoLlama 3.1 8B & 54.4$_{\scriptstyle\pm0.7}$$^{\dagger}$ & 54.8$_{\scriptstyle\pm1.5}$$^{\dagger}$ & 47.7$_{\scriptstyle\pm2.3}$$^{\dagger}$ & 46.9$_{\scriptstyle\pm2.5}$$^{\dagger}$ & 67.3$_{\scriptstyle\pm0.4}$$^{\dagger}$ & 66.3$_{\scriptstyle\pm0.9}$$^{\dagger}$ \\
RoGemma 2 9B & 56.9$_{\scriptstyle\pm1.5}$$^{\dagger}$ & 56.8$_{\scriptstyle\pm1.5}$$^{\dagger}$ & 62.1$_{\scriptstyle\pm1.0}$ & 62.1$_{\scriptstyle\pm0.9}$ & 70.7$_{\scriptstyle\pm1.5}$ & 68.9$_{\scriptstyle\pm1.5}$$^{\dagger}$ \\
GPT-4.1 mini & 54.1$_{\scriptstyle\pm3.3}$$^{\dagger}$ & 50.4$_{\scriptstyle\pm0.8}$$^{\dagger}$ & 59.8 & 59.8 & - & - \\
GPT-5 mini & 60.1$_{\scriptstyle\pm2.8}$ & 59.6$_{\scriptstyle\pm2.0}$ & - & - & - & - \\
GPT-5 & 61.2$_{\scriptstyle\pm1.5}$ & 61.1$_{\scriptstyle\pm1.6}$ & - & - & - & - \\
GPT-6 Luna & 59.1$_{\scriptstyle\pm2.4}$ & 58.1$_{\scriptstyle\pm1.7}$ & - & - & - & - \\
GPT-6 Astra & 59.4$_{\scriptstyle\pm0.3}$$^{\dagger}$ & 58.3$_{\scriptstyle\pm0.3}$$^{\dagger}$ & - & - & - & - \\
GPT-6 Sol & 56.7$_{\scriptstyle\pm0.7}$$^{\dagger}$ & 55.0$_{\scriptstyle\pm1.0}$$^{\dagger}$ & - & - & - & - \\
GPT-6.1 Sol & 60.1$_{\scriptstyle\pm1.6}$ & 59.1$_{\scriptstyle\pm1.7}$ & - & - & - & - \\
\bottomrule
\end{tabular}
\caption{Results on the gold dataset across zero-shot (ZS), chain-of-thought (CoT), and fine-tuning (FT) settings. Bold indicates the best performance. $^{\dagger}$ indicates statistically significant difference compared to Gemma 4 31B or its fine-tuned variant in the FT setting ($p < 0.05$; for $N>1$, Welch's $t$-test; for $N=1$, paired McNemar's test for accuracy and a paired bootstrap for F1).}
\label{tab:overall_gs_results}
\end{table*}

We present the results in Table~\ref{tab:results} on the entire test set, and Table~\ref{tab:overall_gs_results} on the gold evaluation dataset. In the zero-shot setting, Gemma 4 31B (68.1\% F1-score) achieves comparable results with proprietary reasoning models (e.g., GPT-5 mini, 69.5\%) or even surpasses them (e.g., GPT-6 Luna, 67.0\%). However, including chain-of-thought reasoning traces, the model underperforms, with both metrics achieving 66.5\%, while fine-tuning the model achieves over 77\% in both metrics. Among Romanian-adapted models, RoGemma 2 9B achieves the strongest results across all regimes (reaching 75.4\% under fine-tuning), outperforming other Romanian-adapted architectures like RoLlama and RoMistral.

When evaluating only on the gold set, the discrepancy between open-source and proprietary models remains diminished. Gemma 4 31B remains the best-performing model, achieving around 62--63\% in both metrics in zero-shot and chain-of-thought settings, and around 72--73\% in the fine-tuning setting. We observe that frontier models do not manage to surpass it, and, in fact, proprietary models tend to plateau just slightly under Gemma 4 31B. Romanian pre-trained models continue to transfer better than their base model counterparts in the zero-shot setting (e.g., RoLlama 52.6\% vs base Llama 44.8\% on CS-SeLeRoSa, and 54.8\% vs 43.0\% on CS-SeLeRoSa-Gold).

\subsection{Detailed Method and Language Breakdown}
\label{app:breakdown_results}

\begin{table*}[!th]
\centering
\small
\setlength{\tabcolsep}{3pt}
\resizebox{\textwidth}{!}{
\begin{tabular}{lcccccccccccc}
\toprule
\multicolumn{1}{c}{\multirow{2}{*}{\textbf{Model}}} & \multicolumn{3}{c}{\textbf{Random}} & \multicolumn{3}{c}{\textbf{Phrase}} & \multicolumn{3}{c}{\textbf{Dependency}} & \multicolumn{3}{c}{\textbf{LLM}} \\
 & \textbf{ZS} & \textbf{CoT} & \textbf{FT} & \textbf{ZS} & \textbf{CoT} & \textbf{FT} & \textbf{ZS} & \textbf{CoT} & \textbf{FT} & \textbf{ZS} & \textbf{CoT} & \textbf{FT} \\
\midrule
\multicolumn{13}{c}{\textit{RO $\rightarrow$ EN}} \\
\midrule
Llama 3.1 8B & 44.3$_{\scriptstyle\pm16.5}$$^{\dagger}$ & 44.0$_{\scriptstyle\pm4.0}$$^{\dagger}$ & 75.6$_{\scriptstyle\pm1.1}$ & 45.4$_{\scriptstyle\pm15.9}$ & 49.8$_{\scriptstyle\pm2.1}$$^{\dagger}$ & 75.0$_{\scriptstyle\pm0.8}$ & 47.5$_{\scriptstyle\pm13.2}$ & 52.3$_{\scriptstyle\pm2.0}$ & 77.0$_{\scriptstyle\pm0.2}$ & 45.1$_{\scriptstyle\pm15.0}$ & 49.6$_{\scriptstyle\pm2.9}$$^{\dagger}$ & 74.5$_{\scriptstyle\pm0.8}$ \\
Gemma 3 27B & 30.3$_{\scriptstyle\pm0.1}$$^{\dagger}$ & 31.0$_{\scriptstyle\pm0.9}$$^{\dagger}$ & 74.0$_{\scriptstyle\pm0.4}$ & 38.6$_{\scriptstyle\pm0.2}$$^{\dagger}$ & 41.3$_{\scriptstyle\pm1.5}$$^{\dagger}$ & 74.4$_{\scriptstyle\pm0.8}$ & 40.8$_{\scriptstyle\pm0.6}$$^{\dagger}$ & 44.8$_{\scriptstyle\pm2.5}$$^{\dagger}$ & 75.9$_{\scriptstyle\pm1.0}$ & 33.4$_{\scriptstyle\pm0.4}$$^{\dagger}$ & 40.0$_{\scriptstyle\pm3.9}$$^{\dagger}$ & 75.3$_{\scriptstyle\pm1.2}$ \\
Gemma 4 26B A4B & 58.4$_{\scriptstyle\pm0.2}$$^{\dagger}$ & 54.1$_{\scriptstyle\pm0.3}$$^{\dagger}$ & 67.7$_{\scriptstyle\pm1.4}$$^{\dagger}$ & 63.8$_{\scriptstyle\pm0.1}$$^{\dagger}$ & 63.5$_{\scriptstyle\pm0.1}$$^{\dagger}$ & 69.5$_{\scriptstyle\pm0.7}$ & 66.9$_{\scriptstyle\pm0.1}$ & 67.0$_{\scriptstyle\pm0.2}$$^{\dagger}$ & 71.1$_{\scriptstyle\pm0.5}$ & 66.7$_{\scriptstyle\pm0.3}$ & 67.4$_{\scriptstyle\pm0.2}$ & 69.1$_{\scriptstyle\pm0.2}$ \\
Gemma 4 31B & 64.3$_{\scriptstyle\pm0.2}$$^{\dagger}$ & 58.3$_{\scriptstyle\pm0.2}$$^{\dagger}$ & \textbf{76.4$_{\scriptstyle\pm0.4}$$^{\dagger}$} & 67.8$_{\scriptstyle\pm0.0}$$^{\dagger}$ & 66.2$_{\scriptstyle\pm0.3}$$^{\dagger}$ & \textbf{77.0$_{\scriptstyle\pm0.2}$$^{\dagger}$} & 69.5$_{\scriptstyle\pm0.1}$ & 68.4$_{\scriptstyle\pm0.2}$$^{\dagger}$ & \textbf{78.5$_{\scriptstyle\pm0.1}$} & 69.9$_{\scriptstyle\pm0.2}$ & 68.9$_{\scriptstyle\pm0.1}$ & \textbf{77.7$_{\scriptstyle\pm0.3}$} \\
RoMistral 7B & 36.3$_{\scriptstyle\pm14.3}$$^{\dagger}$ & 45.2$_{\scriptstyle\pm2.3}$$^{\dagger}$ & 74.8$_{\scriptstyle\pm0.5}$ & 38.7$_{\scriptstyle\pm14.3}$$^{\dagger}$ & 47.6$_{\scriptstyle\pm1.3}$$^{\dagger}$ & 74.5$_{\scriptstyle\pm0.3}$ & 40.6$_{\scriptstyle\pm12.7}$ & 50.2$_{\scriptstyle\pm0.5}$ & 75.9$_{\scriptstyle\pm0.3}$ & 40.2$_{\scriptstyle\pm13.6}$ & 48.7$_{\scriptstyle\pm1.4}$ & 74.6$_{\scriptstyle\pm0.2}$ \\
RoLlama 3.1 8B & 44.0$_{\scriptstyle\pm1.3}$$^{\dagger}$ & 41.7$_{\scriptstyle\pm0.5}$$^{\dagger}$ & 73.5$_{\scriptstyle\pm0.5}$ & 55.9$_{\scriptstyle\pm0.6}$$^{\dagger}$ & 45.5$_{\scriptstyle\pm1.3}$$^{\dagger}$ & 73.4$_{\scriptstyle\pm0.6}$ & 58.5$_{\scriptstyle\pm0.7}$$^{\dagger}$ & 48.3$_{\scriptstyle\pm2.1}$$^{\dagger}$ & 74.1$_{\scriptstyle\pm0.3}$ & 56.1$_{\scriptstyle\pm0.6}$$^{\dagger}$ & 46.1$_{\scriptstyle\pm1.5}$$^{\dagger}$ & 74.2$_{\scriptstyle\pm0.4}$ \\
RoGemma 2 9B & 42.7$_{\scriptstyle\pm0.7}$$^{\dagger}$ & 42.5$_{\scriptstyle\pm0.5}$$^{\dagger}$ & 75.8$_{\scriptstyle\pm0.1}$ & 53.9$_{\scriptstyle\pm0.5}$$^{\dagger}$ & 55.9$_{\scriptstyle\pm0.9}$$^{\dagger}$ & 75.8$_{\scriptstyle\pm0.2}$ & 56.6$_{\scriptstyle\pm0.6}$$^{\dagger}$ & 60.3$_{\scriptstyle\pm0.9}$$^{\dagger}$ & 75.7$_{\scriptstyle\pm0.2}$ & 54.0$_{\scriptstyle\pm0.4}$$^{\dagger}$ & 58.5$_{\scriptstyle\pm1.6}$$^{\dagger}$ & 75.5$_{\scriptstyle\pm0.3}$ \\
GPT-4.1 mini & 45.0$^{\dagger}$ & 50.8$^{\dagger}$ & - & 56.9$^{\dagger}$ & 60.9$^{\dagger}$ & - & 62.1 & 64.4$^{\dagger}$ & - & 58.3$^{\dagger}$ & 63.2$^{\dagger}$ & - \\
GPT-5 mini & 65.4$^{\dagger}$ & - & - & 69.3$^{\dagger}$ & - & - & 70.6 & - & - & 70.8 & - & - \\
GPT-6 Luna & 64.5$^{\dagger}$ & - & - & 66.6$^{\dagger}$ & - & - & 68.2 & - & - & 68.1 & - & - \\
\midrule
\multicolumn{13}{c}{\textit{RO $\rightarrow$ DE}} \\
\midrule
Llama 3.1 8B & 41.4$_{\scriptstyle\pm17.6}$$^{\dagger}$ & 42.8$_{\scriptstyle\pm2.7}$$^{\dagger}$ & 75.6$_{\scriptstyle\pm1.2}$ & 43.1$_{\scriptstyle\pm17.2}$$^{\dagger}$ & 48.0$_{\scriptstyle\pm2.3}$$^{\dagger}$ & 74.5$_{\scriptstyle\pm1.0}$ & 45.2$_{\scriptstyle\pm14.7}$ & 49.2$_{\scriptstyle\pm2.7}$$^{\dagger}$ & 71.3$_{\scriptstyle\pm0.9}$ & 42.1$_{\scriptstyle\pm15.5}$$^{\dagger}$ & 47.3$_{\scriptstyle\pm2.1}$$^{\dagger}$ & 75.2$_{\scriptstyle\pm0.3}$ \\
Gemma 3 27B & 30.2$_{\scriptstyle\pm0.2}$$^{\dagger}$ & 33.1$_{\scriptstyle\pm0.8}$$^{\dagger}$ & 74.7$_{\scriptstyle\pm0.7}$ & 39.6$_{\scriptstyle\pm0.3}$$^{\dagger}$ & 42.6$_{\scriptstyle\pm1.2}$$^{\dagger}$ & 74.7$_{\scriptstyle\pm0.8}$ & 40.8$_{\scriptstyle\pm0.4}$$^{\dagger}$ & 44.3$_{\scriptstyle\pm0.8}$$^{\dagger}$ & 75.6$_{\scriptstyle\pm1.0}$ & 33.5$_{\scriptstyle\pm0.3}$$^{\dagger}$ & 38.7$_{\scriptstyle\pm0.7}$$^{\dagger}$ & 75.5$_{\scriptstyle\pm0.7}$ \\
Gemma 4 26B A4B & 56.0$_{\scriptstyle\pm0.2}$$^{\dagger}$ & 54.4$_{\scriptstyle\pm0.2}$$^{\dagger}$ & 69.7$_{\scriptstyle\pm0.7}$ & 61.6$_{\scriptstyle\pm0.1}$$^{\dagger}$ & 63.7$_{\scriptstyle\pm0.3}$$^{\dagger}$ & 70.3$_{\scriptstyle\pm0.6}$ & 65.2$_{\scriptstyle\pm0.0}$$^{\dagger}$ & 67.3$_{\scriptstyle\pm0.3}$ & 69.3$_{\scriptstyle\pm0.9}$ & 63.8$_{\scriptstyle\pm0.2}$$^{\dagger}$ & 65.6$_{\scriptstyle\pm0.1}$$^{\dagger}$ & 70.5$_{\scriptstyle\pm0.5}$ \\
Gemma 4 31B & 64.0$_{\scriptstyle\pm0.2}$$^{\dagger}$ & 60.9$_{\scriptstyle\pm0.2}$$^{\dagger}$ & \textbf{77.9$_{\scriptstyle\pm0.3}$} & 67.8$_{\scriptstyle\pm0.1}$$^{\dagger}$ & 66.1$_{\scriptstyle\pm0.1}$$^{\dagger}$ & \textbf{78.1$_{\scriptstyle\pm0.3}$} & 68.9$_{\scriptstyle\pm0.1}$$^{\dagger}$ & 68.2$_{\scriptstyle\pm0.3}$$^{\dagger}$ & \textbf{78.2$_{\scriptstyle\pm0.3}$} & 68.4$_{\scriptstyle\pm0.2}$$^{\dagger}$ & 68.3$_{\scriptstyle\pm0.1}$$^{\dagger}$ & \textbf{76.8$_{\scriptstyle\pm0.1}$$^{\dagger}$} \\
RoMistral 7B & 34.7$_{\scriptstyle\pm14.4}$$^{\dagger}$ & 46.3$_{\scriptstyle\pm0.7}$$^{\dagger}$ & 74.5$_{\scriptstyle\pm1.0}$ & 36.9$_{\scriptstyle\pm14.6}$$^{\dagger}$ & 48.2$_{\scriptstyle\pm0.8}$$^{\dagger}$ & 72.3$_{\scriptstyle\pm1.3}$ & 37.5$_{\scriptstyle\pm13.4}$$^{\dagger}$ & 49.7$_{\scriptstyle\pm0.8}$ & 73.2$_{\scriptstyle\pm2.0}$ & 36.8$_{\scriptstyle\pm14.3}$$^{\dagger}$ & 48.9$_{\scriptstyle\pm1.0}$ & 74.4$_{\scriptstyle\pm0.0}$ \\
RoLlama 3.1 8B & 41.4$_{\scriptstyle\pm1.2}$$^{\dagger}$ & 41.7$_{\scriptstyle\pm0.8}$$^{\dagger}$ & 73.8$_{\scriptstyle\pm0.4}$ & 51.9$_{\scriptstyle\pm0.8}$$^{\dagger}$ & 45.8$_{\scriptstyle\pm1.9}$$^{\dagger}$ & 72.4$_{\scriptstyle\pm0.4}$ & 54.8$_{\scriptstyle\pm0.9}$$^{\dagger}$ & 47.2$_{\scriptstyle\pm1.4}$$^{\dagger}$ & 72.6$_{\scriptstyle\pm0.3}$ & 50.7$_{\scriptstyle\pm1.1}$$^{\dagger}$ & 44.9$_{\scriptstyle\pm0.9}$$^{\dagger}$ & 74.1$_{\scriptstyle\pm0.3}$ \\
RoGemma 2 9B & 49.3$_{\scriptstyle\pm0.7}$$^{\dagger}$ & 56.8$_{\scriptstyle\pm1.4}$$^{\dagger}$ & 75.4$_{\scriptstyle\pm0.2}$ & 54.6$_{\scriptstyle\pm0.7}$$^{\dagger}$ & 60.2$_{\scriptstyle\pm0.8}$$^{\dagger}$ & 75.2$_{\scriptstyle\pm0.3}$ & 58.6$_{\scriptstyle\pm0.5}$$^{\dagger}$ & 62.7$_{\scriptstyle\pm0.8}$ & 75.0$_{\scriptstyle\pm0.6}$ & 54.4$_{\scriptstyle\pm0.5}$$^{\dagger}$ & 61.3$_{\scriptstyle\pm1.1}$$^{\dagger}$ & 75.6$_{\scriptstyle\pm0.1}$ \\
GPT-4.1 mini & 48.0$^{\dagger}$ & 55.4$^{\dagger}$ & - & 54.4$^{\dagger}$ & 62.4$^{\dagger}$ & - & 59.4$^{\dagger}$ & 65.6$^{\dagger}$ & - & 52.9$^{\dagger}$ & 62.9$^{\dagger}$ & - \\
GPT-5 mini & 65.4$^{\dagger}$ & - & - & 70.2 & - & - & 70.2 & - & - & 69.7$^{\dagger}$ & - & - \\
GPT-6 Luna & 63.4$^{\dagger}$ & - & - & 65.8$^{\dagger}$ & - & - & 67.4 & - & - & 66.2$^{\dagger}$ & - & - \\
\midrule
\multicolumn{13}{c}{\textit{RO $\rightarrow$ FR}} \\
\midrule
Llama 3.1 8B & 40.1$_{\scriptstyle\pm21.0}$$^{\dagger}$ & 44.2$_{\scriptstyle\pm2.4}$$^{\dagger}$ & \textbf{76.6$_{\scriptstyle\pm0.8}$} & 43.5$_{\scriptstyle\pm17.9}$$^{\dagger}$ & 49.5$_{\scriptstyle\pm1.5}$$^{\dagger}$ & 75.0$_{\scriptstyle\pm1.2}$ & 44.7$_{\scriptstyle\pm17.4}$$^{\dagger}$ & 50.5$_{\scriptstyle\pm1.5}$$^{\dagger}$ & 74.4$_{\scriptstyle\pm3.2}$ & 42.1$_{\scriptstyle\pm18.0}$$^{\dagger}$ & 48.2$_{\scriptstyle\pm2.1}$$^{\dagger}$ & 75.7$_{\scriptstyle\pm0.1}$ \\
Gemma 3 27B & 31.2$_{\scriptstyle\pm0.1}$$^{\dagger}$ & 34.9$_{\scriptstyle\pm1.7}$$^{\dagger}$ & 75.2$_{\scriptstyle\pm1.2}$ & 40.6$_{\scriptstyle\pm0.4}$$^{\dagger}$ & 44.4$_{\scriptstyle\pm2.3}$$^{\dagger}$ & 76.0$_{\scriptstyle\pm0.6}$ & 42.2$_{\scriptstyle\pm0.3}$$^{\dagger}$ & 46.4$_{\scriptstyle\pm1.7}$$^{\dagger}$ & 75.6$_{\scriptstyle\pm1.0}$ & 34.6$_{\scriptstyle\pm0.3}$$^{\dagger}$ & 41.0$_{\scriptstyle\pm3.2}$$^{\dagger}$ & 74.7$_{\scriptstyle\pm0.6}$ \\
Gemma 4 26B A4B & 58.4$_{\scriptstyle\pm0.1}$$^{\dagger}$ & 56.2$_{\scriptstyle\pm0.4}$$^{\dagger}$ & 68.9$_{\scriptstyle\pm0.9}$$^{\dagger}$ & 64.3$_{\scriptstyle\pm0.1}$$^{\dagger}$ & 65.3$_{\scriptstyle\pm0.2}$$^{\dagger}$ & 70.4$_{\scriptstyle\pm0.6}$ & 65.8$_{\scriptstyle\pm0.2}$$^{\dagger}$ & 67.3$_{\scriptstyle\pm0.1}$ & 71.6$_{\scriptstyle\pm0.5}$ & 65.1$_{\scriptstyle\pm0.1}$$^{\dagger}$ & 66.6$_{\scriptstyle\pm0.3}$$^{\dagger}$ & 71.0$_{\scriptstyle\pm0.5}$ \\
Gemma 4 31B & 66.8$_{\scriptstyle\pm0.0}$$^{\dagger}$ & 62.4$_{\scriptstyle\pm0.3}$$^{\dagger}$ & 76.5$_{\scriptstyle\pm0.4}$$^{\dagger}$ & 69.1$_{\scriptstyle\pm0.1}$ & 67.1$_{\scriptstyle\pm0.2}$$^{\dagger}$ & \textbf{77.9$_{\scriptstyle\pm0.3}$} & 69.5$_{\scriptstyle\pm0.1}$ & 68.9$_{\scriptstyle\pm0.2}$ & \textbf{77.2$_{\scriptstyle\pm0.3}$$^{\dagger}$} & 68.6$_{\scriptstyle\pm0.1}$$^{\dagger}$ & 68.2$_{\scriptstyle\pm0.1}$$^{\dagger}$ & \textbf{77.3$_{\scriptstyle\pm0.4}$} \\
RoMistral 7B & 34.2$_{\scriptstyle\pm14.8}$$^{\dagger}$ & 46.0$_{\scriptstyle\pm1.4}$$^{\dagger}$ & 75.0$_{\scriptstyle\pm0.2}$ & 36.5$_{\scriptstyle\pm15.1}$$^{\dagger}$ & 48.1$_{\scriptstyle\pm1.9}$$^{\dagger}$ & 74.7$_{\scriptstyle\pm0.5}$ & 37.5$_{\scriptstyle\pm14.5}$$^{\dagger}$ & 48.5$_{\scriptstyle\pm1.3}$ & 74.2$_{\scriptstyle\pm1.1}$ & 36.0$_{\scriptstyle\pm15.2}$$^{\dagger}$ & 47.3$_{\scriptstyle\pm1.1}$$^{\dagger}$ & 74.8$_{\scriptstyle\pm0.4}$ \\
RoLlama 3.1 8B & 47.1$_{\scriptstyle\pm1.2}$$^{\dagger}$ & 44.2$_{\scriptstyle\pm1.0}$$^{\dagger}$ & 73.6$_{\scriptstyle\pm0.4}$ & 55.5$_{\scriptstyle\pm1.0}$$^{\dagger}$ & 47.5$_{\scriptstyle\pm1.0}$$^{\dagger}$ & 73.9$_{\scriptstyle\pm0.3}$ & 57.5$_{\scriptstyle\pm0.9}$$^{\dagger}$ & 48.7$_{\scriptstyle\pm1.0}$$^{\dagger}$ & 73.0$_{\scriptstyle\pm0.3}$ & 53.5$_{\scriptstyle\pm0.8}$$^{\dagger}$ & 47.2$_{\scriptstyle\pm0.7}$$^{\dagger}$ & 72.7$_{\scriptstyle\pm0.3}$ \\
RoGemma 2 9B & 51.9$_{\scriptstyle\pm0.9}$$^{\dagger}$ & 58.8$_{\scriptstyle\pm1.5}$$^{\dagger}$ & 75.3$_{\scriptstyle\pm0.1}$ & 57.1$_{\scriptstyle\pm1.1}$$^{\dagger}$ & 61.4$_{\scriptstyle\pm0.6}$$^{\dagger}$ & 75.6$_{\scriptstyle\pm0.2}$ & 58.1$_{\scriptstyle\pm0.6}$$^{\dagger}$ & 63.3$_{\scriptstyle\pm0.7}$ & 75.5$_{\scriptstyle\pm0.4}$ & 55.4$_{\scriptstyle\pm0.4}$$^{\dagger}$ & 62.1$_{\scriptstyle\pm1.3}$ & 74.9$_{\scriptstyle\pm0.3}$ \\
GPT-4.1 mini & 49.6$^{\dagger}$ & 56.2$^{\dagger}$ & - & 56.6$^{\dagger}$ & 62.9$^{\dagger}$ & - & 60.3$^{\dagger}$ & 64.4$^{\dagger}$ & - & 56.3$^{\dagger}$ & 62.5$^{\dagger}$ & - \\
GPT-5 mini & 68.2$^{\dagger}$ & - & - & 70.5 & - & - & 70.4 & - & - & 70.0$^{\dagger}$ & - & - \\
GPT-6 Luna & 67.0 & - & - & 67.0 & - & - & 68.5 & - & - & 66.8$^{\dagger}$ & - & - \\
\midrule
\multicolumn{13}{c}{\textit{RO $\rightarrow$ IT}} \\
\midrule
Llama 3.1 8B & 47.4$_{\scriptstyle\pm10.3}$ & 44.3$_{\scriptstyle\pm2.2}$$^{\dagger}$ & 70.7$_{\scriptstyle\pm2.5}$ & 46.9$_{\scriptstyle\pm13.6}$ & 49.8$_{\scriptstyle\pm2.4}$$^{\dagger}$ & 74.2$_{\scriptstyle\pm1.2}$ & 48.1$_{\scriptstyle\pm12.1}$ & 50.1$_{\scriptstyle\pm2.1}$$^{\dagger}$ & 75.8$_{\scriptstyle\pm2.9}$ & 46.3$_{\scriptstyle\pm13.2}$ & 46.3$_{\scriptstyle\pm2.0}$$^{\dagger}$ & 74.7$_{\scriptstyle\pm1.3}$ \\
Gemma 3 27B & 32.6$_{\scriptstyle\pm0.3}$$^{\dagger}$ & 38.9$_{\scriptstyle\pm2.4}$$^{\dagger}$ & 75.7$_{\scriptstyle\pm0.6}$ & 42.6$_{\scriptstyle\pm0.3}$$^{\dagger}$ & 45.6$_{\scriptstyle\pm1.6}$$^{\dagger}$ & 76.7$_{\scriptstyle\pm0.8}$ & 43.9$_{\scriptstyle\pm0.3}$$^{\dagger}$ & 48.2$_{\scriptstyle\pm1.4}$$^{\dagger}$ & 75.4$_{\scriptstyle\pm1.2}$ & 35.2$_{\scriptstyle\pm0.4}$$^{\dagger}$ & 42.7$_{\scriptstyle\pm2.5}$$^{\dagger}$ & 76.4$_{\scriptstyle\pm1.2}$ \\
Gemma 4 26B A4B & 63.4$_{\scriptstyle\pm0.1}$$^{\dagger}$ & 64.0$_{\scriptstyle\pm0.1}$$^{\dagger}$ & 70.8$_{\scriptstyle\pm0.4}$ & 65.6$_{\scriptstyle\pm0.3}$$^{\dagger}$ & 67.3$_{\scriptstyle\pm0.1}$ & 69.8$_{\scriptstyle\pm0.3}$ & 66.9$_{\scriptstyle\pm0.2}$ & 69.1$_{\scriptstyle\pm0.2}$ & 72.2$_{\scriptstyle\pm0.4}$ & 65.8$_{\scriptstyle\pm0.1}$ & 67.8$_{\scriptstyle\pm0.2}$ & 71.2$_{\scriptstyle\pm0.8}$ \\
Gemma 4 31B & 67.5$_{\scriptstyle\pm0.1}$$^{\dagger}$ & 66.2$_{\scriptstyle\pm0.2}$$^{\dagger}$ & \textbf{77.3$_{\scriptstyle\pm0.3}$} & 68.9$_{\scriptstyle\pm0.0}$$^{\dagger}$ & 68.2$_{\scriptstyle\pm0.1}$$^{\dagger}$ & \textbf{77.7$_{\scriptstyle\pm0.3}$} & 69.9$_{\scriptstyle\pm0.1}$ & 69.0$_{\scriptstyle\pm0.1}$ & \textbf{77.7$_{\scriptstyle\pm0.6}$} & 69.1$_{\scriptstyle\pm0.1}$ & 68.4$_{\scriptstyle\pm0.1}$ & \textbf{77.9$_{\scriptstyle\pm0.1}$} \\
RoMistral 7B & 33.5$_{\scriptstyle\pm15.1}$$^{\dagger}$ & 48.7$_{\scriptstyle\pm2.0}$ & 74.0$_{\scriptstyle\pm2.0}$ & 36.8$_{\scriptstyle\pm15.0}$$^{\dagger}$ & 49.0$_{\scriptstyle\pm0.9}$ & 74.4$_{\scriptstyle\pm0.3}$ & 37.1$_{\scriptstyle\pm14.8}$$^{\dagger}$ & 50.3$_{\scriptstyle\pm1.5}$ & 75.5$_{\scriptstyle\pm0.1}$ & 33.7$_{\scriptstyle\pm14.4}$$^{\dagger}$ & 48.9$_{\scriptstyle\pm0.8}$ & 75.7$_{\scriptstyle\pm0.3}$ \\
RoLlama 3.1 8B & 46.2$_{\scriptstyle\pm1.0}$$^{\dagger}$ & 44.1$_{\scriptstyle\pm1.3}$$^{\dagger}$ & 73.7$_{\scriptstyle\pm0.8}$ & 55.6$_{\scriptstyle\pm0.9}$$^{\dagger}$ & 47.1$_{\scriptstyle\pm0.9}$$^{\dagger}$ & 73.7$_{\scriptstyle\pm0.8}$ & 56.1$_{\scriptstyle\pm1.0}$$^{\dagger}$ & 47.5$_{\scriptstyle\pm1.6}$$^{\dagger}$ & 72.6$_{\scriptstyle\pm0.4}$ & 51.3$_{\scriptstyle\pm1.3}$$^{\dagger}$ & 46.7$_{\scriptstyle\pm0.0}$$^{\dagger}$ & 73.6$_{\scriptstyle\pm0.7}$ \\
RoGemma 2 9B & 55.9$_{\scriptstyle\pm1.2}$$^{\dagger}$ & 61.4$_{\scriptstyle\pm0.5}$ & 75.7$_{\scriptstyle\pm0.2}$ & 57.6$_{\scriptstyle\pm0.8}$$^{\dagger}$ & 62.4$_{\scriptstyle\pm0.8}$ & 76.2$_{\scriptstyle\pm0.2}$ & 58.9$_{\scriptstyle\pm0.7}$$^{\dagger}$ & 63.7$_{\scriptstyle\pm0.8}$ & 74.9$_{\scriptstyle\pm0.2}$ & 57.8$_{\scriptstyle\pm0.7}$$^{\dagger}$ & 63.1$_{\scriptstyle\pm0.3}$ & 75.1$_{\scriptstyle\pm0.3}$ \\
GPT-4.1 mini & 54.2$^{\dagger}$ & 61.8$^{\dagger}$ & - & 59.1$^{\dagger}$ & 64.3$^{\dagger}$ & - & 61.4$^{\dagger}$ & 65.9$^{\dagger}$ & - & 57.8$^{\dagger}$ & 64.2$^{\dagger}$ & - \\
GPT-5 mini & 68.8$^{\dagger}$ & - & - & 70.5 & - & - & 70.8 & - & - & 71.2 & - & - \\
GPT-6 Luna & 67.0 & - & - & 68.4 & - & - & 69.1 & - & - & 67.7 & - & - \\
\bottomrule
\end{tabular}}
\caption{F1-scores breakdown for each model and code-switching generation method, in the zero-shot (ZS), chain-of-thought (CoT), and fine-tuning (FT) settings, for CS-SeLeRoSa test set. Bold indicates the best performance for each method and language pair. $^{\dagger}$ indicates statistically significant degradation compared to the monolingual counterpart (original Romanian input with the same model for ZS/CoT; the Romanian-only fine-tuned counterpart for FT; sentence-level paired bootstrap, $p < 0.05$).}
\label{tab:cs_results}
\end{table*}

\begin{table*}[!th]
\centering
\small
\setlength{\tabcolsep}{3pt}
\resizebox{\textwidth}{!}{
\begin{tabular}{lcccccccccccc}
\toprule
\multicolumn{1}{c}{\multirow{2}{*}{\textbf{Model}}} & \multicolumn{3}{c}{\textbf{Random}} & \multicolumn{3}{c}{\textbf{Phrase}} & \multicolumn{3}{c}{\textbf{Dependency}} & \multicolumn{3}{c}{\textbf{LLM}} \\
 & \textbf{ZS} & \textbf{CoT} & \textbf{FT} & \textbf{ZS} & \textbf{CoT} & \textbf{FT} & \textbf{ZS} & \textbf{CoT} & \textbf{FT} & \textbf{ZS} & \textbf{CoT} & \textbf{FT} \\
\midrule
\multicolumn{13}{c}{\textit{RO $\rightarrow$ EN}} \\
\midrule
Llama 3.1 8B & 50.0$_{\scriptstyle\pm16.7}$ & 31.7$_{\scriptstyle\pm16.1}$ & \textbf{73.3$_{\scriptstyle\pm0.0}$} & 50.3$_{\scriptstyle\pm13.1}$ & 48.7$_{\scriptstyle\pm7.5}$ & 68.6$_{\scriptstyle\pm1.8}$ & 44.6$_{\scriptstyle\pm18.4}$ & 46.9$_{\scriptstyle\pm2.6}$ & 63.6$_{\scriptstyle\pm2.2}$ & 50.7$_{\scriptstyle\pm11.1}$ & 46.7$_{\scriptstyle\pm5.3}$ & 61.7$_{\scriptstyle\pm3.6}$ \\
Gemma 3 27B & 28.9$_{\scriptstyle\pm7.7}$ & 33.3$_{\scriptstyle\pm0.0}$ & \textbf{73.3$_{\scriptstyle\pm0.0}$} & 33.2$_{\scriptstyle\pm0.0}$ & 38.3$_{\scriptstyle\pm1.7}$ & 62.8$_{\scriptstyle\pm3.5}$ & 33.6$_{\scriptstyle\pm1.7}$ & 35.0$_{\scriptstyle\pm4.1}$ & 69.2$_{\scriptstyle\pm3.9}$ & 33.6$_{\scriptstyle\pm0.0}$ & 38.4$_{\scriptstyle\pm6.8}$ & 61.6$_{\scriptstyle\pm2.9}$ \\
Gemma 4 26B A4B & 57.8$_{\scriptstyle\pm13.5}$ & 50.0$_{\scriptstyle\pm0.0}$ & 65.6$_{\scriptstyle\pm13.5}$ & 64.0$_{\scriptstyle\pm0.0}$ & 61.3$_{\scriptstyle\pm6.1}$ & 70.3$_{\scriptstyle\pm2.1}$ & 54.7$_{\scriptstyle\pm2.9}$ & 54.2$_{\scriptstyle\pm2.7}$ & 58.8$_{\scriptstyle\pm3.5}$ & 65.2$_{\scriptstyle\pm1.3}$ & 65.1$_{\scriptstyle\pm2.5}$ & 63.7$_{\scriptstyle\pm1.3}$ \\
Gemma 4 31B & 50.0$_{\scriptstyle\pm0.0}$ & 50.0$_{\scriptstyle\pm0.0}$ & \textbf{73.3$_{\scriptstyle\pm0.0}$} & 70.3$_{\scriptstyle\pm0.0}$ & 67.9$_{\scriptstyle\pm2.1}$ & 72.5$_{\scriptstyle\pm2.2}$ & 59.7$_{\scriptstyle\pm1.2}$ & 60.4$_{\scriptstyle\pm1.2}$ & 72.3$_{\scriptstyle\pm2.1}$ & 64.7$_{\scriptstyle\pm1.2}$ & 66.1$_{\scriptstyle\pm2.1}$ & \textbf{72.5$_{\scriptstyle\pm3.5}$} \\
RoMistral 7B & 17.8$_{\scriptstyle\pm16.8}$ & 26.1$_{\scriptstyle\pm6.7}$ & \textbf{73.3$_{\scriptstyle\pm0.0}$} & 40.3$_{\scriptstyle\pm13.3}$ & 53.0$_{\scriptstyle\pm4.7}$ & 67.0$_{\scriptstyle\pm0.5}$ & 39.0$_{\scriptstyle\pm9.5}$ & 45.6$_{\scriptstyle\pm8.3}$ & \textbf{73.1$_{\scriptstyle\pm0.7}$} & 36.7$_{\scriptstyle\pm14.8}$ & 52.2$_{\scriptstyle\pm6.7}$ & 71.5$_{\scriptstyle\pm4.4}$ \\
RoLlama 3.1 8B & 50.0$_{\scriptstyle\pm0.0}$ & 47.8$_{\scriptstyle\pm26.7}$ & \textbf{73.3$_{\scriptstyle\pm0.0}$} & 61.8$_{\scriptstyle\pm4.1}$ & 56.9$_{\scriptstyle\pm4.8}$ & \textbf{73.7$_{\scriptstyle\pm5.0}$} & 51.4$_{\scriptstyle\pm3.6}$ & 40.8$_{\scriptstyle\pm2.6}$ & 60.5$_{\scriptstyle\pm1.1}$ & 57.9$_{\scriptstyle\pm1.6}$ & 47.5$_{\scriptstyle\pm5.0}$ & 69.3$_{\scriptstyle\pm1.1}$ \\
RoGemma 2 9B & 20.0$_{\scriptstyle\pm0.0}$ & 24.4$_{\scriptstyle\pm7.7}$ & \textbf{73.3$_{\scriptstyle\pm0.0}$} & 54.2$_{\scriptstyle\pm2.6}$ & 61.3$_{\scriptstyle\pm2.3}$ & 69.4$_{\scriptstyle\pm2.4}$ & 56.5$_{\scriptstyle\pm1.3}$ & 60.3$_{\scriptstyle\pm9.9}$ & 66.8$_{\scriptstyle\pm4.2}$ & 59.0$_{\scriptstyle\pm4.7}$ & 60.9$_{\scriptstyle\pm4.7}$ & 69.2$_{\scriptstyle\pm4.4}$ \\
GPT-4.1 mini & 30.0$_{\scriptstyle\pm17.3}$ & 20.0 & - & 52.5$_{\scriptstyle\pm2.0}$ & 64.0 & - & 51.0$_{\scriptstyle\pm2.3}$ & 58.1 & - & 53.9$_{\scriptstyle\pm1.3}$ & 55.5 & - \\
GPT-5 mini & 47.8$_{\scriptstyle\pm26.7}$ & - & - & 72.9$_{\scriptstyle\pm2.4}$ & - & - & 66.4$_{\scriptstyle\pm1.5}$ & - & - & 62.0$_{\scriptstyle\pm1.0}$ & - & - \\
GPT-5 & 50.0$_{\scriptstyle\pm0.0}$ & - & - & 70.5$_{\scriptstyle\pm2.2}$ & - & - & 68.7$_{\scriptstyle\pm3.7}$ & - & - & 70.3$_{\scriptstyle\pm1.9}$ & - & - \\
GPT-6 Luna & 37.8$_{\scriptstyle\pm30.8}$ & - & - & 55.9$_{\scriptstyle\pm0.2}$ & - & - & 52.6$_{\scriptstyle\pm5.3}$ & - & - & 60.7$_{\scriptstyle\pm1.3}$ & - & - \\
GPT-6 Astra & 50.0$_{\scriptstyle\pm0.0}$ & - & - & 64.0$_{\scriptstyle\pm0.0}$ & - & - & 59.5$_{\scriptstyle\pm2.7}$ & - & - & 62.1$_{\scriptstyle\pm3.8}$ & - & - \\
GPT-6 Sol & 40.0$_{\scriptstyle\pm17.3}$ & - & - & 59.9$_{\scriptstyle\pm0.0}$ & - & - & 60.2$_{\scriptstyle\pm3.6}$ & - & - & 54.0$_{\scriptstyle\pm1.3}$ & - & - \\
GPT-6.1 Sol & 47.8$_{\scriptstyle\pm26.7}$ & - & - & 66.6$_{\scriptstyle\pm2.3}$ & - & - & 61.9$_{\scriptstyle\pm1.3}$ & - & - & 64.3$_{\scriptstyle\pm2.1}$ & - & - \\
\midrule
\multicolumn{13}{c}{\textit{RO $\rightarrow$ DE}} \\
\midrule
Llama 3.1 8B & 33.3$_{\scriptstyle\pm28.9}$ & 0.0$_{\scriptstyle\pm0.0}$ & 0.0$_{\scriptstyle\pm0.0}$ & 36.4$_{\scriptstyle\pm23.6}$ & 26.6$_{\scriptstyle\pm20.3}$ & 51.6$_{\scriptstyle\pm3.5}$ & 38.5$_{\scriptstyle\pm15.0}$ & 33.5$_{\scriptstyle\pm4.1}$ & 56.8$_{\scriptstyle\pm6.4}$ & 50.0$_{\scriptstyle\pm14.9}$ & 53.0$_{\scriptstyle\pm10.3}$ & 71.8$_{\scriptstyle\pm3.4}$ \\
Gemma 3 27B & \textbf{50.0$_{\scriptstyle\pm0.0}$} & 33.3$_{\scriptstyle\pm28.9}$ & 0.0$_{\scriptstyle\pm0.0}$ & 38.0$_{\scriptstyle\pm5.7}$ & 31.4$_{\scriptstyle\pm0.0}$ & \textbf{72.4$_{\scriptstyle\pm4.6}$} & 34.5$_{\scriptstyle\pm0.0}$ & 32.2$_{\scriptstyle\pm3.9}$ & 62.9$_{\scriptstyle\pm2.0}$ & 39.1$_{\scriptstyle\pm3.2}$ & 38.4$_{\scriptstyle\pm2.7}$ & 75.0$_{\scriptstyle\pm1.8}$ \\
Gemma 4 26B A4B & 0.0$_{\scriptstyle\pm0.0}$ & 0.0$_{\scriptstyle\pm0.0}$ & 0.0$_{\scriptstyle\pm0.0}$ & 55.6$_{\scriptstyle\pm0.0}$ & 55.9$_{\scriptstyle\pm6.5}$ & 52.4$_{\scriptstyle\pm15.0}$ & 53.6$_{\scriptstyle\pm0.0}$ & 55.2$_{\scriptstyle\pm4.2}$ & 58.3$_{\scriptstyle\pm3.5}$ & 60.9$_{\scriptstyle\pm0.0}$ & 69.2$_{\scriptstyle\pm1.7}$ & 64.5$_{\scriptstyle\pm3.2}$ \\
Gemma 4 31B & 0.0$_{\scriptstyle\pm0.0}$ & 0.0$_{\scriptstyle\pm0.0}$ & 0.0$_{\scriptstyle\pm0.0}$ & 49.6$_{\scriptstyle\pm0.0}$ & 55.9$_{\scriptstyle\pm6.5}$ & 70.0$_{\scriptstyle\pm0.0}$ & 51.1$_{\scriptstyle\pm2.1}$ & 57.4$_{\scriptstyle\pm3.7}$ & \textbf{73.3$_{\scriptstyle\pm5.5}$} & 62.3$_{\scriptstyle\pm1.8}$ & 66.4$_{\scriptstyle\pm1.8}$ & 63.1$_{\scriptstyle\pm1.9}$ \\
RoMistral 7B & 0.0$_{\scriptstyle\pm0.0}$ & 16.7$_{\scriptstyle\pm28.9}$ & 0.0$_{\scriptstyle\pm0.0}$ & 28.3$_{\scriptstyle\pm8.7}$ & 39.7$_{\scriptstyle\pm16.5}$ & 44.1$_{\scriptstyle\pm10.0}$ & 37.0$_{\scriptstyle\pm18.3}$ & 37.3$_{\scriptstyle\pm7.1}$ & 59.1$_{\scriptstyle\pm2.7}$ & 35.0$_{\scriptstyle\pm14.9}$ & 55.1$_{\scriptstyle\pm5.0}$ & \textbf{81.7$_{\scriptstyle\pm1.8}$} \\
RoLlama 3.1 8B & 0.0$_{\scriptstyle\pm0.0}$ & 33.3$_{\scriptstyle\pm28.9}$ & 0.0$_{\scriptstyle\pm0.0}$ & 46.9$_{\scriptstyle\pm9.7}$ & 28.8$_{\scriptstyle\pm7.8}$ & 45.4$_{\scriptstyle\pm9.5}$ & 51.9$_{\scriptstyle\pm1.4}$ & 45.9$_{\scriptstyle\pm4.9}$ & 59.1$_{\scriptstyle\pm2.6}$ & 54.6$_{\scriptstyle\pm7.0}$ & 40.8$_{\scriptstyle\pm4.8}$ & 73.1$_{\scriptstyle\pm4.6}$ \\
RoGemma 2 9B & 0.0$_{\scriptstyle\pm0.0}$ & 0.0$_{\scriptstyle\pm0.0}$ & 0.0$_{\scriptstyle\pm0.0}$ & 54.5$_{\scriptstyle\pm4.1}$ & 45.6$_{\scriptstyle\pm6.8}$ & 62.1$_{\scriptstyle\pm0.0}$ & 67.9$_{\scriptstyle\pm2.3}$ & 59.5$_{\scriptstyle\pm4.3}$ & 63.6$_{\scriptstyle\pm3.5}$ & 67.7$_{\scriptstyle\pm8.6}$ & 72.0$_{\scriptstyle\pm1.9}$ & 74.3$_{\scriptstyle\pm1.3}$ \\
GPT-4.1 mini & 16.7$_{\scriptstyle\pm28.9}$ & 0.0 & - & 41.2$_{\scriptstyle\pm7.6}$ & 49.6 & - & 45.7$_{\scriptstyle\pm6.1}$ & 64.9 & - & 51.7$_{\scriptstyle\pm3.2}$ & 67.7 & - \\
GPT-5 mini & 0.0$_{\scriptstyle\pm0.0}$ & - & - & 44.0$_{\scriptstyle\pm4.8}$ & - & - & 42.7$_{\scriptstyle\pm5.7}$ & - & - & 56.1$_{\scriptstyle\pm8.0}$ & - & - \\
GPT-5 & 0.0$_{\scriptstyle\pm0.0}$ & - & - & 35.3$_{\scriptstyle\pm5.2}$ & - & - & 37.3$_{\scriptstyle\pm3.1}$ & - & - & 52.4$_{\scriptstyle\pm3.7}$ & - & - \\
GPT-6 Luna & 0.0$_{\scriptstyle\pm0.0}$ & - & - & 42.3$_{\scriptstyle\pm16.6}$ & - & - & 48.1$_{\scriptstyle\pm11.5}$ & - & - & 61.6$_{\scriptstyle\pm1.4}$ & - & - \\
GPT-6 Astra & 0.0$_{\scriptstyle\pm0.0}$ & - & - & 31.9$_{\scriptstyle\pm10.3}$ & - & - & 35.0$_{\scriptstyle\pm1.6}$ & - & - & 57.2$_{\scriptstyle\pm3.1}$ & - & - \\
GPT-6 Sol & 0.0$_{\scriptstyle\pm0.0}$ & - & - & 20.0$_{\scriptstyle\pm0.0}$ & - & - & 37.8$_{\scriptstyle\pm5.2}$ & - & - & 55.8$_{\scriptstyle\pm3.6}$ & - & - \\
GPT-6.1 Sol & 0.0$_{\scriptstyle\pm0.0}$ & - & - & 23.8$_{\scriptstyle\pm6.6}$ & - & - & 38.1$_{\scriptstyle\pm6.8}$ & - & - & 55.4$_{\scriptstyle\pm0.0}$ & - & - \\
\midrule
\multicolumn{13}{c}{\textit{RO $\rightarrow$ FR}} \\
\midrule
Llama 3.1 8B & 15.3$_{\scriptstyle\pm13.2}$ & 28.5$_{\scriptstyle\pm9.0}$ & \textbf{78.7$_{\scriptstyle\pm6.4}$} & 37.7$_{\scriptstyle\pm17.0}$ & 42.1$_{\scriptstyle\pm8.0}$ & 74.1$_{\scriptstyle\pm2.4}$ & 42.6$_{\scriptstyle\pm23.0}$ & 40.1$_{\scriptstyle\pm5.8}$ & 63.1$_{\scriptstyle\pm2.7}$ & 36.9$_{\scriptstyle\pm7.6}$ & 41.1$_{\scriptstyle\pm4.4}$ & 66.2$_{\scriptstyle\pm1.5}$ \\
Gemma 3 27B & 25.0$_{\scriptstyle\pm0.0}$ & 20.5$_{\scriptstyle\pm3.9}$ & 61.6$_{\scriptstyle\pm5.7}$ & 30.3$_{\scriptstyle\pm1.5}$ & 37.4$_{\scriptstyle\pm7.7}$ & 75.2$_{\scriptstyle\pm2.7}$ & 34.3$_{\scriptstyle\pm1.0}$ & 40.8$_{\scriptstyle\pm5.5}$ & 64.8$_{\scriptstyle\pm5.7}$ & 31.8$_{\scriptstyle\pm2.1}$ & 38.3$_{\scriptstyle\pm9.1}$ & 67.4$_{\scriptstyle\pm2.3}$ \\
Gemma 4 26B A4B & 32.5$_{\scriptstyle\pm0.0}$ & 44.4$_{\scriptstyle\pm0.0}$ & 55.0$_{\scriptstyle\pm0.0}$ & 49.9$_{\scriptstyle\pm0.0}$ & 58.6$_{\scriptstyle\pm5.7}$ & 62.9$_{\scriptstyle\pm4.0}$ & 61.5$_{\scriptstyle\pm0.0}$ & 69.0$_{\scriptstyle\pm3.4}$ & 67.6$_{\scriptstyle\pm1.3}$ & 57.4$_{\scriptstyle\pm1.4}$ & 65.9$_{\scriptstyle\pm0.0}$ & 69.4$_{\scriptstyle\pm1.2}$ \\
Gemma 4 31B & 44.4$_{\scriptstyle\pm0.0}$ & 32.5$_{\scriptstyle\pm0.0}$ & 60.0$_{\scriptstyle\pm8.6}$ & 70.9$_{\scriptstyle\pm2.1}$ & 70.6$_{\scriptstyle\pm5.9}$ & \textbf{83.8$_{\scriptstyle\pm0.0}$} & 63.5$_{\scriptstyle\pm0.0}$ & 64.4$_{\scriptstyle\pm1.6}$ & 72.2$_{\scriptstyle\pm1.5}$ & 62.5$_{\scriptstyle\pm1.4}$ & 63.2$_{\scriptstyle\pm4.8}$ & 69.0$_{\scriptstyle\pm1.2}$ \\
RoMistral 7B & 32.1$_{\scriptstyle\pm17.8}$ & 44.2$_{\scriptstyle\pm2.6}$ & 75.0$_{\scriptstyle\pm0.0}$ & 22.8$_{\scriptstyle\pm7.9}$ & 37.9$_{\scriptstyle\pm4.5}$ & 72.7$_{\scriptstyle\pm1.9}$ & 37.3$_{\scriptstyle\pm12.0}$ & 45.7$_{\scriptstyle\pm5.1}$ & \textbf{75.2$_{\scriptstyle\pm1.7}$} & 32.6$_{\scriptstyle\pm16.5}$ & 42.0$_{\scriptstyle\pm3.4}$ & 69.0$_{\scriptstyle\pm1.2}$ \\
RoLlama 3.1 8B & 36.5$_{\scriptstyle\pm6.9}$ & 30.7$_{\scriptstyle\pm11.8}$ & 75.0$_{\scriptstyle\pm0.0}$ & 57.8$_{\scriptstyle\pm1.7}$ & 56.6$_{\scriptstyle\pm1.7}$ & 71.3$_{\scriptstyle\pm2.0}$ & 53.0$_{\scriptstyle\pm3.6}$ & 45.5$_{\scriptstyle\pm3.4}$ & 64.1$_{\scriptstyle\pm5.3}$ & 50.5$_{\scriptstyle\pm5.1}$ & 48.1$_{\scriptstyle\pm4.7}$ & 64.4$_{\scriptstyle\pm5.8}$ \\
RoGemma 2 9B & 36.5$_{\scriptstyle\pm6.9}$ & 50.0$_{\scriptstyle\pm7.3}$ & \textbf{78.7$_{\scriptstyle\pm6.4}$} & 35.2$_{\scriptstyle\pm2.5}$ & 59.1$_{\scriptstyle\pm8.2}$ & 72.9$_{\scriptstyle\pm4.7}$ & 58.1$_{\scriptstyle\pm5.9}$ & 63.2$_{\scriptstyle\pm6.9}$ & 71.6$_{\scriptstyle\pm5.0}$ & 53.9$_{\scriptstyle\pm2.3}$ & 61.1$_{\scriptstyle\pm6.2}$ & \textbf{74.6$_{\scriptstyle\pm3.3}$} \\
GPT-4.1 mini & 18.2$_{\scriptstyle\pm0.0}$ & 18.2 & - & 49.6$_{\scriptstyle\pm2.7}$ & 49.3 & - & 51.5$_{\scriptstyle\pm3.2}$ & 69.2 & - & 46.0$_{\scriptstyle\pm5.8}$ & 59.1 & - \\
GPT-5 mini & 54.8$_{\scriptstyle\pm10.2}$ & - & - & 72.2$_{\scriptstyle\pm5.4}$ & - & - & 72.5$_{\scriptstyle\pm2.9}$ & - & - & 60.5$_{\scriptstyle\pm2.3}$ & - & - \\
GPT-5 & 51.5$_{\scriptstyle\pm6.1}$ & - & - & 75.6$_{\scriptstyle\pm0.0}$ & - & - & 72.1$_{\scriptstyle\pm1.6}$ & - & - & 68.4$_{\scriptstyle\pm4.8}$ & - & - \\
GPT-6 Luna & 44.4$_{\scriptstyle\pm0.0}$ & - & - & 57.6$_{\scriptstyle\pm6.6}$ & - & - & 66.5$_{\scriptstyle\pm4.3}$ & - & - & 59.0$_{\scriptstyle\pm4.4}$ & - & - \\
GPT-6 Astra & 48.0$_{\scriptstyle\pm6.1}$ & - & - & 61.5$_{\scriptstyle\pm0.1}$ & - & - & 64.9$_{\scriptstyle\pm1.5}$ & - & - & 62.9$_{\scriptstyle\pm1.3}$ & - & - \\
GPT-6 Sol & 32.5$_{\scriptstyle\pm0.0}$ & - & - & 61.5$_{\scriptstyle\pm0.0}$ & - & - & 62.3$_{\scriptstyle\pm1.5}$ & - & - & 60.6$_{\scriptstyle\pm2.6}$ & - & - \\
GPT-6.1 Sol & 40.5$_{\scriptstyle\pm6.9}$ & - & - & 62.7$_{\scriptstyle\pm7.9}$ & - & - & 68.4$_{\scriptstyle\pm1.5}$ & - & - & 63.6$_{\scriptstyle\pm0.0}$ & - & - \\
\midrule
\multicolumn{13}{c}{\textit{RO $\rightarrow$ IT}} \\
\midrule
Llama 3.1 8B & 27.8$_{\scriptstyle\pm4.8}$ & \textbf{77.8$_{\scriptstyle\pm19.2}$} & 25.0$_{\scriptstyle\pm0.0}$ & 42.6$_{\scriptstyle\pm11.1}$ & 55.4$_{\scriptstyle\pm3.0}$ & 68.1$_{\scriptstyle\pm10.2}$ & 35.6$_{\scriptstyle\pm11.2}$ & 38.2$_{\scriptstyle\pm5.8}$ & 58.4$_{\scriptstyle\pm1.6}$ & 43.5$_{\scriptstyle\pm18.5}$ & 39.0$_{\scriptstyle\pm10.4}$ & \textbf{75.6$_{\scriptstyle\pm1.5}$} \\
Gemma 3 27B & 30.0$_{\scriptstyle\pm8.7}$ & 25.0$_{\scriptstyle\pm0.0}$ & 8.3$_{\scriptstyle\pm14.4}$ & 44.0$_{\scriptstyle\pm0.0}$ & 48.4$_{\scriptstyle\pm11.6}$ & 56.5$_{\scriptstyle\pm10.2}$ & 49.3$_{\scriptstyle\pm3.4}$ & 48.6$_{\scriptstyle\pm1.9}$ & 67.4$_{\scriptstyle\pm3.1}$ & 38.5$_{\scriptstyle\pm0.0}$ & 43.5$_{\scriptstyle\pm8.3}$ & 74.3$_{\scriptstyle\pm2.0}$ \\
Gemma 4 26B A4B & 25.0$_{\scriptstyle\pm0.0}$ & 25.0$_{\scriptstyle\pm0.0}$ & 25.0$_{\scriptstyle\pm0.0}$ & 53.7$_{\scriptstyle\pm3.9}$ & 54.3$_{\scriptstyle\pm6.5}$ & 47.0$_{\scriptstyle\pm2.6}$ & 63.4$_{\scriptstyle\pm3.7}$ & 61.3$_{\scriptstyle\pm3.2}$ & 69.5$_{\scriptstyle\pm3.5}$ & 59.8$_{\scriptstyle\pm2.9}$ & 63.7$_{\scriptstyle\pm1.5}$ & 66.5$_{\scriptstyle\pm6.1}$ \\
Gemma 4 31B & 0.0$_{\scriptstyle\pm0.0}$ & 0.0$_{\scriptstyle\pm0.0}$ & 25.0$_{\scriptstyle\pm0.0}$ & 59.7$_{\scriptstyle\pm2.9}$ & 66.4$_{\scriptstyle\pm0.0}$ & \textbf{76.8$_{\scriptstyle\pm9.3}$} & 64.5$_{\scriptstyle\pm0.0}$ & 66.6$_{\scriptstyle\pm3.7}$ & \textbf{80.4$_{\scriptstyle\pm3.3}$} & 64.1$_{\scriptstyle\pm1.5}$ & 62.6$_{\scriptstyle\pm2.9}$ & 74.7$_{\scriptstyle\pm2.4}$ \\
RoMistral 7B & 16.7$_{\scriptstyle\pm14.4}$ & 36.1$_{\scriptstyle\pm42.8}$ & 38.9$_{\scriptstyle\pm24.1}$ & 41.1$_{\scriptstyle\pm15.5}$ & 51.9$_{\scriptstyle\pm11.2}$ & 71.7$_{\scriptstyle\pm0.0}$ & 35.7$_{\scriptstyle\pm13.7}$ & 43.4$_{\scriptstyle\pm6.2}$ & 67.2$_{\scriptstyle\pm0.5}$ & 38.1$_{\scriptstyle\pm18.1}$ & 50.2$_{\scriptstyle\pm7.2}$ & 70.6$_{\scriptstyle\pm4.5}$ \\
RoLlama 3.1 8B & 52.8$_{\scriptstyle\pm24.1}$ & 21.7$_{\scriptstyle\pm20.2}$ & 38.9$_{\scriptstyle\pm24.1}$ & 62.2$_{\scriptstyle\pm2.3}$ & 51.6$_{\scriptstyle\pm22.8}$ & 67.6$_{\scriptstyle\pm11.8}$ & 58.8$_{\scriptstyle\pm3.8}$ & 47.4$_{\scriptstyle\pm4.2}$ & 64.3$_{\scriptstyle\pm5.6}$ & 58.2$_{\scriptstyle\pm2.1}$ & 48.3$_{\scriptstyle\pm11.6}$ & 70.2$_{\scriptstyle\pm1.5}$ \\
RoGemma 2 9B & 52.8$_{\scriptstyle\pm24.1}$ & 25.0$_{\scriptstyle\pm0.0}$ & 8.3$_{\scriptstyle\pm14.4}$ & 68.7$_{\scriptstyle\pm11.7}$ & 73.7$_{\scriptstyle\pm10.7}$ & 74.0$_{\scriptstyle\pm5.8}$ & 60.1$_{\scriptstyle\pm6.7}$ & 62.4$_{\scriptstyle\pm5.5}$ & 62.1$_{\scriptstyle\pm6.7}$ & 56.1$_{\scriptstyle\pm1.6}$ & 66.9$_{\scriptstyle\pm4.1}$ & 62.8$_{\scriptstyle\pm1.9}$ \\
GPT-4.1 mini & 38.9$_{\scriptstyle\pm24.1}$ & 25.0 & - & 40.0$_{\scriptstyle\pm16.4}$ & 61.4 & - & 43.9$_{\scriptstyle\pm9.1}$ & 58.1 & - & 43.4$_{\scriptstyle\pm6.6}$ & 67.6 & - \\
GPT-5 mini & 8.3$_{\scriptstyle\pm14.4}$ & - & - & 34.5$_{\scriptstyle\pm19.9}$ & - & - & 44.2$_{\scriptstyle\pm12.0}$ & - & - & 46.6$_{\scriptstyle\pm10.3}$ & - & - \\
GPT-5 & 0.0$_{\scriptstyle\pm0.0}$ & - & - & 33.8$_{\scriptstyle\pm2.1}$ & - & - & 39.8$_{\scriptstyle\pm1.0}$ & - & - & 44.3$_{\scriptstyle\pm2.4}$ & - & - \\
GPT-6 Luna & 16.7$_{\scriptstyle\pm14.4}$ & - & - & 50.4$_{\scriptstyle\pm12.8}$ & - & - & 53.0$_{\scriptstyle\pm7.3}$ & - & - & 55.1$_{\scriptstyle\pm2.0}$ & - & - \\
GPT-6 Astra & 25.0$_{\scriptstyle\pm0.0}$ & - & - & 45.1$_{\scriptstyle\pm3.0}$ & - & - & 41.0$_{\scriptstyle\pm4.9}$ & - & - & 49.3$_{\scriptstyle\pm3.0}$ & - & - \\
GPT-6 Sol & 35.0$_{\scriptstyle\pm8.7}$ & - & - & 31.9$_{\scriptstyle\pm2.7}$ & - & - & 37.3$_{\scriptstyle\pm3.5}$ & - & - & 45.8$_{\scriptstyle\pm0.0}$ & - & - \\
GPT-6.1 Sol & 25.0$_{\scriptstyle\pm0.0}$ & - & - & 39.6$_{\scriptstyle\pm6.5}$ & - & - & 44.6$_{\scriptstyle\pm0.0}$ & - & - & 46.3$_{\scriptstyle\pm4.0}$ & - & - \\
\bottomrule
\end{tabular}}
\caption{F1-scores breakdown for each model and code-switching generation method, in the zero-shot (ZS), chain-of-thought (CoT), and fine-tuning (FT) settings, for the CS gold standard test set. Bold indicates the best performance for each method and language pair.}
\label{tab:cs_gold_results}
\end{table*}

To better quantify performance differences across target languages and code-switching generation methods, Table~\ref{tab:cs_results} presents the detailed breakdown for all models and language pairs on the full test set, and Table~\ref{tab:cs_gold_results} reports the corresponding breakdown on the gold standard benchmark. We also report fine-tuning results for open-source models, which show significant improvements across all languages and generation methods over zero-shot and chain-of-thought prompting.

We note that the random-replacement strategy naturally yields degraded outputs. However, we applied the method exactly as proposed by \citet{krishnan-etal-2021-multilingual}, without additional filtering, to maintain consistency with prior work. Broken down by model and method, models generally underperform under this strategy. Nevertheless, the performance drop varies substantially across architectures: for GPT-5 mini, the variance across methods is at most 5\%, demonstrating higher robustness to noisy lexical substitutions. In other cases, random insertion acts as an adversarial perturbation, reducing performance by more than 10\% compared to syntactically informed methods. Fine-tuning on this data yields results competitive with other code-switching generation methods, indicating that models can adapt effectively even in noisy environments. While not aimed at maximizing naturalness or simulating real-world usage, it serves as a valid, fast data augmentation strategy and provides a useful stress test for assessing model robustness. The dependency-based method is the most resilient, as it preserves syntactic dependencies from the original matrix language. For top-performing models like Gemma 4 31B and GPT-5 mini, zero-shot performance shows no statistically significant degradation compared to the monolingual counterpart.

When comparing Romance languages against Germanic ones, we note that higher F1-scores are achieved in French and Italian code-mixing, with less degradation, than in English and German. Across zero-shot settings on open-weights models, German sees the steepest degradation (e.g., down to approx. 35\% on RoMistral 7B).

Chain-of-thought may help improve performance in smaller models; for example, RoMistral 7B sees gains ranging from +8 to +13 points in F1-score. On the proprietary GPT-4.1 mini model, the gains remain more limited between +2 to +7 points in F1-score. In contrast, on RoLlama 3.1 8B, we notice a consistent performance degradation across all methods and languages, of about -5 to -10 percentage points in macro-F1. Under fine-tuning, on the other hand, models tend to recover, reaching over 70\% in noisy random code-mixing, reducing the performance gap across models. Gemma 4 31B achieves top performance across nearly all language and method combinations, in the fine-tuning regime.

On the CS-SeLeRoSa-Gold set, we see similar trends, with the highest performance being achieved on phrase- and LLM-based generation, where GPT-5 reaches approx. 70\% in F1-score, and GPT-5 mini outperforms it by 2\%. Likewise, Gemma 4 31B achieves the highest performance while fine-tuned. Despite filtering out unnatural code-mixed sentences after the human annotation step, models still tend to underperform, especially under the random method.

\subsection{Method-Level Degradation Analysis}
\label{app:degradation_analysis}

\begin{table}[!th]
\centering
\small
\setlength{\tabcolsep}{4pt}
\begin{tabular}{llcc}
\toprule
\textbf{Model} & \textbf{Method} & \textbf{F1 (\%)} & $\Delta$ (\%) \\
\midrule
Llama 3.1 8B & None & 46.3 & - \\
 & Random & \underline{43.3} & \underline{-3.0} \\
 & Phrase & 44.7 & -1.6 \\
 & Dependency & \textbf{46.4} & \textbf{+0.0} \\
 & LLM & 43.9 & -2.4 \\
\midrule
Gemma 3 27B & None & 49.9 & - \\
 & Random & \underline{31.1} & \underline{-18.8} \\
 & Phrase & 40.4 & -9.6 \\
 & Dependency & \textbf{41.9} & \textbf{-8.0} \\
 & LLM & 34.2 & -15.7 \\
\midrule
Gemma 4 26B A4B & None & 67.0 & - \\
 & Random & \underline{59.1} & \underline{-7.9} \\
 & Phrase & 63.8 & -3.1 \\
 & Dependency & \textbf{66.2} & \textbf{-0.8} \\
 & LLM & 65.3 & -1.6 \\
\midrule
Gemma 4 31B & None & 70.2 & - \\
 & Random & \underline{65.6} & \underline{-4.5} \\
 & Phrase & 68.4 & -1.8 \\
 & Dependency & \textbf{69.5} & \textbf{-0.7} \\
 & LLM & 69.0 & -1.2 \\
\midrule
RoMistral 7B & None & 41.2 & - \\
 & Random & \underline{34.6} & \underline{-6.6} \\
 & Phrase & 37.2 & -4.0 \\
 & Dependency & \textbf{38.2} & \textbf{-3.0} \\
 & LLM & 36.7 & -4.5 \\
\midrule
RoLlama 3.1 8B & None & 61.6 & - \\
 & Random & \underline{44.7} & \underline{-16.9} \\
 & Phrase & 54.7 & -6.9 \\
 & Dependency & \textbf{56.7} & \textbf{-4.9} \\
 & LLM & 52.9 & -8.7 \\
\midrule
RoGemma 2 9B & None & 60.3 & - \\
 & Random & \underline{49.9} & \underline{-10.4} \\
 & Phrase & 55.8 & -4.5 \\
 & Dependency & \textbf{58.1} & \textbf{-2.3} \\
 & LLM & 55.4 & -4.9 \\
\midrule
GPT-4.1 mini & None & 63.3 & - \\
 & Random & \underline{49.2} & \underline{-14.1} \\
 & Phrase & 56.7 & -6.6 \\
 & Dependency & \textbf{60.8} & \textbf{-2.5} \\
 & LLM & 56.3 & -7.0 \\
\midrule
GPT-5 mini & None & 71.3 & - \\
 & Random & \underline{66.9} & \underline{-4.4} \\
 & Phrase & 70.1 & -1.2 \\
 & Dependency & \textbf{70.5} & \textbf{-0.9} \\
 & LLM & 70.5 & -0.9 \\
\midrule
GPT-6 Luna & None & 68.1 & - \\
 & Random & \underline{65.5} & \underline{-2.7} \\
 & Phrase & 66.9 & -1.2 \\
 & Dependency & \textbf{68.3} & \textbf{+0.2} \\
 & LLM & 67.2 & -0.9 \\
\bottomrule
\end{tabular}
\caption{F1-scores and performance degradation ($\Delta$) relative to the monolingual baseline (None) across code-switching generation methods. Bold indicates the lowest degradation (i.e., best performance) among code-switching methods for each model, while underline indicates the highest degradation (i.e., worst performance).}
\label{tab:degradation_method}
\end{table}

Degradation varies across models and methods. Analyzing the results broken down by method (see Table~\ref{tab:degradation_method}, reporting the mean F1 degradation across 3 seeds for open-weights models, and a single run for closed-source models), we observe that for most models, the worst degradation is observed when employing the random code-switching method, reaching up to 18.8\% (Gemma 3 27B) and 16.9\% (RoLlama 3.1 8B). On the other side of the spectrum, the dependency-based method, which preserves syntactic structure, causes a much smaller degradation, under 5\% for almost all models (and near zero for Gemma 4 31B, GPT-5 mini, and GPT-6 Luna). GPT-5 mini remains robust against all code-switching methods, obtaining a maximum degradation of only -4.4\% under the random method. Code-switching generation causes statistically significant degradation when compared with the original Romanian sentences for closed-source models ($p < 0.001$ under paired McNemar's test on instance-level predictions) and open-source models ($p < 0.01$ under Welch's $t$-test across $N=3$ runs), except for Llama 3.1 8B and RoMistral 7B due to high run-to-run variance caused by instruction-following failures and invalid outputs.

\subsection{Failure Modes Analysis}
\label{sec:error_analysis}

\begin{table}[!th]
\centering
\small
\setlength{\tabcolsep}{4pt}
\begin{tabular}{llcc}
\toprule
\textbf{Model} & \textbf{Method} & \textbf{FPR (\%)} & \textbf{FNR (\%)} \\
\midrule
Llama 3.1 8B & None & 47.2 & 47.3 \\
 & Random & 57.4 & 34.1 \\
 & Phrase & 52.4 & 39.3 \\
 & Dependency & 49.7 & 40.3 \\
 & LLM & 52.5 & 40.2 \\
\midrule
Gemma 3 27B & None & 76.3 & 4.2 \\
 & Random & 97.9 & 0.3 \\
 & Phrase & 88.6 & 0.9 \\
 & Dependency & 86.5 & 2.7 \\
 & LLM & 94.8 & 1.1 \\
\midrule
Gemma 4 26B A4B & None & 73.1 & 12.2 \\
 & Random & 83.9 & 11.8 \\
 & Phrase & 79.5 & 11.1 \\
 & Dependency & 75.2 & 13.9 \\
 & LLM & 80.7 & 12.3 \\
\midrule
Gemma 4 31B & None & 37.5 & 19.7 \\
 & Random & 58.8 & 14.9 \\
 & Phrase & 48.1 & 15.8 \\
 & Dependency & 41.8 & 20.2 \\
 & LLM & 47.0 & 16.7 \\
\midrule
RoMistral 7B & None & 56.3 & 44.3 \\
 & Random & 62.9 & 53.5 \\
 & Phrase & 60.7 & 47.0 \\
 & Dependency & 55.2 & 53.3 \\
 & LLM & 60.3 & 49.3 \\
\midrule
RoLlama 3.1 8B & None & 51.2 & 18.8 \\
 & Random & 82.5 & 4.6 \\
 & Phrase & 66.9 & 10.7 \\
 & Dependency & 62.2 & 14.2 \\
 & LLM & 69.5 & 10.8 \\
\midrule
RoGemma 2 9B & None & 52.2 & 20.4 \\
 & Random & 72.1 & 13.0 \\
 & Phrase & 62.5 & 15.9 \\
 & Dependency & 55.8 & 21.6 \\
 & LLM & 61.4 & 18.8 \\
\midrule
GPT-4.1 mini & None & 50.6 & 16.0 \\
 & Random & 75.0 & 8.7 \\
 & Phrase & 64.2 & 10.2 \\
 & Dependency & 54.8 & 16.2 \\
 & LLM & 63.8 & 11.8 \\
\midrule
GPT-5 mini & None & 28.2 & 28.0 \\
 & Random & 26.0 & 40.3 \\
 & Phrase & 29.6 & 29.1 \\
 & Dependency & 26.4 & 32.2 \\
 & LLM & 29.2 & 28.7 \\
\midrule
GPT-6 Luna & None & 46.6 & 10.0 \\
 & Random & 49.5 & 12.2 \\
 & Phrase & 49.4 & 8.9 \\
 & Dependency & 46.5 & 10.1 \\
 & LLM & 48.7 & 9.1 \\
\bottomrule
\end{tabular}
\caption{Breakdown of the False Positive Rate (FPR) and False Negative Rate (FNR) for each model across code-switching generation methods. The \textit{None} method indicates the original monolingual baseline. Unparseable model outputs are counted as classification errors.}
\label{tab:fpr_fnr_method}
\end{table}

\begin{table}[!th]
\centering
\small
\setlength{\tabcolsep}{4pt}
\begin{tabular}{llcc}
\toprule
\textbf{Model} & \textbf{Language} & \textbf{FPR (\%)} & \textbf{FNR (\%)} \\
\midrule
Llama 3.1 8B & EN & 51.6 & 37.9 \\
 & DE & 56.0 & 38.9 \\
 & FR & 51.3 & 42.0 \\
 & IT & 53.2 & 35.3 \\
 & RO & 47.2 & 47.3 \\
\midrule
Gemma 3 27B & EN & 93.2 & 1.2 \\
 & DE & 93.0 & 1.0 \\
 & FR & 91.8 & 1.2 \\
 & IT & 90.1 & 1.5 \\
 & RO & 76.3 & 4.2 \\
\midrule
Gemma 4 26B A4B & EN & 83.0 & 9.4 \\
 & DE & 80.0 & 12.2 \\
 & FR & 79.0 & 14.1 \\
 & IT & 77.3 & 13.4 \\
 & RO & 73.1 & 12.2 \\
\midrule
Gemma 4 31B & EN & 51.7 & 16.4 \\
 & DE & 50.7 & 16.5 \\
 & FR & 49.2 & 15.9 \\
 & IT & 44.3 & 18.9 \\
 & RO & 37.5 & 19.7 \\
\midrule
RoMistral 7B & EN & 54.9 & 48.9 \\
 & DE & 57.7 & 56.8 \\
 & FR & 61.8 & 54.4 \\
 & IT & 64.7 & 43.1 \\
 & RO & 56.3 & 44.3 \\
\midrule
RoLlama 3.1 8B & EN & 66.9 & 11.8 \\
 & DE & 74.9 & 8.0 \\
 & FR & 68.9 & 10.8 \\
 & IT & 70.6 & 9.7 \\
 & RO & 51.2 & 18.8 \\
\midrule
RoGemma 2 9B & EN & 69.0 & 14.0 \\
 & DE & 65.0 & 15.6 \\
 & FR & 61.1 & 19.3 \\
 & IT & 57.0 & 20.5 \\
 & RO & 52.2 & 20.4 \\
\midrule
GPT-4.1 mini & EN & 64.9 & 11.0 \\
 & DE & 67.9 & 11.2 \\
 & FR & 64.9 & 11.5 \\
 & IT & 60.5 & 13.1 \\
 & RO & 50.6 & 16.0 \\
\midrule
GPT-5 mini & EN & 29.3 & 31.8 \\
 & DE & 28.6 & 32.9 \\
 & FR & 28.1 & 31.6 \\
 & IT & 25.2 & 34.0 \\
 & RO & 28.2 & 28.0 \\
\midrule
GPT-6 Luna & EN & 47.9 & 11.3 \\
 & DE & 51.1 & 9.3 \\
 & FR & 48.3 & 9.6 \\
 & IT & 46.8 & 10.1 \\
 & RO & 46.6 & 10.0 \\
\bottomrule
\end{tabular}
\caption{Breakdown of False Positive Rate (FPR) and False Negative Rate (FNR) for each model across different target languages. \textit{RO} indicates the original monolingual Romanian baseline. Unparseable model outputs are counted as classification errors.}
\label{tab:fpr_fnr_language}
\end{table}

\begin{table}[!th]
\centering
\small
\setlength{\tabcolsep}{4pt}
\begin{tabular}{llcc}
\toprule
\textbf{Model} & \textbf{Setting} & \textbf{FPR (\%)} & \textbf{FNR (\%)} \\
\midrule
Llama 3.1 8B & Zero-Shot & 63.8 & 49.6 \\
 & CoT & 74.9 & 19.1 \\
 & Fine-Tuning & 20.1 & 30.2 \\
\midrule
Gemma 3 27B & Zero-Shot & 92.0 & 1.2 \\
 & CoT & 87.0 & 2.6 \\
 & Fine-Tuning & 26.1 & 21.5 \\
\midrule
Gemma 4 26B A4B & Zero-Shot & 51.1 & 14.5 \\
 & CoT & 49.8 & 15.1 \\
 & Fine-Tuning & 38.0 & 17.8 \\
\midrule
Gemma 4 31B & Zero-Shot & 40.6 & 19.1 \\
 & CoT & 44.9 & 16.9 \\
 & Fine-Tuning & 22.2 & 21.7 \\
\midrule
RoMistral 7B & Zero-Shot & 64.1 & 64.6 \\
 & CoT & 69.8 & 30.2 \\
 & Fine-Tuning & 17.9 & 33.4 \\
\midrule
RoLlama 3.1 8B & Zero-Shot & 70.5 & 10.9 \\
 & CoT & 78.1 & 12.9 \\
 & Fine-Tuning & 26.1 & 26.0 \\
\midrule
RoGemma 2 9B & Zero-Shot & 62.7 & 17.6 \\
 & CoT & 52.1 & 22.5 \\
 & Fine-Tuning & 22.2 & 26.4 \\
\midrule
GPT-4.1 mini & Zero-Shot & 64.5 & 11.7 \\
 & CoT & 52.2 & 17.3 \\
\midrule
GPT-5 mini & Zero-Shot & 27.8 & 32.6 \\
\midrule
GPT-6 Luna & Zero-Shot & 48.5 & 10.1 \\
\bottomrule
\end{tabular}
\caption{Comparison of False Positive Rate (FPR) and False Negative Rate (FNR) across zero-shot (ZS), chain-of-thought (CoT), and fine-tuning (FT) settings. Unparseable model outputs are counted as classification errors.}
\label{tab:fpr_fnr_setting}
\end{table}

We showcase the failure modes (i.e., False Positive Rate (FPR) and False Negative Rate (FNR)) for every model across methods, languages, and prompting/fine-tuning settings. In the language and method breakdown (Tables~\ref{tab:fpr_fnr_method} and \ref{tab:fpr_fnr_language}), for open-weights models, we report the results aggregated using majority voting across 3 runs, whereas for closed-source models, we report single runs.

This analysis reveals that in zero-shot settings, most models tend to exhibit a very high FPR combined with a very low FNR, as seen in the cases of Gemma 3 27B and RoLlama 3.1 8B. This means that these models rarely miss actual satire (i.e., low FNR), but they constantly misclassify regular news as satire (i.e., high FPR). Investigating these results more closely, we notice a consistent pattern where the models perform best (lowest FPR) on syntactically structured methods such as dependency-based, followed by phrase-based and LLM-based generation methods, and worst on random code-switching. For well-performing localized models (i.e., RoLlama 3.1 8B, RoGemma 2 9B, and Gemma 3 27B) as well as GPT-4.1 mini, the increase in FPR from the monolingual baseline to any code-switching setting is statistically significant (p < 0.05 under Welch's t-test for open-source models and p < 0.01 under a Chi-squared test for GPT-4.1 mini). GPT-5 mini remains an exception, maintaining a robust and statistically invariant FPR between 26--30\% (p > 0.09) regardless of the code-switching method. For Llama 3.1 8B and RoMistral 7B, this observation is not consistent due to high variance in the predictions, which matches their overall low performance and instability on the dataset. Conversely, when looking at the FNR, it drops as the generation method becomes noisier. Since models over-classify these noisy sentences as satirical, they tend to miss very few true satirical sentences.

The language breakdown in Table~\ref{tab:fpr_fnr_language} demonstrates that, across almost all models (with the exception of RoMistral 7B), mixing Romanian (i.e., a Romance language) with other Romance languages (i.e., French or Italian) consistently obtains a lower FPR compared to Germanic languages (i.e., English or German). This difference is highly statistically significant at the instance level for the Gemma models (p < 0.001 under a Welch's t-test) and both GPT models (p < 1e-9 under a Chi-squared test), suggesting that cognate sharing and syntactic compatibility mitigate false-positive satire detection.

Finally, Table~\ref{tab:fpr_fnr_setting} (reporting the mean FPR and FNR across 3 seeds for open-weights models, and a single run for proprietary models) shows that while zero-shot and chain-of-thought prompting strategies are vulnerable to code-switching noise, supervised fine-tuning on the CS-SeLeRoSa dataset successfully mitigates this bias. Fine-tuning drastically reduces the FPR (e.g., Gemma 3 27B drops from 92.0\% to 26.1\% FPR; RoLlama 3.1 8B drops from 70.5\% to 26.1\% FPR) but also increases the FNR. As the decision boundary shifts during fine-tuning, the models become more selective and miss some of the harder satire examples.

\subsection{Baseline RoLlama 3.1 8B Chain-of-Thought Degradation}
\label{app:rollama_baseline_degradation}

\begin{table}[t]
\centering
\small
\setlength{\tabcolsep}{6pt}
\begin{tabular}{lccc}
\toprule
\textbf{Metric} & \textbf{Seed 1} & \textbf{Seed 2} & \textbf{Seed 3} \\
\midrule
\multicolumn{4}{c}{\textit{Zero-Shot}} \\
\midrule
Predicted Positive & 1,343 & 1,325 & 1,170 \\
Predicted Negative & 721 & 737 & 780 \\
Unparseable & 8 & 10 & 122 \\
\midrule
\multicolumn{4}{c}{\textit{Chain-of-Thought}} \\
\midrule
Predicted Positive & 1,665 & 1,549 & 1,613 \\
Predicted Negative & 389 & 493 & 446 \\
Unparseable & 18 & 30 & 13 \\
\bottomrule
\end{tabular}
\caption{Prediction counts of RoLlama 3.1 8B on the monolingual baseline test split, across each seed.}
\label{tab:rollama_prediction_distribution}
\end{table}

\begin{table*}[t]
\centering
\small
\setlength{\tabcolsep}{4pt}
\begin{tabular}{cc>{\raggedright\arraybackslash}p{4.8cm}c>{\raggedright\arraybackslash}p{6.4cm}}
\toprule
\textbf{ID} & \textbf{Count} & \textbf{Topic Label} & \textbf{Share (\%)} & \textbf{Top c-TF-IDF Keywords} \\
\midrule
\multicolumn{5}{c}{\textit{Reasoning Traces}} \\
\midrule
0 & 319 & Interpreting Satire and Irony & 96.1 & cuvânt, afirmație, exprima, sentiment, satiric, spune \\
1 & 13 & Romanian Satire Explanations & 3.9 & românesc, român, satiric, cuvânt, sarcasm, comentariu \\
\midrule
\multicolumn{5}{c}{\textit{Input Sentences}} \\
\midrule
$-1$ & 131 & People Leaving and Changing Positions (Outliers) & 39.5 & pleca, rămâne, veni, începe, urmare, întâmpla \\
0 & 71 & Reported Problems and Conditions & 21.4 & totuşi, pericol, fapt, explica, spune, convinşi \\
1 & 36 & Romania Satirical Discoveries & 10.8 & gpe, susţine, panică, deveni, confirma, crede \\
2 & 35 & Personal Experiences And Problems & 10.5 & simţi, ştiam, rău, povesti, greu, puţin \\
3 & 25 & Disjointed News Snippets & 7.5 & arăta, vedea, recunoaște, crede, povesti, respectiv \\
4 & 12 & Romanian Society and Everyday Life & 3.6 & român, mulţi, comun, călător, printre, internaţional \\
5 & 11 & Fragmented News Excerpts & 3.3 & viu, genial, spune, urmă, exista, veni \\
6 & 11 & Office Email and Time Off & 3.3 & mail, trimite, pleca, acasă, oră, muncă \\
\bottomrule
\end{tabular}
\caption{Topics found by BERTopic for reasoning traces and input sentences across the 332 false positive errors on the RoLlama 3.1 8B, seed 1, when evaluated on the monolingual SeLeRoSa test set.}
\label{tab:cot_bertopic_clusters}
\end{table*}

In Table~\ref{tab:baseline_results}, as opposed to the other open-source models, RoLlama 3.1 8B underperforms in the chain-of-thought setting on the monolingual baseline results, when compared to the zero-shot setting. Specifically, in the zero-shot setting, the model achieved $61.0_{\scriptstyle\pm0.3}$\% accuracy and $61.6_{\scriptstyle\pm0.8}$\% in F1-score, while in the chain-of-thought setting, the model achieved $51.9_{\scriptstyle\pm1.5}$\% accuracy and $50.4_{\scriptstyle\pm2.1}$\% in F1-score. This represents a roughly 10 percentage point decrease across both metrics; hence, we further analyze this negative result. Table~\ref{tab:rollama_prediction_distribution} presents the prediction distribution across each seed. Since the trends are similar across seeds, we focus our analysis on seed 1. Out of 2,072 samples, in the zero-shot setting the model predicted 1,343 samples as satire (64.8\%), 721 samples as factual (34.8\%), and 8 samples were unparseable (0.4\%). Using chain-of-thought, the model predicted 1,665 samples as positive (80.4\%), 389 as negative (18.8\%), and 18 were unparseable (0.9\%). We notice a considerable increase in the predicted positive rate of about 15.6 percentage points. Furthermore, among samples correctly classified in zero-shot, 330 factual samples switched to satirical and 6 became unparseable (336 new false positives under CoT), while only 45 satirical samples switched to factual and 5 became unparseable (50 new false negatives).

We inspect the input sentences and reasoning traces as a proxy to assess the model's reasons for predicting satire on samples that switched their label. We use BERTopic~\cite{DBLP:journals/corr/abs-2203-05794} to extract the main topics from the text. First, we generate an embedding representation using the multilingual Sentence-BERT\footnote{We employ the \texttt{paraphrase-\allowbreak multilingual-\allowbreak MiniLM-\allowbreak L12-v2} model.}~\cite{reimers-2019-sentence-bert} model, use UMAP~\cite{mcinnes2018umap} dimensionality reduction to 5 components with cosine metric and 8 neighbours, clustering with at least 10 samples, and c-TF-IDF~\cite{DBLP:journals/corr/abs-2203-05794} for filtering filler words. The representative keywords were extracted following maximal marginal relevance with a diversity of 0.3, and KeyBERT~\cite{grootendorst2020keybert}, and we employ OpenAI's GPT-6 Luna for human-readable representation labels. Table~\ref{tab:cot_bertopic_clusters} shows the topics identified. 

Analyzing the reasoning traces, we notice that the model overassumes that certain situations or phrases are associated with satire. For example, in the sentence \textit{``Îmi pot imagina mii de aplicaţii pentru această invenţie revoluţionară.''} (Engl., \textit{``I can imagine lots of applications for this revolutionary invention''}), the model assumes that \textit{``a high level of optimism and positivity is associated with satire''}, and \textit{``even without enough context, [it] considers it satirical''}. In another example, \textit{``Prompți, ca întotdeauna!''} (Engl., \textit{``Prompt, as always!''}) the model assumes that the tone is too formal for an everyday conversation, and \textit{``as always''} suggests irony, although the ground truth suggests otherwise. 

When looking at the input sentences that changed their label, we notice several main topics that were affected, including \textit{``Reported Problems and Conditions''}, \textit{``Romania Satirical Discoveries''}, and \textit{``Personal Experiences and Problems''}. Additionally, a large portion of the affected sentences (39.5\%, corresponding to the outlier cluster \textit{``People Leaving and Changing Positions''}) describe specific individuals and actions, which may also suggest a bias, as satirical news outlets may frequently caricature public figures. This further emphasizes that the model considered colloquial situations as possibly satirical.

\subsection{Instruction Following Error Rate}
\label{app:instruction_following_error_rate}

In Tables~\ref{tab:parse_failure_rates} and \ref{tab:parse_failure_rates_gs}, we present the instruction-following error rates where models did not produce the expected output given in the prompt instructions, or used during fine-tuning. The reported values represent the average invalid rate across the 3 seeds for open-weights models, and a single run for proprietary models. The highest error rates are obtained by Llama 3.1 8B and RoMistral 7B in zero-shot (23\%--25\%) and chain-of-thought (6\%--11\%) settings, while in the fine-tuning setting, error rates drop to almost 0\%. Gemma 3 27B, Gemma 4 31B, and the GPT model family achieve under 0.1\% formatting error rates. In most cases, the models produced an invalid label while otherwise adhering to the XML format requirement (e.g., instead of ``satiră'', they produced ``satire'', ``satură'', ``ironie'', or ``ironic''). In some cases, the models produced different XML tags (e.g., using the Romanian translation for ``response'', ``răspuns''), failed to generate the XML tags, or refused to solve the task.

\begin{table}[!th]
\centering
\small
\setlength{\tabcolsep}{4pt}
\begin{tabular}{lccc}
\toprule
\textbf{Model} & \textbf{ZS (\%)} & \textbf{CoT (\%)} & \textbf{FT (\%)} \\
\midrule
Llama 3.1 8B & 24.00 & 7.91 & 0.13 \\
Gemma 3 27B & 0.00 & 0.00 & 0.02 \\
Gemma 4 26B A4B & 0.12 & 0.29 & 0.15 \\
Gemma 4 31B & 0.07 & 0.04 & 0.00 \\
RoMistral 7B & 24.52 & 10.97 & 0.00 \\
RoLlama 3.1 8B & 2.02 & 0.88 & 0.02 \\
RoGemma 2 9B & 0.00 & 0.46 & 0.00 \\
GPT-4.1 mini & 0.00 & 0.00 & - \\
GPT-5 mini & 0.00 & - & - \\
GPT-6 Luna & 0.03 & - & - \\
\bottomrule
\end{tabular}
\caption{Rate of unparseable model outputs on the CS-SeLeRoSa test set.}
\label{tab:parse_failure_rates}
\end{table}

\begin{table}[!th]
\centering
\small
\setlength{\tabcolsep}{4pt}
\begin{tabular}{lccc}
\toprule
\textbf{Model} & \textbf{ZS (\%)} & \textbf{CoT (\%)} & \textbf{FT (\%)} \\
\midrule
Llama 3.1 8B & 23.10 & 6.56 & 0.00 \\
Gemma 3 27B & 0.00 & 0.00 & 0.00 \\
Gemma 4 26B A4B & 0.00 & 0.43 & 0.34 \\
Gemma 4 31B & 0.00 & 0.00 & 0.00 \\
RoMistral 7B & 24.72 & 11.25 & 0.00 \\
RoLlama 3.1 8B & 2.81 & 1.36 & 0.09 \\
RoGemma 2 9B & 0.00 & 0.85 & 0.00 \\
GPT-4.1 mini & 0.00 & 0.00 & - \\
GPT-5 mini & 0.00 & - & - \\
GPT-5 & 0.00 & - & - \\
GPT-6 Luna & 0.00 & - & - \\
GPT-6 Astra & 0.00 & - & - \\
GPT-6 Sol & 0.00 & - & - \\
GPT-6.1 Sol & 0.00 & - & - \\
\bottomrule
\end{tabular}
\caption{Rate of unparseable model outputs on the CS-SeLeRoSa-Gold test set.}
\label{tab:parse_failure_rates_gs}
\end{table}

\clearpage
\section{t-SNE Visualization}
\label{app:tsne}

\begin{figure*}[!b]
\centering
\includegraphics[width=0.98\textwidth]{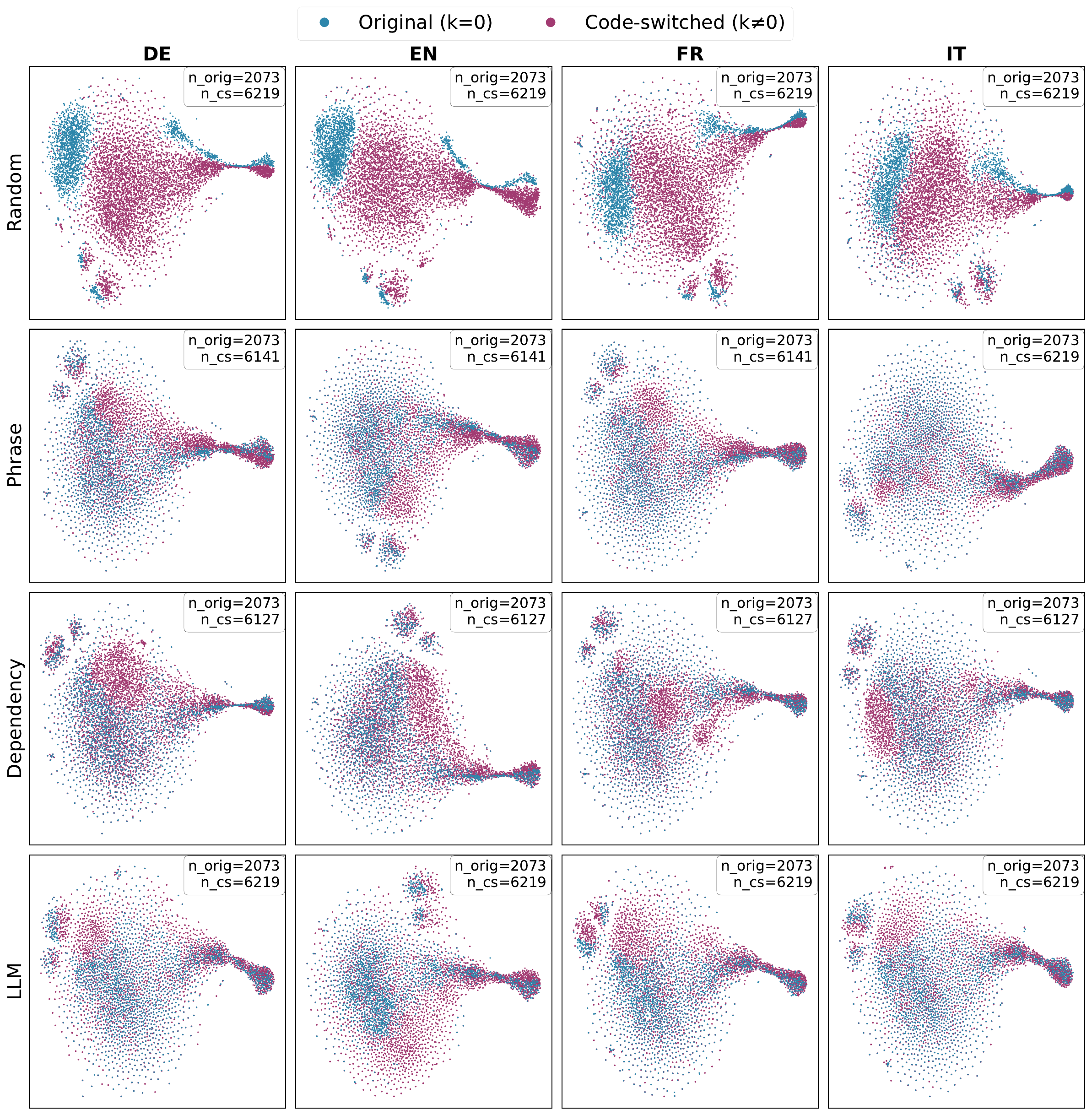}
\caption{t-SNE visualization of sentence embeddings from different languages (EN - English, DE - German, FR - French, IT - Italian) and code-switching generation methods.}
\label{fig:tsne}
\end{figure*}

In Figure~\ref{fig:tsne}, we present the t-SNE visualization (perplexity of 500) of the original and code-mixed sentences, generated using each method on the four languages: English (EN), German (DE), French (FR), and Italian (IT). We embed the text using the multilingual BERT-cased model\footnote{\url{https://huggingface.co/google-bert/bert-base-multilingual-cased}}. The number of generated samples (i.e., \texttt{n\_cs}) may differ between methods since not all produced exactly $k=3$ variants.
We observe that random code-mixing yields separable representations compared to other approaches, indicating a semantic shift when words are randomly translated. This is also emphasized by the semantic similarity results in Table~\ref{tab:cs_semantic_jaccard_scores}, and the annotation process that filtered out a large portion of the random and phrase-generated samples. On the other hand, the other approaches generate code-mixed examples that are closer to the original distribution by considering more context when translating text chunks, thereby better preserving semantics. Across languages, we do not see meaningful differences in the representations, which is expected, given that the goal of the synthetic generation was to preserve the original meaning of the sentence by providing translated variations.

\clearpage
\section{LLM Prompts}
\label{app:prompts}

\prompt{18em}{Prompt for Machine Translation}
{
You are an expert translator. Your task is to translate text from \{src\_lang\} to \{trg\_lang\}.\\
Translation Instructions:\\
~\\
- Translate all content inside <input> tags to \{trg\_lang\}.\\
- Preserve the original formatting, punctuation, and line breaks.\\
- If any words, phrases, or named entities are already translated or are in the target language, leave them unchanged (even if this results in duplicates).\\
- Do not translate code, numbers, URLs, or named entities; keep them as they are.\\
- If the text is code-mixed or partially translated, only translate the parts in \{src\_lang\}.\\
- Output ONLY the translation, wrapped between <output> and </output> tags, with no extra commentary or explanation.
}

\prompt{18em}{Prompt for Generating Code-Mixed Sentences}
{
You are a bilingual speaker of \{src\_lang\} and \{trg\_lang\} who frequently mixes words from both languages in conversations. \\
~\\
You will receive a sentence in \{src\_lang\}. Your task is to generate code-mixed text by naturally mixing words or phrases from the \{src\_lang\} language with the \{trg\_lang\} language.\\
You **must** preserve the meaning and maintain grammatical coherence of the original sentence. Try to mix the languages as naturally as possible, similar to how bilingual speakers would.\\
At most several word or phrases should be translated \{trg\_lang\} language. Keep most words from the sentence in \{src\_lang\} language. \\
~\\
You will be provided with a sentence in the source language between <input> and </input> tags.\\
Return only the code-mixed text without explanations, between <output> and </output> tags.
}

\prompt{18em}{Romanian Prompt for Satire Detection}
{
Ești un expert in detecția satirei și a ironiei în propoziții code-mixed cu limba română și \{target\_lang\}. O să primești o propoziție și trebuie să o adnotezi cu \textasciigrave\{class0\}\textasciigrave\ sau \textasciigrave\{class1\}\textasciigrave.\\
~\\
Răspunsul final îl vei oferi în următorul format:\\
~\\
\textasciigrave\textasciigrave\textasciigrave\\
<response>{[RĂSPUNS]}</response>\\
\textasciigrave\textasciigrave\textasciigrave\\
~\\
unde {[RĂSPUNS]} este \textasciigrave\{class0\}\textasciigrave\ sau \textasciigrave\{class1\}\textasciigrave.\\
~\\
Nu oferi explicații sau să zici altceva decât "<response>{[RĂSPUNS]}</response>". Te rog să respecți formatul cerut, altfel răspunsul va fi considerat greșit.
}

\prompt{18em}{English Translated Prompt for Satire Detection}
{
You are an expert in detecting satire and irony in code-mixed sentences combining Romanian and \{target\_lang\}. You will receive a sentence and must annotate it with \textasciigrave\{class0\}\textasciigrave\ or \textasciigrave\{class1\}\textasciigrave.\\
~\\
Provide the final answer in the following format:\\
~\\
\textasciigrave\textasciigrave\textasciigrave\\
<response>{[RESPONSE]}</response>\\
\textasciigrave\textasciigrave\textasciigrave\\
~\\
where {[RESPONSE]} is \textasciigrave\{class0\}\textasciigrave\ or \textasciigrave\{class1\}\textasciigrave.\\
~\\
Do not provide explanations or say anything other than "<response>{[RESPONSE]}</response>". Please respect the required format, otherwise the answer will be considered incorrect.
}

\prompt{18em}{Romanian Prompt for Satire Detection using Chain-of-Thought}
{
Ești un expert in detecția satirei și a ironiei în propoziții code-mixed cu limba română și \{target\_lang\}. O să primești o propoziție și trebuie să o adnotezi cu \textasciigrave\{class0\}\textasciigrave\ sau \textasciigrave\{class1\}\textasciigrave.\\
~\\
Te rog să te gândești pas cu pas înainte de a răspunde. \\
~\\
Răspunsul final îl vei oferi în următorul format: \\
~\\
\textasciigrave\textasciigrave\textasciigrave\\
<reasoning>\\
{[RAȚIONAMENTUL TĂU]}\\
</reasoning>\\
<response>{[RĂSPUNS]}</response>\\
\textasciigrave\textasciigrave\textasciigrave\\
~\\
unde {[RAȚIONAMENTUL TĂU]} este raționamentul tău pas cu pas de maxim 50 de cuvinte și {[RĂSPUNS]} este \textasciigrave\{class0\}\textasciigrave\ sau \textasciigrave\{class1\}\textasciigrave.\\
~\\
Te rog să respecți formatul cerut, altfel răspunsul va fi considerat greșit.
}

\prompt{18em}{English Translated Prompt for Satire Detection using Chain-of-Thought}
{
You are an expert in detecting satire and irony in code-mixed sentences combining Romanian and \{target\_lang\}. You will receive a sentence and must annotate it with \textasciigrave\{class0\}\textasciigrave\ or \textasciigrave\{class1\}\textasciigrave.\\
~\\
Please think step by step before answering. \\
~\\
Provide the final answer in the following format: \\
~\\
\textasciigrave\textasciigrave\textasciigrave\\
<reasoning>\\
{[YOUR REASONING]}\\
</reasoning>\\
<response>{[RESPONSE]}</response>\\
\textasciigrave\textasciigrave\textasciigrave\\
~\\
where {[YOUR REASONING]} is your step-by-step reasoning of maximum 50 words and {[RESPONSE]} is \textasciigrave\{class0\}\textasciigrave\ or \textasciigrave\{class1\}\textasciigrave.\\
~\\
Please respect the required format, otherwise the answer will be considered incorrect.
}

\prompt{18em}{Romanian Fine-tuning System Prompt}
{
Ești un expert în detecția satirei și a ironiei în propoziții code-mixed cu limba română și \{target\_lang\}. O să primești o propoziție și trebuie să o adnotezi cu \textasciigrave\{class0\}\textasciigrave\ sau \textasciigrave\{class1\}\textasciigrave.
}

\prompt{18em}{English Translation for Fine-tuning System Prompt}
{
You are an expert in detecting satire and irony in code-mixed sentences combining Romanian and \{target\_lang\}. You will receive a sentence and must annotate it with \textasciigrave\{class0\}\textasciigrave\ or \textasciigrave\{class1\}\textasciigrave.
}

\prompt{18em}{Romanian Fine-tuning User Input Prompt}
{
PROPOZIȚIE: \{sentence\}
}

\prompt{18em}{English Translation User Input Prompt}
{
SENTENCE: \{sentence\}
}

\prompt{18em}{Fine-tuning Assistant Prompt}
{
<response>\{answer\}</response>
}

\end{document}